\documentclass{article}

\usepackage{iclr2027_conference,times}
\usepackage[T1]{fontenc}
\usepackage{amsmath,amssymb}
\usepackage{graphicx}
\usepackage{booktabs,longtable,array,calc}
\usepackage{xcolor}
\usepackage{microtype}
\usepackage{xurl}
\usepackage{hyperref}
\usepackage{url}

\hypersetup{
  colorlinks=true,
  linkcolor=blue!55!black,
  citecolor=blue!55!black,
  urlcolor=blue!55!black
}

\title{Do Generative Priors Align with Human\\Naturalness Perception?}

\author{Taiki Fukiage\\
Communication Science Laboratories, NTT, Inc.\\
\texttt{t.fukiage@gmail.com}
}

\iclrfinalcopy

\begin{document}

\maketitle
\lhead{Preprint} %

\begin{abstract}
Visual generative models are trained to capture the probability distributions of natural images, yet whether their native priors reflect the regularities governing human perception of image naturalness remains an open question. Here, we probe these priors through native prediction errors across 25 open image and video generators. Because raw single-image losses are dominated by scene content and visual complexity, we evaluate directional loss differences using content-preserving, paired relational interventions that selectively disrupt facial configurations or physical illumination consistency while limiting changes in low-level image statistics. Across both domains, these loss differences reproduce human-like selective sensitivities and tolerances, capturing the classic Thatcher effect on faces and shape-dependent responses to illumination inconsistencies. Notably, these loss differences reliably track continuous gradations of human naturalness judgments across individual stimulus pairs (peaking at $r = .84$ on faces and $.64$ on physical scenes) and retain unique human-aligned signals even after controlling for feature distances from frozen vision encoders and standard image quality metrics. We also find that while overall sensitivity to these violations broadly covaries with human alignment across models, the two systematically decouple along denoising schedules, with alignment peaking earlier than sensitivity, revealing that human-like naturalness judgments dissociate from generic violation detection. Together, these findings demonstrate that learning visual distributions yields generative loss landscapes that capture distinct aspects of human naturalness perception.
\end{abstract}

\section{Introduction}\label{introduction}

Human perception of image naturalness draws heavily on sensitivity to the statistical regularities of natural visual environments: cognitive theories formalize this as inference against internal distributional priors acquired through visual experience \citep{Yuille2006-vision}. Just as human vision relies on approximate statistical expectations rather than exact physical simulation, visual generative models produce strikingly realistic images from learned image distributions while frequently violating strict physical and geometric constraints \citep{Farid2022-lighting,Sarkar2023-es}. This parallel raises the possibility that generative priors, learned from visual distributions rather than explicit physical simulation, naturally capture the selective sensitivities and tolerances of human vision.

\begin{figure}[t]
\centering
\includegraphics[width=\linewidth]{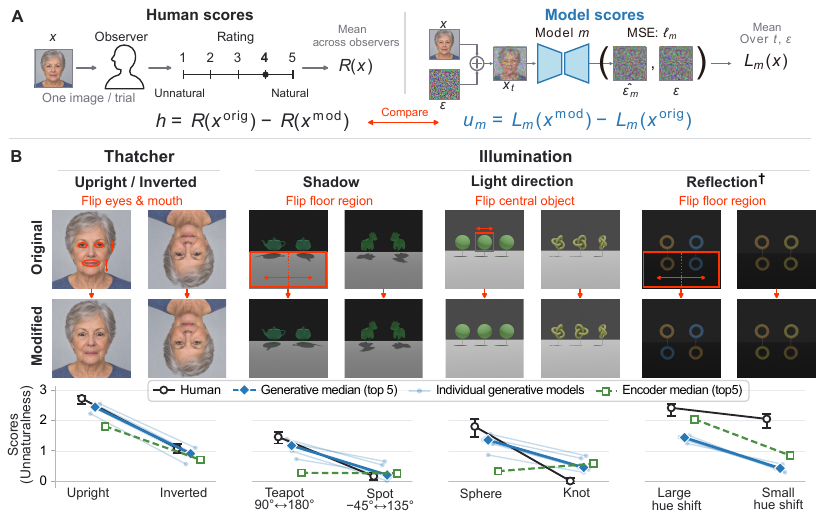}
\caption{\textbf{Overview of experimental paradigm and paired response profiles.}
(A) Measurement pipelines. Left: human naturalness ratings are averaged across observers to define the paired perceptual effect $h$. Right: native prediction errors from frozen models are averaged over matched timesteps and noise to obtain image losses and the paired directional loss difference $u_m$ (schematic shows a noise-predicting diffusion model; prediction targets vary across architectures).
(B) Controlled stimulus pairs and representative response profiles. Content-preserving relational interventions flip local elements (eyes/mouth, floor, or central objects; red boxes), leaving overall scene composition intact.
\textbf{Thatcher}: both humans and generative models penalize facial distortions far more strongly in upright than in inverted faces.
\textbf{Illumination}: factor-level contrasts across cast shadows, light directions, and reflections; generative models mirror human sensitivities to shadows and lighting. $^\dagger$Reflection color is a representative divergence, where humans remain sensitive to small hue mismatches while generative responses decrease.
Markers indicate human means with 95\% bootstrap intervals (Appendix~\ref{detailed-statistical-methods}), top-five generative models (median and individuals), and top-five vision encoders (median). Complete profiles: Appendix~\ref{illumination-factor-profiles}. Face images depict a synthetic person for illustration and were not used in experiments.}
\label{fig:stimuli-reference-conditioning}
\end{figure}

Prior studies utilizing generative models for perceptual assessment have largely treated them as feature extractors, training supervised regressors on internal activations \citep{Fu2024-gk,Raviv2024-ce} or engineering tuning-free attention-based metrics \citep{Song2025-diffsim}. To test whether generative models naturally internalize human-aligned perceptual regularities without relying on such task-specific readouts, we turn directly to their native prediction objectives. In diffusion and flow architectures, denoising and velocity losses relate to score matching and variational bounds on data log-likelihood \citep{Ho2020-tw,Kingma2023-zc,Esser2024-sd3}, providing an empirical surrogate for negative log-likelihood that reflects an image's plausibility under the learned distribution \citep{Clark2023-bt}. We therefore evaluate whether these native prediction errors directly track human naturalness ratings in a zero-shot manner, probing the perceptual sensitivities captured by native generative prediction objectives.

Evaluating native loss directly, however, faces a known obstacle: high-dimensional density estimates and unconditioned prediction errors are sensitive to low-level image complexity, background statistics, and data geometry, which can decouple density from perceived plausibility \citep{Nalisnick2018-jh,Serra2020-input,Ren2019-likelihoodratio,Kamkari2024-pz}. Consistent with these concerns, in our control evaluation on a standard distortion benchmark (CSIQ \citep{Larson2010-xm}), native prediction errors fail to track human quality judgments across distortion types, and removing visual elements like cast shadows can lower loss even though the scene becomes physically unnatural (Appendix~\ref{diagnostic-controls-generic-distortions-element-deletion}). To separate relational plausibility from low-level statistical shifts, we adopt psychophysical paradigms based on content-preserving relational interventions \citep{Thompson1980-thatcher,Ostrovsky2005-ev,Nightingale2019-ef}. 
Instead of adding or deleting objects, our manipulations alter spatial or physical configurations by reflecting existing image elements, which leaves coarse image statistics largely intact. The paired directional loss difference ($L_{\mathrm{modified}} - L_{\mathrm{original}}$) then controls for shared scene composition, focusing evaluation on whether the model penalizes the relational violations that reduce human-perceived naturalness.

We evaluate this approach across two controlled domains: the Thatcher face task (feature inversions in upright and inverted faces) and the Illumination task (shadow, reflection, and light-direction interventions in rendered scenes; Figure~\ref{fig:stimuli-reference-conditioning}). In representative response profiles, paired generative loss differences capture characteristic human perceptual patterns, including the classic Thatcher effect and shape-dependent shadow sensitivities, while also revealing specific divergences (Figure~\ref{fig:stimuli-reference-conditioning}B). To test whether these correspondences reflect systematic perceptual alignment, we evaluate 25 open image and video generators on individual stimulus pairs, comparing native loss differences against human naturalness ratings, 12 frozen vision and multimodal encoders, and ten standard image quality metrics.

Our investigation yields three primary findings. First, zero-shot directional loss differences reproduce human-like selective sensitivities and tolerances across representative physical and face conditions. Across individual items, these loss differences reliably track continuous human ratings (peaking at $r = .84$ on faces and $.64$ on physical scenes) and concentrate spatially within manipulated scene regions. Second, top generative models achieve higher human alignment than standard image quality metrics and frozen vision encoders, and retain non-redundant perceptual signals in bidirectional partial correlation analyses. Third, while coarse violation sensitivity broadly covaries with perceptual alignment across models, the two systematically decouple along denoising schedules, with alignment peaking earlier than sensitivity. 
Together, our work provides empirical evidence for a functional convergence between visual generative priors and human perception, showing that modern generative loss landscapes capture key aspects of the selective sensitivities and tolerances of human naturalness judgments.

\section{Related Work}\label{related-work}

\subsection{Generative Models as Perceptual Priors}

Evaluating departures from natural image statistics is a long-standing approach in perceptual assessment. Classical no-reference image quality assessment (NR-IQA) measures distortions as deviations from parametric models of natural scene statistics \citep{Mittal2012-brisque,Mittal2013-niqe,Saad2012-bliinds}. From a Bayesian perspective, visual quality assessment formalizes this principle as statistical inference against learned priors of natural visual environments \citep{Duanmu2021-of}, echoing cognitive theories of biological vision as Bayesian inference or inverse graphics \citep{Yuille2006-vision,Yildirim2020-efficient}. Modern deep generative models scale this principle by fitting high-dimensional distributions directly on large image corpora. However, raw density estimates and unconditioned prediction errors do not directly track perceptual quality. Likelihood-based deep models frequently assign higher likelihoods to out-of-distribution inputs than to in-distribution training data \citep{Nalisnick2018-jh,Kirichenko2020-im}, reflecting sensitivity to input complexity \citep{Serra2020-input}, background statistics \citep{Ren2019-likelihoodratio}, and manifold geometry \citep{Kamkari2024-pz}.

To bypass these density confounds, recent approaches treat generative models as feature extractors, using supervised regressors on internal activations \citep{Fu2024-gk,Raviv2024-ce} or tuning-free attention correspondences \citep{Song2025-diffsim}. While effective for metric engineering, these feature-derived methods do not evaluate whether the model's learned distribution itself aligns with human naturalness perception. Rather than engineering an applied IQA metric, our goal is a scientific inquiry: testing whether human perceptual priors are reflected directly within the native loss landscapes of generative models, using content-preserving relational interventions to probe prediction errors without intermediate readouts.

\subsection{Physical World Models vs.\ Perceptual Naturalness}

Several recent benchmarks evaluate whether generative models learn physical and causal regularities as predictive world models. While VideoPhy evaluates physical commonsense in video generation \citep{Bansal2025-videophy} and Physics-IQ tests real-world physical continuation \citep{Motamed2025-physicsiq}, LikePhys \citep{Yuan2026-likephys} and YoCausal \citep{Xie2026-yocausal} use paired denoising-loss contrasts to test model preferences for physically valid or temporally plausible sequences, verifying their benchmarks against aggregate human model rankings. However, adherence to objective physical laws does not directly imply alignment with human perceptual naturalness. Psychophysical studies demonstrate that human observers frequently fail to detect or tolerate substantial physical inconsistencies in lighting, shadows, and reflections \citep{Mamassian2004-ou,Ostrovsky2005-ev,Cavanagh2005-artist,Wilder2019-ae,Nightingale2019-ef}. A model that strictly penalizes all physical departures may therefore diverge from human naturalness judgments. Rather than evaluating whether models act as strict physical simulators, our Illumination task examines whether generative loss landscapes mirror the selective tolerances that characterize human perception, testing item-level correspondence with continuous human naturalness ratings under controlled physical interventions.

\subsection{Visual Illusions in Deep Networks}

Visual illusions and perceptual biases provide a classic benchmark for testing whether artificial neural networks reproduce biological visual phenomena \citep{GomezVilla2019-illusions,Jacob2021-wr,Wang2024-illusionvqa}. Prior computational studies have investigated these effects across several paradigms: convolutional filters adapt to natural statistics to reproduce low-level brightness illusions \citep{GomezVilla2019-illusions}, networks trained on edge-based image reconstruction replicate surface filling-in illusions \citep{Saha2026-edge}, and diffusion models reveal human-like brightness shifts along their DDIM inversion trajectories \citep{GomezVilla2025-art}. In higher-level perception, configural phenomena such as the Thatcher illusion have been investigated in feedforward neural networks \citep{Jacob2021-wr}, while structured inverse graphics models have captured other face-perception illusions such as the hollow-face effect \citep{Yildirim2020-efficient}. In addition, multimodal benchmarks probe illusions via linguistic question answering \citep{Wang2024-illusionvqa}. Extending this line of inquiry, we test whether characteristic human perceptual biases are expressed directly within the native loss landscapes of visual generative models.

\section{Methods}\label{methods}

\subsection{Stimulus Design and Psychophysical Experiments}
\label{human-judgments-and-stimuli}

\textbf{Probing human naturalness perception across two domains.}
To investigate whether generative priors reflect human naturalness perception, we examine two distinct visual domains: face configurations and physical scene illumination. In face perception, observers exhibit pronounced orientation dependence: in the Thatcher illusion, inverted eyes and mouth appear grotesque in an upright face, but this distortion becomes difficult to detect when the entire face is inverted \citep{Thompson1980-thatcher,de-Haas2018-en}. Testing Thatcherized faces evaluates whether generative models detect these localized feature inversions and whether their sensitivity exhibits orientation dependence. In physical scenes, observers can fail to detect substantial inconsistencies in illumination, shadows, and reflections \citep{Mamassian2004-ou,Ostrovsky2005-ev,Cavanagh2005-artist,Wilder2019-ae,Nightingale2019-ef}. Testing illumination interventions determines whether models identify physical inconsistencies at all, and whether their responses reflect physical mechanics strictly or follow human perceptual tolerances and naturalness ratings.

\textbf{Content-preserving relational interventions.}
To probe relational regularities while limiting low-level statistical shifts, we design two tasks using \emph{content-preserving relational interventions}. Existing image content is rearranged while depicted objects remain in place, avoiding object deletion or insertion. This design largely preserves color distributions and edge content, reducing the density confounds observed in evaluation of raw native losses. The Thatcher task contains 140 public-domain face scenes from FFHQ~\citep{Karras2019-stylegan}, modified by in-place landmark-guided vertical reflection of the eye and mouth regions~\citep{Alimohammadi2020-thatcher} using dlib-based facial landmark detection~\citep{King2009-dlib,Kazemi2014-face} and presented upright and inverted (280 original--modified pairs; Appendix~\ref{stimulus-construction}). The Illumination task contains 96 scenes rendered in Kubric~\citep{Greff2022-kubric} each for cast-shadow, reflection, and light-direction interventions (288 pairs): for shadow and reflection conditions, the rendered floor region was mirrored to exchange shadows or reflections while objects remained fixed; for the light-direction condition, the central object was horizontally mirrored while the flanking objects and floor, including any cast shadows, remained unchanged. To produce graded item difficulty, we varied object geometry, orientation or color, height, material, and lighting parameters across scenes (Appendix~\ref{stimulus-construction}).

\textbf{Human psychophysical experiments.}
Human observers rated the naturalness of each image on a 5-point scale (1: \emph{very unnatural} to 5: \emph{natural}) in online psychophysical experiments. Each trial presented a single image (Figure~\ref{fig:stimuli-reference-conditioning}A). In the Thatcher task, each observer viewed upright and inverted faces from different identity sets, whereas illumination conditions were tested between subjects. In both tasks, original and modified versions of each scene were evaluated by the same observer across separate, randomized trials (Appendix~\ref{human-experiment-details}). Following response-consistency screening (143 observers retained for Thatcher, 105 for Illumination), each item received an average of 35.8 ratings in Thatcher and 26.2 in Illumination. Item-mean ratings \(R(x)\) define two measures: single-image unnaturalness \(a(x)=5-R(x)\) and the paired unnaturalness effect \(h_i=R(x_i^{\mathrm{orig}})-R(x_i^{\mathrm{mod}})\), which quantifies the perceived unnaturalness introduced by each modification (\(i\) indexes stimulus pairs). Only item-level aggregates enter subsequent model analyses.

\subsection{Generative Models and Native Loss Probing}\label{model-suite}

We evaluate 25 open generative models (16 image, 9 video) across 11 families: (i) pixel-space image models (\emph{JiT}, \emph{PixelGen}, \emph{HiDream-O1-Image})~\citep{Li2025-jit,Ma2026-pixelgen,Cai2026-hidreamo1}; (ii) latent-space image models (\emph{Stable Diffusion v1.5/XL/3, FLUX.1/2, Qwen-Image})~\citep{Rombach2022-ldm,Podell2023-sdxl,Esser2024-sd3,BlackForestLabs2024-flux,BlackForestLabs2025-flux2,Wu2025-qwenimage}; and (iii) spatiotemporal video models (\emph{CogVideoX, LTX-Video, Wan2.1/2.2, HunyuanVideo})~\citep{Yang2024-cogvideox,HaCohen2024-ltxvideo,WanTeam2025-wan,Kong2024-hunyuanvideo}. Static images are repeated as 17-frame clips, with 21 frames for CogVideoX1.5 to satisfy its temporal patching constraint. Text-conditioned models use the neutral or empty-prompt single-pass convention without classifier-free guidance; a control evaluation on a text-conditioned subset confirms that adding descriptive prompts does not alter our findings (Appendix~\ref{conditioning-and-objective-controls}). Model checkpoints, schedules, and evaluation details appear in Appendix~\ref{model-and-runtime-manifest}.

Let \(\ell_m(x,t,\epsilon)\) denote model \(m\)'s native prediction loss (e.g., squared denoising error or flow-matching velocity loss) for input \(x\), timestep \(t\), and noise \(\epsilon \sim \mathcal{N}(0, I)\). We aggregate over stored timesteps \(T_m\) (100 for image, 50 for video) and noise realizations \(E_m\) (20 draws):
\[
L_m(x)=\frac{1}{|T_m||E_m|}\sum_{t\in T_m}\sum_{\epsilon\in E_m}\ell_m(x,t,\epsilon).
\]
We use the average native prediction loss \(L_m(x)\) as an empirical surrogate for negative log-likelihood. Because loss formulations, target spaces (pixel vs.\ latent), and latent dimensions differ across architectures, raw loss magnitudes \(L_m(x)\) are not comparable across models. We evaluate two scores corresponding to human judgments: (i) the \emph{unconditioned single-image loss} \(L_m(x)\), compared against single-image unnaturalness \(a(x)\) as a baseline control, and (ii) the \emph{controlled paired loss change} \(u_{m,i}=L_m(x_i^{\mathrm{mod}})-L_m(x_i^{\mathrm{orig}})\), evaluated against the paired human unnaturalness effect \(h_i\). A positive value (\(u_{m,i} > 0\)) indicates that the relational violation increased native denoising loss, consistent with human ratings where modifications reduce perceived naturalness (\(h_i > 0\)). Paired subtraction controls for shared scene content, evaluating model sensitivity to targeted relational violations.

\subsection{Perceptual Evaluation Metrics}
\label{outcome-measures}
\label{frozen-reference-based-baselines}

\textbf{Standardized unnaturalness and condition profiles (Section~\ref{zero-shot-alignment-and-spatial-localization}).}
Because raw loss scales differ across models and share no common units with human perceptual ratings, raw paired differences ($u_{m,i} = L_m(x_i^{\mathrm{mod}}) - L_m(x_i^{\mathrm{orig}})$ for models, $h_i$ for humans) cannot be directly compared across models or against human judgments. To place models and human ratings on a common scale for condition- and factor-level comparisons, we define the \textbf{standardized unnaturalness score} for each stimulus pair $i$ as the paired difference divided by the task-wide sample standard deviation: \(\tilde{u}_{m,i} = u_{m,i}/s_{m,\mathrm{task}}\) and \(\tilde{h}_i = h_i/s_{h,\mathrm{task}}\), where $s_{m,\mathrm{task}}$ and $s_{h,\mathrm{task}}$ are computed across all $N$ pairs within the task ($N=280$ for Thatcher, $N=288$ for Illumination, using $N-1$ degrees of freedom). Averaging $\tilde{u}_{m,i}$ across specific stimulus factor levels yields the paired response profiles that probe selective sensitivities and tolerances (e.g., face orientation, object geometry, or lighting shifts; Figure~\ref{fig:stimuli-reference-conditioning}B). Human sensitivity profiles are defined analogously from $\tilde{h}_i$. To examine where loss penalties arise, we compute pixel-level paired prediction error maps (Figure~\ref{fig:spatial-loss-maps}).

\textbf{Item-level human alignment (Section~\ref{score-pattern-results}).}
To determine whether this perceptual correspondence generalizes beyond condition-level averages to continuous scene-by-scene judgments, human alignment measures whether graded model penalties track continuous human naturalness ratings across individual items, quantified by the item-level Pearson correlation $r_m = \operatorname{corr}_i(u_{m,i}, h_i)$ within each task. Furthermore, to test whether this correspondence reflects unique perceptual structure beyond standard visual features, we evaluate generative models alongside standard quality metrics and frozen vision encoders, using bidirectional partial correlations to isolate non-redundant signals (Section~\ref{baseline-comparators}).

\textbf{Coupling and dissociation of sensitivity and alignment (Section~\ref{sensitivity-alignment-and-noise-scale-dynamics}).}
Finally, to test whether violation detection and perceptual naturalness represent unified or dissociable properties, we contrast human alignment against coarse \textbf{violation sensitivity}. Task-level sensitivity $S_m$ is defined as the mean standardized score across all pairs in the task ($S_m = N^{-1}\sum_i \tilde{u}_{m,i}$), and condition-level sensitivity $S_{m,c}$ averages over the pairs $I_c$ within condition $c$ ($S_{m,c} = |I_c|^{-1}\sum_{i\in I_c}\tilde{u}_{m,i}$). We then evaluate how this coarse violation sensitivity relates to fine-grained human alignment across model architectures and along denoising schedules.

\textbf{Uncertainty estimation.}
Confidence intervals for human-dependent estimates use 10,000 participant-and-scene bootstrap resamples, whereas those for generative-model sensitivities use scene resampling alone. For cross-model sensitivity--alignment correlations, we additionally perform cluster bootstrap resampling over model families (Appendix~\ref{detailed-statistical-methods}). To verify that associations between sensitivity and alignment are not driven by shared Monte Carlo sampling error, we estimated the two metrics from independent noise draws and observed closely matching results (Appendix~\ref{control-for-shared-sampling-noise}).

\subsection{Baseline Models}
\label{baseline-comparators}
\label{score-pattern-methods}

We compare generative scores with a prespecified pool of 12 frozen image encoders (24M--8.1B parameters) and two five-metric IQA batteries (Appendix~\ref{comparator-baseline-details}). The encoder panel includes ResNet-50~\citep{He2016-resnet}, CLIP~\citep{Radford2021-clip}, DINOv2/v3~\citep{Oquab2023-dinov2,Simeoni2025-dinov3}, SigLIP~2~\citep{Tschannen2025-siglip2}, PE-Core~\citep{Bolya2025-pe}, Qwen3-VL-Embedding~\citep{Li2026-qwen3vlembedding}, and Jina CLIP v2~\citep{Koukounas2024-jinaclipv2}. Encoder scores use cosine distances between global vectors, or the mean distance between corresponding spatial patches for DINOv2/v3, capturing representational divergence across paired images. We evaluate representations across all intermediate blocks alongside each model's default final output, extracting features via fixed architecture-specific pooling (Appendix~\ref{comparator-baseline-details}). For each encoder and task, we select the candidate with the highest Pearson correlation with human judgments. While this post-hoc layer selection deliberately favors the encoder baselines over zero-shot generative scores, fixed feature distances measure the magnitude of change rather than the direction of degradation. Thus, they serve as a diagnostic baseline rather than an upper bound on encoder representations. The IQA batteries comprise FR-IQA (PSNR, SSIM, VIF, LPIPS, DISTS; \citealp{Wang2004-ssim,Sheikh2006-vif,Zhang2018-jh,Ding2022-dists}) and NR-IQA (NIQE, BRISQUE, HyperIQA, MUSIQ, LIQE; \citealp{Mittal2013-niqe,Mittal2012-brisque,Su2020-hyperiqa,Ke2021-musiq,Zhang2023-liqe}).

To test whether generative loss differences capture human-aligned naturalness judgments that provide non-redundant information relative to standard image quality metrics and high-level visual encoders, we evaluate human alignment and bidirectional partial correlations against top-performing baselines. After selecting each encoder's best-performing layer, we assess human alignment across candidate models, retaining the top five Generative models and top five Encoders alongside all five metrics from each prespecified IQA battery. For every selected generative--encoder pair $(G, X)$, we compute $r(h, G \mid X)$ and $r(h, X \mid G)$, summarizing each $5 \times 5$ comparison by its medians. Top-five models are selected over each entire task and retained, together with the selected encoder layers, across its conditions (Appendix~\ref{condition-specific-partial-correlation}). To account for selection uncertainty, confidence intervals are estimated via 10,000 participant-and-scene bootstrap resamples, repeating layer selection followed by model selection within each draw (Appendix~\ref{selection-aware-partial-correlation-protocol}).

\section{Results}\label{results}

\subsection{Qualitative Validation of Generative Losses}
\label{zero-shot-alignment-and-spatial-localization}

We first examined standardized unnaturalness profiles across representative stimulus manipulations (Figure~\ref{fig:stimuli-reference-conditioning}B). In the Thatcher task, humans and top generative models strongly penalized upright facial feature inversions, with sensitivity dropping sharply under inversion (Figure~\ref{fig:stimuli-reference-conditioning}B; top-five generative median: $2.44$ upright vs.\ $0.91$ inverted; across all 25 models, median sensitivity was $2.24$ vs.\ $0.84$; humans: $2.73$ vs.\ $1.09$; Appendix~\ref{model-wise-condition-sensitivity-breakdown}). In the Illumination task, generative model losses tracked human unnaturalness across scene geometry: for cast shadows, both humans and models showed pronounced sensitivity to shadow rotations for objects with distinctive directional landmarks (e.g., teapot), but remained tolerant for shapes lacking prominent in-plane asymmetries (e.g., Spot; Figure~\ref{fig:stimuli-reference-conditioning}B). Crucially, top vision encoders failed to capture this selective sensitivity, showing uniformly low responses across both objects ($0.28$ vs.\ $0.27$), and similarly missed human sensitivity to light-direction changes on simple spheres (Figure~\ref{fig:stimuli-reference-conditioning}B). Conversely, human and model responses diverged on reflection color: whereas humans strongly penalized subtle hue mismatches, generative models showed relatively weak sensitivity.

\begin{figure}[t]
\noindent\begin{minipage}[c]{0.56\linewidth}
\centering
\includegraphics[width=\linewidth]{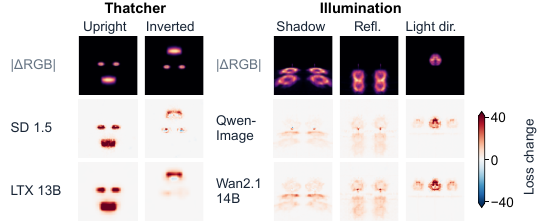}
\end{minipage}\hfill
\begin{minipage}[c]{0.41\linewidth}
\caption{\textbf{Spatial loss changes.} Mean paired loss maps for two high-alignment models per task. Top-row pixel differences identify the modified regions. Complete 25-model maps: Appendix~\ref{mean-spatial-loss-maps}.}
\label{fig:spatial-loss-maps}
\end{minipage}
\end{figure}

Spatial loss-difference maps reveal where violation penalties arise across the scene (Figure~\ref{fig:spatial-loss-maps}). For upright Thatcher faces, loss increases concentrated around the inverted eyes and mouth, with marked attenuation under face inversion. For illumination interventions, loss increases were not confined to manipulated regions, extending from floor surfaces to objects directly above altered shadows or reflections, and from the central object to unmodified flanking objects under light-direction shifts. These maps indicate that generative models evaluate contextual relationships across the scene rather than isolated local modifications.

\subsection{Item-Level Human Alignment}
\label{score-pattern-results}

\begin{figure}[t]
\centering
\includegraphics[width=\linewidth]{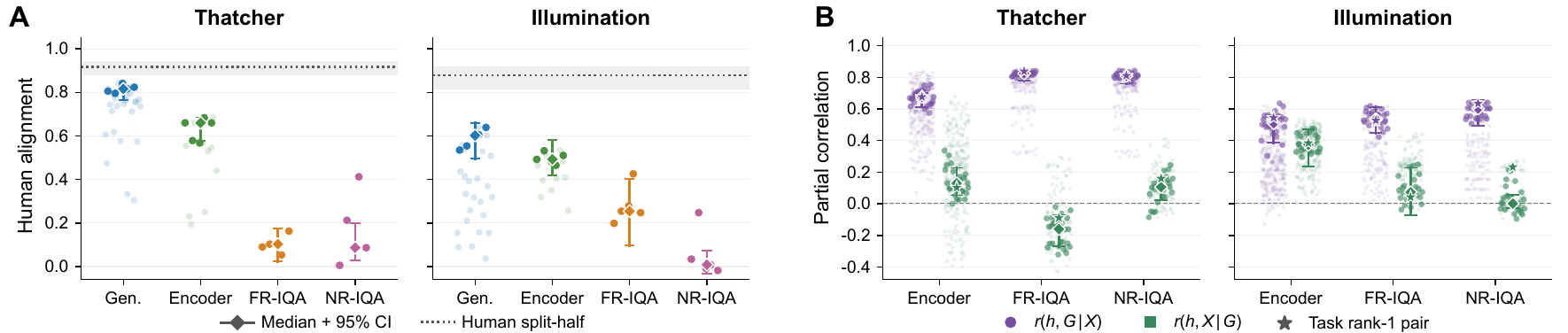}
\caption{\textbf{Item-level human alignment and comparison with baselines.}
(A) Model--human Pearson correlations on individual stimulus pairs. Saturated points mark top-five Generative and Encoder models and all IQA metrics; diamonds show medians. Dotted lines and bands show mean human split-half reliability and its 95\% interval.
(B) Bidirectional partial correlations ($r(h,G\mid X)$, purple circles; $r(h,X\mid G)$, green circles). Faint points show all pairs; saturated points show the selected 5 $\times$ 5 model pairs; stars mark the rank-1 pair. Large markers and bars in all panels show medians and 95\% bootstrap CIs (Section~\ref{score-pattern-methods}, Appendix~\ref{selection-aware-partial-correlation-protocol}).}
\label{fig:comparator-alignment-and-partial-correlation}
\end{figure}

\textbf{Item-level agreement across generative models.}
We next tested whether these selective sensitivities generalize to continuous, item-level agreement across the full 25-model benchmark. Across individual stimulus pairs, directional loss changes ($u_{m,i} = L_m(x_i^{\mathrm{mod}}) - L_m(x_i^{\mathrm{orig}})$) reliably tracked continuous human unnaturalness ratings ($h_i$), with median Pearson correlations of $.747$ on Thatcher (peaking at $.841$) and $.334$ on Illumination (peaking at $.639$; Figure~\ref{fig:comparator-alignment-and-partial-correlation}A). This alignment was not driven merely by coarse mean differences across conditions: after condition centering (subtracting condition-specific means from both model losses and human ratings to isolate item-level rankings within each condition), correlations remained positive for all 25 models on both tasks (medians of $.411$ and $.271$; peaking at $.534$ and $.619$; condition breakdowns in Appendix~\ref{condition-specific-sensitivity-alignment}). In sharp contrast, unconditioned single-image losses ($L_m(x)$) showed little correlation with human ratings (25-model medians of $-.20$ to $.24$; Appendix~\ref{additional-score-analyses}), confirming that paired differencing is essential to separate relational naturalness from shared scene composition.

Notably, human alignment was observed across diverse generative formulations (diffusion versus flow), modeling spaces (pixel versus latent), and modalities (image versus video), with substantial within-category variation (model-wise breakdowns across conditions in Figure~\ref{fig:condition-sensitivity-heatmap-supplement}; model specifications in Table~\ref{tab:generative-model-suite}). Furthermore, high alignment emerged in models whose training recipes include no explicit preference optimization (e.g., RLHF or DPO). On Thatcher, class-conditional JiT models \citep{Li2025-jit} trained solely on ImageNet exhibited alignment that scaled systematically with model capacity (reaching $r = .577$ for JiT-H/32), alongside strong agreement in SD~v1.5 ($r = .825$). On Illumination, Wan2.1 14B \citep{WanTeam2025-wan} achieved the highest overall correlation ($r = .639$, remaining $.619$ after condition centering), with solid alignment also sustained in video models such as CogVideoX \citep{Yang2024-cogvideox}. These findings indicate that perceptual alignment with human naturalness judgments emerges directly from learning visual data distributions.

\textbf{Comparison with visual encoder and IQA baselines.} Across both tasks, top generative models achieved higher human alignment than all evaluated baselines (Figure~\ref{fig:comparator-alignment-and-partial-correlation}A), even after selecting each encoder's best-performing layer (top-five median $r = .816$ vs.\ $.660$ on Thatcher; $.602$ vs.\ $.492$ on Illumination). In contrast, standard image quality metrics showed substantially weaker correspondence: NR-IQA correlations remained near zero throughout ($.087$ on Thatcher, $.009$ on Illumination), arguing against generic degradation or manipulation artifacts, while weak FR-IQA results ($.103$ and $.254$) indicate that alignment is not driven merely by low-level change magnitude.

\textbf{Bidirectional partial correlations with baselines.} Generative scores also retained robust positive residual associations after controlling for baseline metrics in bidirectional partial correlation analyses (Figure~\ref{fig:comparator-alignment-and-partial-correlation}B). Controlling for standard IQA metrics had little effect on generative alignment (medians of $r(h,G\mid\mathrm{IQA}) = .809$--$.825$ on Thatcher and $.535$--$.602$ on Illumination), while IQA residual associations remained near or below zero ($-.160$ to $.106$), confirming that generative signals do not simply reflect low-level degradation or distortion magnitude. When evaluated alongside frozen vision encoders, generative scores likewise preserved positive residual associations. On Thatcher, generative scores retained strong incremental explanatory power (median partial correlation of $.676$, 95\% CI $[.612,.701]$ vs.\ $.130$, $[.054,.228]$ for encoders), whereas on Illumination, both representations retained substantial non-redundant associations ($.510$ for generative models vs.\ $.375$ for encoders), demonstrating complementary perceptual structure. Broken down by condition, generative scores contributed distinct variance for upright faces, cast shadows, and light direction, while encoders retained larger associations for surface reflections (Figure~\ref{fig:condition-specific-comparator-partial-correlations}, Appendix~\ref{condition-specific-partial-correlation}). Together, these results indicate that native generative loss landscapes capture human-aligned perceptual regularities that are distinct from standard quality metrics and complementary to discriminative visual encoders.

\subsection{Coupling and Dissociation of Sensitivity and Alignment}
\label{sensitivity-alignment-and-noise-scale-dynamics}

\begin{figure}[t]
\centering
\includegraphics[width=\linewidth]{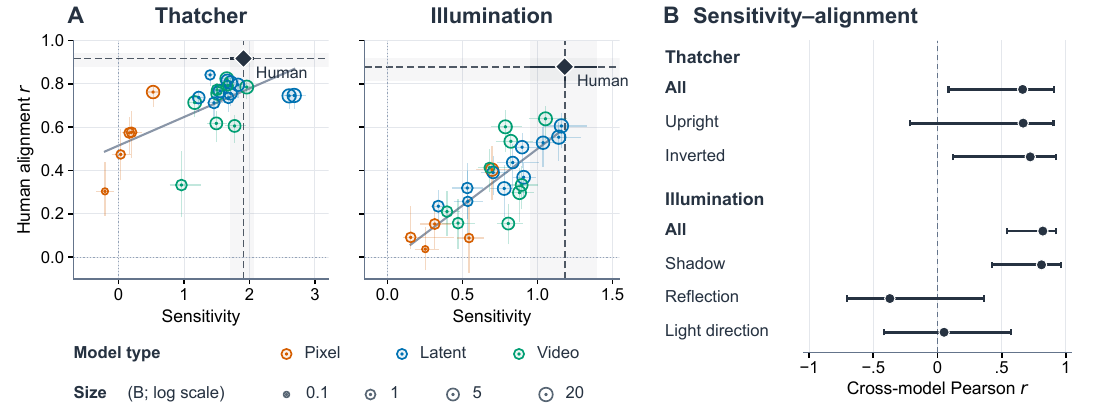}
\caption{\textbf{Cross-model association between sensitivity and human alignment.}
(A) Violation sensitivity versus human alignment ($r$) for 25 generative models on Thatcher and Illumination. Solid lines show linear fits; black diamonds and dashed crosshairs mark human sensitivity and split-half reliability. Marker colors indicate model types (pixel, latent, video); sizes scale with parameter count (bottom legend).
(B) Cross-model Pearson correlations ($r$) between sensitivity and alignment for each task as a whole (All) and within individual conditions. Error bars denote 95\% bootstrap CIs.}
\label{fig:cross-model-sensitivity-alignment}
\end{figure}

\textbf{Cross-model association.}
To test whether human alignment simply reflects stronger violation detection, we examined whether models with greater violation sensitivity also showed closer human alignment, evaluating this relationship both across each task as a whole and within individual stimulus conditions (Figure~\ref{fig:cross-model-sensitivity-alignment}). Across the full benchmark, violation sensitivity exhibited an overall positive association with item-level human alignment (Figure~\ref{fig:cross-model-sensitivity-alignment}A; Thatcher: $r = .663$, 95\% CI $[.086,.905]$; Illumination: $r = .821$, $[.542,.922]$), with larger models generally clustering toward the upper right (Appendix~\ref{model-size-diagnostic}, Figure~\ref{fig:model-size-sensitivity-alignment}). Within individual conditions (Figure~\ref{fig:cross-model-sensitivity-alignment}B), positive associations were robust for cast shadows ($r = .810$, 95\% CI $[.424,.960]$) and inverted faces ($r = .723$, $[.118,.923]$). For upright faces, the point estimate remained positive but carried wide estimation uncertainty ($r = .667$, $[-.215,.904]$), whereas associations were near zero for light direction ($r = .051$, $[-.417,.570]$) and negative for surface reflections ($r = -.371$, $[-.704,.364]$; Appendix~\ref{condition-specific-sensitivity-alignment}). Thus, while sensitivity and alignment covary across each task as a whole, this coupling is not uniform across individual stimulus conditions.

\begin{figure}[t]
\centering
\includegraphics[width=0.90\linewidth]{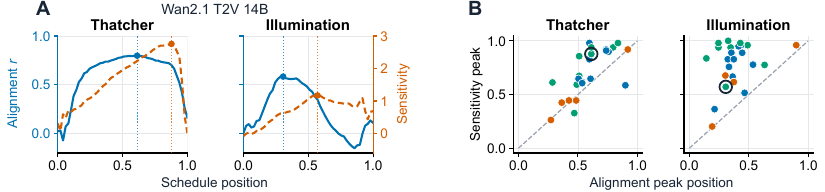}
\caption{\textbf{Dissociation of sensitivity and human alignment along denoising schedules.}
(A) Representative schedule dynamics (Wan2.1 T2V 14B) on Thatcher and Illumination: human alignment $r$ (blue, solid) and sensitivity (orange, dashed) across normalized schedule positions ($0 \to 1$); solid dots mark peak positions.
(B) Peak locations for all 25 models on Thatcher and Illumination across normalized schedules ($[0,1]$); the dashed diagonal marks equal timing. Dots are colored by model category following Figure~\ref{fig:cross-model-sensitivity-alignment}; open black rings identify Wan2.1 T2V 14B.}
\label{fig:timestep-profiles-and-peaks}
\end{figure}

\textbf{Dissociation along denoising schedules.}
Within individual models, violation sensitivity and human alignment systematically peak at different stages of the denoising schedule (Figure~\ref{fig:timestep-profiles-and-peaks}A,B). Across normalized schedules ($[0,1]$), sensitivity peaked later (at cleaner noise levels) than human alignment in 20 of 25 models on Thatcher and in all 25 models on Illumination, with median within-model peak shifts of $0.141$ and $0.444$ (condition breakdowns in Appendix~\ref{timestep-profiles}). This separation is consistent with the coarse-to-fine dynamics of generative diffusion \citep{Sclocchi2024-tv,Tinaz2025,Gorgun2026}: human alignment on relational tasks may peak at intermediate noise levels where global layout and contextual consistency are resolved, whereas later stages emphasize fine details that inflate loss penalties without refining perceptual agreement.

\textbf{Connection to practical generation performance.}
Finally, as an exploratory analysis, we investigated whether loss-based perceptual alignment connects to practical generative capabilities (Appendix~\ref{external-generation-evaluation}). We matched 11 image models to human-preference Elo \citep{ArtificialAnalysis2026-image-methodology} and seven video models to VBench Quality \citep{Huang2024-vbench} (Table~\ref{tab:public-generation-models}). Model-level Spearman correlations between generation benchmarks and human alignment were positive on Illumination for image models ($\rho=.827$) and across both tasks for video models ($\rho=.929$ on Thatcher, $\rho=.821$ on Illumination; Table~\ref{tab:public-generation-sensitivity-alignment}), and these associations remained positive after controlling for violation sensitivity (partial $\rho=.670$ for image models; $.900$ and $.778$ for video models). In contrast, the association on Thatcher images was weak both before and after adjustment ($\rho=.191$ and $.182$), which may reflect ceiling effects as many models approach human-level agreement on faces (Figure~\ref{fig:cross-model-sensitivity-alignment}A). Conversely, reciprocal associations between generation benchmarks and violation sensitivity after controlling for alignment were uniformly weak across all four comparisons ($\rho=-.045$ to $.218$). While based on modest sample sizes, these exploratory associations suggest that loss-based perceptual alignment may reflect visual regularities that support practical generation quality.

\section{Discussion}\label{discussion}

This study evaluated whether the native prediction losses of visual generative models reflect human naturalness perception under controlled relational modifications. Across diverse architectures, zero-shot loss differences track continuous human ratings and retain complementary perceptual structure beyond standard vision encoders and image quality metrics. While coarse violation sensitivity broadly covaries with perceptual alignment across models, sensitivity and fine-grained alignment systematically dissociate along denoising schedules and vary across visual domains. These findings suggest that current generative models internalize structured relational regularities that partially mirror the selective sensitivities and tolerances of human naturalness perception, rather than merely signaling physical violations or generic visual divergence. Beyond characterizing existing architectures, this diagnostic framework provides a foundation to track how generative representations converge toward or diverge from biological perception as models scale.

\textbf{Orientation dependence in face perception.}
Human sensitivity to relational facial distortions deteriorates sharply under inversion, a hallmark effect exemplified by the Thatcher illusion \citep{Thompson1980-thatcher,de-Haas2018-en}.
While this orientation dependence is known to emerge in feedforward networks \citep{Jacob2021-wr} and was observed in our encoder baselines (Figure~\ref{fig:stimuli-reference-conditioning}B), generative models capture human judgments far more closely.
Top generative models markedly outperform even layer-optimized encoders in human alignment on Thatcher, and bidirectional partial correlations confirm that they retain substantial explanatory power that encoder representations cannot account for (Figure~\ref{fig:comparator-alignment-and-partial-correlation}).
Notably, the ImageNet-trained JiT models \citep{Li2025-jit} demonstrate that this orientation penalty can arise without training on a dedicated face dataset, scaling systematically with model capacity (Appendix~\ref{model-wise-condition-sensitivity-breakdown}).
Generative losses also track naturalness variations within upright faces alone (median $r = .493$), echoing findings that generative evidence tracks human facial attractiveness \citep{Lopez2026-elbo}.
These results demonstrate that key characteristics of human face perception can emerge directly from statistical generative modeling of 2D images.

\textbf{Physical scene illumination.}
In our Illumination experiments, human sensitivity to shadow and reflection inconsistencies depended strongly on scene geometry (Figure~\ref{fig:stimuli-reference-conditioning}B): observers readily detected violations for objects with distinctive directional landmarks (e.g., teapot), but tolerated errors for shapes lacking prominent spatial asymmetries (e.g., Spot).
This selective sensitivity aligns with psychophysical accounts where directional landmarks facilitate spatial feature matching \citep{Casati2008-copycat,Farid2010-qz,Nightingale2019-ef,Norman2009-shadows}.
Generative losses systematically tracked this geometric selectivity across shadow and reflection rotations as well as lighting directions (Figure~\ref{fig:stimuli-reference-conditioning}B, Appendix~\ref{illumination-factor-profiles}), suggesting that rather than acting as objective detectors of physical violations, models rely on statistical correspondence cues that mirror human perceptual heuristics.
However, models show a much weaker response to reflection hue shifts than humans do (Figure~\ref{fig:stimuli-reference-conditioning}B).
While humans treat an impossible reflection color as a decisive physical violation, generative losses appear to penalize it only as a modest color difference relative to geometric misalignments.
This divergence suggests that native loss landscapes may internalize relevant relational cues without weighting them in line with human perception.

\textbf{Model scaling and perceptual alignment.}
One might expect that scaling model capacity would steadily resolve such discrepancies, bringing models closer to an ideal observer of physical violations that automatically aligns with human perception.
In exploratory comparisons, whole-task alignment did scale positively with model size (Appendix~\ref{model-size-diagnostic}, Figure~\ref{fig:model-size-sensitivity-alignment}), but this relationship did not uniformly extend to judgments within individual conditions.
Most notably, for surface reflections, larger models grew steadily more sensitive to violations without showing any improvement in human alignment (Figure~\ref{fig:model-size-conditions-illumination}).
This dissociation suggests that heightened sensitivity to visual perturbations does not guarantee alignment with the selective tolerances and weightings of human perception.

\textbf{Limitations.}
Several limitations qualify these conclusions.
First, our psychophysical intervention framework isolates relational regularities by minimizing low-level confounds, but this comes at the expense of scene diversity; extending controlled manipulations to unconstrained natural imagery without introducing low-level artifacts remains challenging.
Second, because we evaluate off-the-shelf models, our comparisons cannot fully isolate core generative objectives from differences in model scale, architecture, or post-training recipes, nor can they disentangle generic visual statistics from human-curated preferences in modern training corpora.
Finally, our analyses linking perceptual alignment to downstream generation benchmarks are observational, leaving open whether directly optimizing for human-aligned relational priors causally improves generation quality.

\clearpage
\subsection*{AI use statement}
Generative AI tools were used at several stages of the research and writing process. They were used to identify potentially relevant prior work and related studies; to discuss and obtain feedback on aspects of research methodology, analysis strategy, and interpretation of results; to assist with data analysis and plotting code; and to draft, revise, and polish portions of the manuscript, including \LaTeX{} formatting. Candidate references and related work identified with AI assistance were checked against the original sources, and AI-assisted code and analyses were inspected and verified by the author.

In addition, an image generation model was used to create the synthetic face image shown in Figures~\ref{fig:stimuli-reference-conditioning} and~\ref{fig:thatcher-stimulus-construction} solely for illustrative purposes to depict the stimulus manipulation. This synthetic image was not used in the human experiments or model analyses.

All research decisions, analyses, interpretations, and claims were made and verified by the author. All AI-assisted text, code, and other outputs were reviewed and edited as appropriate, and the author takes full responsibility for the contents of this work.

\subsection*{Ethics statement}

The human behavioral experiments were approved by the Institutional Review Board (\mbox{Ethics Committee}) of the author's institution. All participants were adults and provided informed consent via an electronic consent form prior to participation. Detailed participant demographics, experimental procedures, and task instructions are provided in Appendix~\ref{human-experiment-details}.

\subsection*{Reproducibility statement}

Code, data, and stimulus materials are available at \url{https://github.com/t-fukiage/generative-priors-naturalness}. The repository provides precomputed model scores, anonymized participant ratings, stimulus images and generation scripts, and reproduction code for model scoring, participant screening, and statistical analyses. Model specifications and experimental and statistical procedures are described in Section~\ref{methods} and Appendices~\ref{stimulus-construction}--\ref{detailed-statistical-methods}.

\subsubsection*{Acknowledgments}
We thank Zitang Sun for his work on preliminary analyses in the early stage of this project and for helpful discussions that contributed to its subsequent development. This work was supported by JSPS KAKENHI Grant Number JP24H00721.

\bibliography{references}

\begin{thebibliography}{81}
\providecommand{\natexlab}[1]{#1}
\providecommand{\url}[1]{\texttt{#1}}
\expandafter\ifx\csname urlstyle\endcsname\relax
  \providecommand{\doi}[1]{doi: #1}\else
  \providecommand{\doi}{doi: \begingroup \urlstyle{rm}\Url}\fi

\bibitem[Alimohammadi(2020)]{Alimohammadi2020-thatcher}
Erfan Alimohammadi.
\newblock {Thatcher Effect Dataset Generator}.
\newblock \url{https://github.com/Erfaniaa/thatcher-effect-dataset-generator},
  2020.
\newblock Open-source software under GNU General Public License v3.0.

\bibitem[{Artificial Analysis}(2026)]{ArtificialAnalysis2026-image-methodology}
{Artificial Analysis}.
\newblock Image generation benchmarking methodology.
\newblock \url{https://artificialanalysis.ai/image/methodology}, 2026.
\newblock Accessed September 12, 2026.

\bibitem[Bansal et~al.(2025)Bansal, Lin, Xie, Zong, Yarom, Bitton, Jiang, Sun,
  Chang, and Grover]{Bansal2025-videophy}
Hritik Bansal, Zongyu Lin, Tianyi Xie, Zeshun Zong, Michal Yarom, Yonatan
  Bitton, Chenfanfu Jiang, Yizhou Sun, Kai-Wei Chang, and Aditya Grover.
\newblock {VideoPhy}: Evaluating physical commonsense for video generation.
\newblock In \emph{International Conference on Learning Representations}, 2025.

\bibitem[{Black Forest Labs}(2024)]{BlackForestLabs2024-flux}
{Black Forest Labs}.
\newblock {FLUX}.
\newblock \url{https://github.com/black-forest-labs/flux}, 2024.

\bibitem[{Black Forest Labs}(2025)]{BlackForestLabs2025-flux2}
{Black Forest Labs}.
\newblock {FLUX.2}: Frontier visual intelligence.
\newblock \url{https://bfl.ai/blog/flux-2}, 2025.

\bibitem[Bolya et~al.(2025)Bolya, Huang, Sun, Cho, Madotto, Wei, Ma, Zhi,
  Rajasegaran, Bangalath, Wang, Monteiro, Xu, Dong, Ravi, Li, Doll{\'a}r, and
  Feichtenhofer]{Bolya2025-pe}
Daniel Bolya, Po-Yao Huang, Peize Sun, Jang~Hyun Cho, Andrea Madotto, Chen Wei,
  Tengyu Ma, Jiale Zhi, Jathushan Rajasegaran, Hanoona Bangalath, Junke Wang,
  Marco Monteiro, Hu~Xu, Shiyu Dong, Nikhila Ravi, Shang-Wen Li, Piotr
  Doll{\'a}r, and Christoph Feichtenhofer.
\newblock Perception encoder: The best visual embeddings are not at the output
  of the network.
\newblock In \emph{Advances in Neural Information Processing Systems},
  volume~38, 2025.

\bibitem[Cai et~al.(2026)]{Cai2026-hidreamo1}
Qi~Cai et~al.
\newblock {HiDream-O1-Image}: A natively unified image generative foundation
  model with pixel-level unified transformer.
\newblock \emph{arXiv preprint arXiv:2605.11061}, 2026.

\bibitem[Casati(2008)]{Casati2008-copycat}
Roberto Casati.
\newblock The copycat solution to the shadow correspondence problem.
\newblock \emph{Perception}, 37\penalty0 (4):\penalty0 495--503, 2008.
\newblock \doi{10.1068/p5588}.

\bibitem[Cavanagh(2005)]{Cavanagh2005-artist}
Patrick Cavanagh.
\newblock The artist as neuroscientist.
\newblock \emph{Nature}, 434\penalty0 (7031):\penalty0 301--307, 2005.

\bibitem[Chen \& Mo(2022)Chen and Mo]{Chen2022-pyiqa}
Chaofeng Chen and Jiadi Mo.
\newblock {IQA-PyTorch}: {PyTorch} toolbox for image quality assessment.
\newblock \url{https://github.com/chaofengc/IQA-PyTorch}, 2022.

\bibitem[Clark \& Jaini(2023)Clark and Jaini]{Clark2023-bt}
Kevin Clark and Priyank Jaini.
\newblock Text-to-image diffusion models are zero-shot classifiers.
\newblock In \emph{Advances in Neural Information Processing Systems},
  volume~36, pp.\  46433--46449, 2023.

\bibitem[de~Haas \& Schwarzkopf(2018)de~Haas and Schwarzkopf]{de-Haas2018-en}
Benjamin de~Haas and D.~Samuel Schwarzkopf.
\newblock Feature-location effects in the {Thatcher} illusion.
\newblock \emph{Journal of Vision}, 18\penalty0 (4):\penalty0 16, 2018.

\bibitem[de~Leeuw et~al.(2023)de~Leeuw, Gilbert, and
  Luchterhandt]{deLeeuw2023-jspsych}
Joshua~R. de~Leeuw, Rebecca~A. Gilbert, and Bj{\"o}rn Luchterhandt.
\newblock {jsPsych}: Enabling an open-source collaborative ecosystem of
  behavioral experiments.
\newblock \emph{Journal of Open Source Software}, 8\penalty0 (85):\penalty0
  5351, 2023.

\bibitem[Ding et~al.(2022)Ding, Ma, Wang, and Simoncelli]{Ding2022-dists}
Keyan Ding, Kede Ma, Shiqi Wang, and Eero~P Simoncelli.
\newblock Image quality assessment: Unifying structure and texture similarity.
\newblock \emph{IEEE Transactions on Pattern Analysis and Machine
  Intelligence}, 44\penalty0 (5):\penalty0 2567--2581, 2022.
\newblock \doi{10.1109/TPAMI.2020.3045810}.

\bibitem[Duanmu et~al.(2021)Duanmu, Liu, Wang, and Wang]{Duanmu2021-of}
Zhengfang Duanmu, Wentao Liu, Zhongling Wang, and Zhou Wang.
\newblock Quantifying visual image quality: A {Bayesian} view.
\newblock \emph{Annual Review of Vision Science}, 7:\penalty0 437--464, 2021.

\bibitem[Esser et~al.(2024)Esser, Kulal, Blattmann, Entezari, M{\"u}ller,
  Saini, Levi, Lorenz, Sauer, Boesel, Podell, Dockhorn, English, and
  Rombach]{Esser2024-sd3}
Patrick Esser, Sumith Kulal, Andreas Blattmann, Rahim Entezari, Jonas
  M{\"u}ller, Harry Saini, Yam Levi, Dominik Lorenz, Axel Sauer, Frederic
  Boesel, Dustin Podell, Tim Dockhorn, Zion English, and Robin Rombach.
\newblock Scaling rectified flow transformers for high-resolution image
  synthesis.
\newblock In \emph{Proceedings of the 41st International Conference on Machine
  Learning}, volume 235 of \emph{Proceedings of Machine Learning Research},
  pp.\  12606--12633, 2024.

\bibitem[Farid(2022)]{Farid2022-lighting}
Hany Farid.
\newblock Lighting (in)consistency of paint by text.
\newblock \emph{arXiv preprint arXiv:2207.13744}, 2022.
\newblock URL \url{https://arxiv.org/abs/2207.13744}.

\bibitem[Farid \& Bravo(2010)Farid and Bravo]{Farid2010-qz}
Hany Farid and Mary~J Bravo.
\newblock Image forensic analyses that elude the human visual system.
\newblock In Nasir~D Memon, Jana Dittmann, Adnan~M Alattar, and Edward~J Delp,
  III (eds.), \emph{Media Forensics and Security II}, volume 7541, pp.\
  52--61. SPIE, 4~February 2010.
\newblock \doi{10.1117/12.837788}.

\bibitem[Fu et~al.(2024)Fu, Wang, Yang, Kot, and Wen]{Fu2024-gk}
Honghao Fu, Yufei Wang, Wenhan Yang, Alex~C. Kot, and Bihan Wen.
\newblock {DP}-{IQA}: Utilizing diffusion prior for blind image quality
  assessment in the wild.
\newblock \emph{arXiv preprint arXiv:2405.19996}, 2024.

\bibitem[G{\'o}mez-Villa et~al.(2020)G{\'o}mez-Villa, Mart{\'i}n,
  Vazquez-Corral, Bertalm{\'i}o, and Malo]{GomezVilla2019-illusions}
Alexandra G{\'o}mez-Villa, Adri{\'a}n Mart{\'i}n, Javier Vazquez-Corral,
  Marcelo Bertalm{\'i}o, and Jes{\'u}s Malo.
\newblock Color illusions also deceive {CNNs} for low-level vision tasks:
  Analysis and implications.
\newblock \emph{Vision Research}, 176:\penalty0 156--174, 2020.

\bibitem[G{\'o}mez-Villa et~al.(2025)G{\'o}mez-Villa, Wang, P{\'a}rraga,
  Twardowski, Malo, Vazquez-Corral, and van~de Weijer]{GomezVilla2025-art}
Alexandra G{\'o}mez-Villa, Kai Wang, C.~Alejandro P{\'a}rraga, Bart{\l}omiej
  Twardowski, Jes{\'u}s Malo, Javier Vazquez-Corral, and Joost van~de Weijer.
\newblock The art of deception: Color visual illusions and diffusion models.
\newblock In \emph{Proceedings of the IEEE/CVF Conference on Computer Vision
  and Pattern Recognition}, pp.\  18642--18652, 2025.

\bibitem[G{\"o}rg{\"u}n et~al.(2026)G{\"o}rg{\"u}n, Sammani, Deligiannis,
  Schiele, and Fischer]{Gorgun2026}
Ada G{\"o}rg{\"u}n, Fawaz Sammani, Nikos Deligiannis, Bernt Schiele, and Jonas
  Fischer.
\newblock Temporal concept dynamics in diffusion models via prompt-conditioned
  interventions.
\newblock In \emph{International Conference on Learning Representations}, 2026.

\bibitem[Greff et~al.(2022)Greff, Belletti, Beyer, Doersch, Du, Duckworth,
  Fleet, Gnanapragasam, Golemo, Herrmann, Kipf, Kundu, Lagun, Laradji, Liu,
  Meyer, Miao, Nowrouzezahrai, Oztireli, Pot, Radwan, Rebain, Sabour, Sajjadi,
  Sela, Sitzmann, Stone, Sun, Vora, Wang, Wu, Yi, Zhong, and
  Tagliasacchi]{Greff2022-kubric}
Klaus Greff, Francois Belletti, Lucas Beyer, Carl Doersch, Yilun Du, Daniel
  Duckworth, David~J. Fleet, Dan Gnanapragasam, Florian Golemo, Charles
  Herrmann, Thomas Kipf, Abhijit Kundu, Dmitry Lagun, Issam Laradji,
  Hsueh-Ti~(Derek) Liu, Henning Meyer, Yishu Miao, Derek Nowrouzezahrai, Cengiz
  Oztireli, Etienne Pot, Noha Radwan, Daniel Rebain, Sara Sabour, Mehdi S.~M.
  Sajjadi, Matan Sela, Vincent Sitzmann, Austin Stone, Deqing Sun, Suhani Vora,
  Ziyu Wang, Tianhao Wu, Kwang~Moo Yi, Fangcheng Zhong, and Andrea
  Tagliasacchi.
\newblock Kubric: A scalable dataset generator.
\newblock In \emph{IEEE/CVF Conference on Computer Vision and Pattern
  Recognition}, pp.\  3749--3761, 2022.

\bibitem[HaCohen et~al.(2024)]{HaCohen2024-ltxvideo}
Yoav HaCohen et~al.
\newblock {LTX-Video}: Realtime video latent diffusion.
\newblock \emph{arXiv preprint arXiv:2501.00103}, 2024.

\bibitem[He et~al.(2016)He, Zhang, Ren, and Sun]{He2016-resnet}
Kaiming He, Xiangyu Zhang, Shaoqing Ren, and Jian Sun.
\newblock Deep residual learning for image recognition.
\newblock In \emph{Proceedings of the IEEE Conference on Computer Vision and
  Pattern Recognition}, pp.\  770--778, 2016.

\bibitem[Ho et~al.(2020)Ho, Jain, and Abbeel]{Ho2020-tw}
Jonathan Ho, Ajay Jain, and Pieter Abbeel.
\newblock Denoising diffusion probabilistic models.
\newblock In \emph{NeurIPS 2020}, 2020.

\bibitem[Huang et~al.(2024)Huang, He, Yu, Zhang, Si, Jiang, Zhang, Wu, Jin,
  Chanpaisit, Wang, Chen, Wang, Lin, Qiao, and Liu]{Huang2024-vbench}
Ziqi Huang, Yinan He, Jiashuo Yu, Fan Zhang, Chenyang Si, Yuming Jiang, Yuanhan
  Zhang, Tianxing Wu, Qingyang Jin, Nattapol Chanpaisit, Yaohui Wang, Xinyuan
  Chen, Limin Wang, Dahua Lin, Yu~Qiao, and Ziwei Liu.
\newblock {VBench}: Comprehensive benchmark suite for video generative models.
\newblock In \emph{Proceedings of the IEEE/CVF Conference on Computer Vision
  and Pattern Recognition}, pp.\  21807--21818, 2024.
\newblock URL
  \url{https://openaccess.thecvf.com/content/CVPR2024/html/Huang_VBench_Comprehensive_Benchmark_Suite_for_Video_Generative_Models_CVPR_2024_paper.html}.

\bibitem[Jacob et~al.(2021)Jacob, Pramod, Katti, and Arun]{Jacob2021-wr}
Georgin Jacob, R~T Pramod, Harish Katti, and S~P Arun.
\newblock Qualitative similarities and differences in visual object
  representations between brains and deep networks.
\newblock \emph{Nature Communications}, 12\penalty0 (1):\penalty0 1872,
  25~March 2021.
\newblock ISSN 2041-1723.
\newblock \doi{10.1038/s41467-021-22078-3}.

\bibitem[Kamkari et~al.(2024)Kamkari, Ross, Cresswell, Caterini, Krishnan, and
  Loaiza-Ganem]{Kamkari2024-pz}
Hamidreza Kamkari, Brendan~Leigh Ross, Jesse~C. Cresswell, Anthony~L. Caterini,
  Rahul Krishnan, and Gabriel Loaiza-Ganem.
\newblock A geometric explanation of the likelihood {OOD} detection paradox.
\newblock In \emph{Proceedings of the 41st International Conference on Machine
  Learning}, volume 235 of \emph{Proceedings of Machine Learning Research},
  pp.\  22908--22935, 2024.

\bibitem[Karras et~al.(2019)Karras, Laine, and Aila]{Karras2019-stylegan}
Tero Karras, Samuli Laine, and Timo Aila.
\newblock A style-based generator architecture for generative adversarial
  networks.
\newblock In \emph{IEEE/CVF Conference on Computer Vision and Pattern
  Recognition}, 2019.

\bibitem[Kazemi \& Sullivan(2014)Kazemi and Sullivan]{Kazemi2014-face}
Vahid Kazemi and Josephine Sullivan.
\newblock One millisecond face alignment with an ensemble of regression trees.
\newblock In \emph{IEEE Conference on Computer Vision and Pattern Recognition},
  pp.\  1867--1874, 2014.

\bibitem[Ke et~al.(2021)Ke, Wang, Wang, Milanfar, and Yang]{Ke2021-musiq}
Junjie Ke, Qifei Wang, Yilin Wang, Peyman Milanfar, and Feng Yang.
\newblock {MUSIQ}: Multi-scale image quality transformer.
\newblock In \emph{IEEE/CVF International Conference on Computer Vision}, 2021.

\bibitem[King(2009)]{King2009-dlib}
Davis~E King.
\newblock Dlib-ml: A machine learning toolkit.
\newblock \emph{Journal of Machine Learning Research}, 10:\penalty0 1755--1758,
  2009.

\bibitem[Kingma \& Gao(2023)Kingma and Gao]{Kingma2023-zc}
Diederik~P. Kingma and Ruiqi Gao.
\newblock Understanding diffusion objectives as the {ELBO} with simple data
  augmentation.
\newblock In \emph{Advances in Neural Information Processing Systems},
  volume~36, pp.\  70932--70956, 2023.

\bibitem[Kirichenko et~al.(2020)Kirichenko, Izmailov, and
  Wilson]{Kirichenko2020-im}
Polina Kirichenko, Pavel Izmailov, and Andrew~Gordon Wilson.
\newblock Why normalizing flows fail to detect out-of-distribution data.
\newblock In \emph{Advances in Neural Information Processing Systems},
  volume~33, pp.\  20578--20589, 2020.

\bibitem[Kong et~al.(2024)]{Kong2024-hunyuanvideo}
Weijie Kong et~al.
\newblock {HunyuanVideo}: A systematic framework for large video generative
  models.
\newblock \emph{arXiv preprint arXiv:2412.03603}, 2024.

\bibitem[Koukounas et~al.(2024)]{Koukounas2024-jinaclipv2}
Andreas Koukounas et~al.
\newblock {jina-clip-v2}: Multilingual multimodal embeddings for text and
  images.
\newblock \emph{arXiv preprint arXiv:2412.08802}, 2024.

\bibitem[Larson \& Chandler(2010)Larson and Chandler]{Larson2010-xm}
Eric~Cooper Larson and Damon~Michael Chandler.
\newblock Most apparent distortion: full-reference image quality assessment and
  the role of strategy.
\newblock \emph{Journal of Electronic Imaging}, 19\penalty0 (1):\penalty0
  011006, January 2010.
\newblock ISSN 1017-9909,1560-229X.
\newblock \doi{10.1117/1.3267105}.

\bibitem[Li et~al.(2026)]{Li2026-qwen3vlembedding}
Mingxin Li et~al.
\newblock {Qwen3-VL-Embedding} and {Qwen3-VL-Reranker}: A unified framework for
  state-of-the-art multimodal retrieval and ranking.
\newblock \emph{arXiv preprint arXiv:2601.04720}, 2026.

\bibitem[Li \& He(2026)Li and He]{Li2025-jit}
Tianhong Li and Kaiming He.
\newblock Back to basics: Let denoising generative models denoise.
\newblock In \emph{Proceedings of the IEEE/CVF Conference on Computer Vision
  and Pattern Recognition}, pp.\  36115--36125, 2026.

\bibitem[L{\'o}pez \& Triesch(2026)L{\'o}pez and Triesch]{Lopez2026-elbo}
Francisco~M. L{\'o}pez and Jochen Triesch.
\newblock Beauty is in the {ELBO} of the beholder: A variational account of
  processing fluency in face perception.
\newblock \emph{arXiv preprint arXiv:2608.24219}, 2026.

\bibitem[Ma et~al.(2026)Ma, Xu, and Zhang]{Ma2026-pixelgen}
Zehong Ma, Ruihan Xu, and Shiliang Zhang.
\newblock {PixelGen}: Improving pixel diffusion with perceptual supervision.
\newblock \emph{arXiv preprint arXiv:2602.02493}, 2026.

\bibitem[Mamassian(2004)]{Mamassian2004-ou}
Pascal Mamassian.
\newblock Impossible shadows and the shadow correspondence problem.
\newblock \emph{Perception}, 33\penalty0 (11):\penalty0 1279--1290, 1~November
  2004.
\newblock ISSN 0301-0066,1468-4233.
\newblock \doi{10.1068/p5280}.

\bibitem[Mittal et~al.(2012)Mittal, Moorthy, and Bovik]{Mittal2012-brisque}
Anish Mittal, Anush~Krishna Moorthy, and Alan~Conrad Bovik.
\newblock No-reference image quality assessment in the spatial domain.
\newblock \emph{IEEE Transactions on Image Processing}, 21\penalty0
  (12):\penalty0 4695--4708, 2012.

\bibitem[Mittal et~al.(2013)Mittal, Soundararajan, and Bovik]{Mittal2013-niqe}
Anish Mittal, Rajiv Soundararajan, and Alan~Conrad Bovik.
\newblock Making a ``completely blind'' image quality analyzer.
\newblock \emph{IEEE Signal Processing Letters}, 20\penalty0 (3):\penalty0
  209--212, 2013.

\bibitem[Motamed et~al.(2026)Motamed, Culp, Swersky, Jaini, and
  Geirhos]{Motamed2025-physicsiq}
Saman Motamed, Laura Culp, Kevin Swersky, Priyank Jaini, and Robert Geirhos.
\newblock Do generative video models understand physical principles?
\newblock In \emph{Proceedings of the IEEE/CVF Winter Conference on
  Applications of Computer Vision}, pp.\  948--958, 2026.

\bibitem[Nalisnick et~al.(2019)Nalisnick, Matsukawa, Teh, Gorur, and
  Lakshminarayanan]{Nalisnick2018-jh}
Eric Nalisnick, Akihiro Matsukawa, Yee~Whye Teh, Dilan Gorur, and Balaji
  Lakshminarayanan.
\newblock Do deep generative models know what they don't know?
\newblock In \emph{International Conference on Learning Representations}, 2019.

\bibitem[Nightingale et~al.(2019)Nightingale, Wade, Farid, and
  Watson]{Nightingale2019-ef}
Sophie~J Nightingale, Kimberley~A Wade, Hany Farid, and Derrick~G Watson.
\newblock Can people detect errors in shadows and reflections?
\newblock \emph{Attention, Perception, \& Psychophysics}, 81\penalty0
  (8):\penalty0 2917--2943, 28~November 2019.
\newblock ISSN 1943-3921,1943-393X.
\newblock \doi{10.3758/s13414-019-01773-w}.

\bibitem[Norman et~al.(2009)Norman, Lee, Phillips, Norman, Jennings, and
  McBride]{Norman2009-shadows}
J~Farley Norman, Young-lim Lee, Flip Phillips, Hideko~F Norman, L~RaShae
  Jennings, and T~Ryan McBride.
\newblock The perception of 3-{D} shape from shadows cast onto curved surfaces.
\newblock \emph{Acta Psychologica}, 131\penalty0 (1):\penalty0 1--11, 2009.
\newblock \doi{10.1016/j.actpsy.2009.01.007}.

\bibitem[Oquab et~al.(2024)Oquab, Darcet, Moutakanni, Vo, Szafraniec, Khalidov,
  Fernandez, Haziza, Massa, El-Nouby, et~al.]{Oquab2023-dinov2}
Maxime Oquab, Timoth{\'e}e Darcet, Th{\'e}o Moutakanni, Huy~V. Vo, Marc
  Szafraniec, Vasil Khalidov, Pierre Fernandez, Daniel Haziza, Francisco Massa,
  Alaaeldin El-Nouby, et~al.
\newblock {DINOv2}: Learning robust visual features without supervision.
\newblock \emph{Transactions on Machine Learning Research}, 2024.

\bibitem[Ostrovsky et~al.(2005)Ostrovsky, Cavanagh, and
  Sinha]{Ostrovsky2005-ev}
Yuri Ostrovsky, Patrick Cavanagh, and Pawan Sinha.
\newblock Perceiving illumination inconsistencies in scenes.
\newblock \emph{Perception}, 34\penalty0 (11):\penalty0 1301--1314, 2005.
\newblock ISSN 0301-0066.
\newblock \doi{10.1068/p5418}.

\bibitem[Podell et~al.(2024)Podell, English, Lacey, Blattmann, Dockhorn,
  M{\"u}ller, Penna, and Rombach]{Podell2023-sdxl}
Dustin Podell, Zion English, Kyle Lacey, Andreas Blattmann, Tim Dockhorn, Jonas
  M{\"u}ller, Joe Penna, and Robin Rombach.
\newblock {SDXL}: Improving latent diffusion models for high-resolution image
  synthesis.
\newblock In \emph{International Conference on Learning Representations}, 2024.

\bibitem[Radford et~al.(2021)Radford, Kim, Hallacy, Ramesh, Goh, Agarwal,
  Sastry, Askell, Mishkin, Clark, Krueger, and Sutskever]{Radford2021-clip}
Alec Radford, Jong~Wook Kim, Chris Hallacy, Aditya Ramesh, Gabriel Goh,
  Sandhini Agarwal, Girish Sastry, Amanda Askell, Pamela Mishkin, Jack Clark,
  Gretchen Krueger, and Ilya Sutskever.
\newblock Learning transferable visual models from natural language
  supervision.
\newblock In \emph{Proceedings of the 38th International Conference on Machine
  Learning}, volume 139, pp.\  8748--8763. PMLR, 2021.

\bibitem[Raviv \& Chechik(2024)Raviv and Chechik]{Raviv2024-ce}
Simon Raviv and Gal Chechik.
\newblock Assessing image quality using a simple generative representation.
\newblock \emph{arXiv preprint arXiv:2404.18178}, 2024.

\bibitem[Ren et~al.(2019)Ren, Liu, Fertig, Snoek, Poplin, DePristo, Dillon, and
  Lakshminarayanan]{Ren2019-likelihoodratio}
Jie Ren, Peter~J. Liu, Emily Fertig, Jasper Snoek, Ryan Poplin, Mark~A.
  DePristo, Joshua~V. Dillon, and Balaji Lakshminarayanan.
\newblock Likelihood ratios for out-of-distribution detection.
\newblock In \emph{Advances in Neural Information Processing Systems},
  volume~32, 2019.
\newblock URL
  \url{https://proceedings.neurips.cc/paper/2019/hash/1e79596878b2320cac26dd792a6c51c9-Abstract.html}.

\bibitem[Rombach et~al.(2022)Rombach, Blattmann, Lorenz, Esser, and
  Ommer]{Rombach2022-ldm}
Robin Rombach, Andreas Blattmann, Dominik Lorenz, Patrick Esser, and Bj{\"o}rn
  Ommer.
\newblock High-resolution image synthesis with latent diffusion models.
\newblock In \emph{IEEE/CVF Conference on Computer Vision and Pattern
  Recognition}, 2022.

\bibitem[Saad et~al.(2012)Saad, Bovik, and Charrier]{Saad2012-bliinds}
Michele~A Saad, Alan~Conrad Bovik, and Christophe Charrier.
\newblock Blind image quality assessment: A natural scene statistics approach
  in the {DCT} domain.
\newblock \emph{IEEE Transactions on Image Processing}, 21\penalty0
  (8):\penalty0 3339--3352, 2012.

\bibitem[Saha et~al.(2026)Saha, Konkle, and Alvarez]{Saha2026-edge}
Srijani Saha, Talia Konkle, and George~A Alvarez.
\newblock A unified account of lightness illusions via edge-based
  reconstruction of natural images.
\newblock \emph{bioRxiv preprint}, 2026.
\newblock \doi{10.64898/2026.04.07.716245}.
\newblock URL
  \url{https://www.biorxiv.org/content/10.64898/2026.04.07.716245v2}.

\bibitem[Sarkar et~al.(2024)Sarkar, Mai, Mahapatra, Lazebnik, Forsyth, and
  Bhattad]{Sarkar2023-es}
Ayush Sarkar, Hanlin Mai, Amitabh Mahapatra, Svetlana Lazebnik, David~A.
  Forsyth, and Anand Bhattad.
\newblock Shadows don't lie and lines can't bend! generative models don't know
  projective geometry... for now.
\newblock In \emph{Proceedings of the IEEE/CVF Conference on Computer Vision
  and Pattern Recognition}, pp.\  28140--28149, 2024.

\bibitem[Sclocchi et~al.(2025)Sclocchi, Favero, and Wyart]{Sclocchi2024-tv}
Antonio Sclocchi, Alessandro Favero, and Matthieu Wyart.
\newblock A phase transition in diffusion models reveals the hierarchical
  nature of data.
\newblock \emph{Proceedings of the National Academy of Sciences}, 122\penalty0
  (1):\penalty0 e2408799121, 2025.

\bibitem[Serr\`a et~al.(2020)Serr\`a, \`Alvarez, G\'omez, Slizovskaia,
  N\'u\~nez, and Luque]{Serra2020-input}
Joan Serr\`a, David \`Alvarez, Vicen\c{c} G\'omez, Olga Slizovskaia, Jos\'e~F.
  N\'u\~nez, and Jordi Luque.
\newblock Input complexity and out-of-distribution detection with
  likelihood-based generative models.
\newblock In \emph{International Conference on Learning Representations}, 2020.
\newblock URL \url{https://openreview.net/forum?id=SyxIWpVYvr}.

\bibitem[Shahgir et~al.(2024)Shahgir, Sayeed, Bhattacharjee, Ahmad, Dong, and
  Shahriyar]{Wang2024-illusionvqa}
Haz~Sameen Shahgir, Khondker~Salman Sayeed, Abhik Bhattacharjee, Wasi~Uddin
  Ahmad, Yue Dong, and Rifat Shahriyar.
\newblock {IllusionVQA}: A challenging optical illusion dataset for vision
  language models.
\newblock In \emph{Conference on Language Modeling}, 2024.

\bibitem[Sheikh \& Bovik(2006)Sheikh and Bovik]{Sheikh2006-vif}
Hamid~R Sheikh and Alan~C Bovik.
\newblock Image information and visual quality.
\newblock \emph{IEEE Transactions on Image Processing}, 15\penalty0
  (2):\penalty0 430--444, 2006.
\newblock \doi{10.1109/TIP.2005.859378}.

\bibitem[Silverman(1986)]{Silverman1986-density}
B.~W. Silverman.
\newblock \emph{Density Estimation for Statistics and Data Analysis}.
\newblock Chapman and Hall, London, 1986.

\bibitem[Sim{\'e}oni et~al.(2026)Sim{\'e}oni, Vo, Seitzer, Baldassarre, Oquab,
  Jose, Khalidov, Szafraniec, Yi, Ramamonjisoa, Massa, Haziza, Wehrstedt, Wang,
  Darcet, Moutakanni, Sentana, Roberts, Vedaldi, Tolan, Brandt, Couprie,
  Mairal, J{\'e}gou, Labatut, and Bojanowski]{Simeoni2025-dinov3}
Oriane Sim{\'e}oni, Huy~V. Vo, Maximilian Seitzer, Federico Baldassarre, Maxime
  Oquab, Cijo Jose, Vasil Khalidov, Marc Szafraniec, Seung~Eun Yi, Michael
  Ramamonjisoa, Francisco Massa, Daniel Haziza, Luca Wehrstedt, Jianyuan Wang,
  Timoth{\'e}e Darcet, Th{\'e}o Moutakanni, Leonel Sentana, Claire Roberts,
  Andrea Vedaldi, Jamie Tolan, John Brandt, Camille Couprie, Julien Mairal,
  Herv{\'e} J{\'e}gou, Patrick Labatut, and Piotr Bojanowski.
\newblock {DINOv3}.
\newblock \emph{Transactions on Machine Learning Research}, 2026.

\bibitem[Song et~al.(2025)Song, Liu, and Shou]{Song2025-diffsim}
Yiren Song, Xiaokang Liu, and Mike~Zheng Shou.
\newblock {DiffSim}: Taming diffusion models for evaluating visual similarity.
\newblock In \emph{Proceedings of the IEEE/CVF International Conference on
  Computer Vision}, pp.\  16904--16915, 2025.

\bibitem[Su et~al.(2020)Su, Yan, Zhu, Zhang, Ge, Sun, and
  Zhang]{Su2020-hyperiqa}
Shaolin Su, Qingsen Yan, Yu~Zhu, Cheng Zhang, Xin Ge, Jinqiu Sun, and Yanning
  Zhang.
\newblock Blindly assess image quality in the wild guided by a self-adaptive
  hyper network.
\newblock In \emph{IEEE/CVF Conference on Computer Vision and Pattern
  Recognition}, 2020.

\bibitem[{{Team Wan}} et~al.(2025)]{WanTeam2025-wan}
{{Team Wan}} et~al.
\newblock {Wan}: Open and advanced large-scale video generative models.
\newblock \emph{arXiv preprint arXiv:2503.20314}, 2025.

\bibitem[Thompson(1980)]{Thompson1980-thatcher}
Peter~Gage Thompson.
\newblock {Margaret Thatcher}: A new illusion.
\newblock \emph{Perception}, 9\penalty0 (4):\penalty0 483--484, 1980.
\newblock \doi{10.1068/p090483}.

\bibitem[T{\i}naz et~al.(2025)T{\i}naz, Fabian, and Soltanolkotabi]{Tinaz2025}
Berk T{\i}naz, Zalan Fabian, and Mahdi Soltanolkotabi.
\newblock Emergence and evolution of interpretable concepts in diffusion
  models.
\newblock In \emph{Advances in Neural Information Processing Systems},
  volume~38, pp.\  166943--166986, 2025.

\bibitem[Tschannen et~al.(2025)]{Tschannen2025-siglip2}
Michael Tschannen et~al.
\newblock {SigLIP 2}: Multilingual vision-language encoders with improved
  semantic understanding, localization, and dense features.
\newblock \emph{arXiv preprint arXiv:2502.14786}, 2025.

\bibitem[Wang et~al.(2004)Wang, Bovik, Sheikh, and Simoncelli]{Wang2004-ssim}
Zhou Wang, Alan~C Bovik, Hamid~R Sheikh, and Eero~P Simoncelli.
\newblock Image quality assessment: From error visibility to structural
  similarity.
\newblock \emph{IEEE Transactions on Image Processing}, 13\penalty0
  (4):\penalty0 600--612, 2004.
\newblock \doi{10.1109/TIP.2003.819861}.

\bibitem[Wilder et~al.(2019)Wilder, Adams, and Murray]{Wilder2019-ae}
John~D Wilder, Wendy~J Adams, and Richard~F Murray.
\newblock Shape from shading under inconsistent illumination.
\newblock \emph{Journal of Vision}, 19\penalty0 (6):\penalty0 2, 3~June 2019.
\newblock ISSN 1534-7362.
\newblock \doi{10.1167/19.6.2}.

\bibitem[Wu et~al.(2025)]{Wu2025-qwenimage}
Chenfei Wu et~al.
\newblock {Qwen-Image} technical report.
\newblock \emph{arXiv preprint arXiv:2508.02324}, 2025.

\bibitem[Xie et~al.(2026)Xie, Li, Lee, Zhang, Liu, and Wang]{Xie2026-yocausal}
You-Zhe Xie, Yu-Hsuan Li, Jie-Ying Lee, Kaipeng Zhang, Yu-Lun Liu, and Zhixiang
  Wang.
\newblock {YoCausal}: How far is video generation from world model? a causality
  perspective.
\newblock \emph{arXiv preprint arXiv:2605.30346}, 2026.
\newblock URL \url{https://arxiv.org/abs/2605.30346}.

\bibitem[Yang et~al.(2025)Yang, Teng, Zheng, Ding, Huang, Xu, Yang, Hong,
  Zhang, Feng, Yin, Zhang, Wang, Cheng, Xu, Gu, Dong, and
  Tang]{Yang2024-cogvideox}
Zhuoyi Yang, Jiayan Teng, Wendi Zheng, Ming Ding, Shiyu Huang, Jiazheng Xu,
  Yuanming Yang, Wenyi Hong, Xiaohan Zhang, Guanyu Feng, Da~Yin, Yuxuan Zhang,
  Weihan Wang, Yean Cheng, Bin Xu, Xiaotao Gu, Yuxiao Dong, and Jie Tang.
\newblock {CogVideoX}: Text-to-video diffusion models with an expert
  transformer.
\newblock In \emph{International Conference on Learning Representations}, 2025.

\bibitem[Yildirim et~al.(2020)Yildirim, Belledonne, Freiwald, and
  Tenenbaum]{Yildirim2020-efficient}
Ilker Yildirim, Mario Belledonne, Winrich Freiwald, and Josh Tenenbaum.
\newblock Efficient inverse graphics in biological face processing.
\newblock \emph{Science Advances}, 6\penalty0 (10):\penalty0 eaax5979, 2020.

\bibitem[Yuan et~al.(2026)Yuan, Pizzati, Pinto, Kunze, Laptev, Newman, Torr,
  and De~Martini]{Yuan2026-likephys}
Jianhao Yuan, Fabio Pizzati, Francesco Pinto, Lars Kunze, Ivan Laptev, Paul
  Newman, Philip Torr, and Daniele De~Martini.
\newblock {LikePhys}: Evaluating intuitive physics understanding in video
  diffusion models via likelihood preference.
\newblock In \emph{International Conference on Learning Representations}, 2026.
\newblock URL \url{https://openreview.net/forum?id=6UJf6B8RZ8}.

\bibitem[Yuille \& Kersten(2006)Yuille and Kersten]{Yuille2006-vision}
Alan Yuille and Daniel Kersten.
\newblock Vision as {Bayesian} inference: analysis by synthesis?
\newblock \emph{Trends in Cognitive Sciences}, 10\penalty0 (7):\penalty0
  301--308, 2006.

\bibitem[Zhang et~al.(2018)Zhang, Isola, Efros, Shechtman, and
  Wang]{Zhang2018-jh}
Richard Zhang, Phillip Isola, Alexei~A. Efros, Eli Shechtman, and Oliver Wang.
\newblock The unreasonable effectiveness of deep features as a perceptual
  metric.
\newblock In \emph{Proceedings of the IEEE Conference on Computer Vision and
  Pattern Recognition}, pp.\  586--595, 2018.

\bibitem[Zhang et~al.(2023)Zhang, Zhai, Wei, Yang, and Ma]{Zhang2023-liqe}
Weixia Zhang, Guangtao Zhai, Ying Wei, Xiaokang Yang, and Kede Ma.
\newblock Blind image quality assessment via vision-language correspondence: A
  multitask learning perspective.
\newblock In \emph{IEEE/CVF Conference on Computer Vision and Pattern
  Recognition}, 2023.

\end{thebibliography}
\bibliographystyle{iclr2027_conference}

  \clearpage
  \appendix
  \renewcommand{\thefigure}{S\arabic{figure}}
  \renewcommand{\thetable}{S\arabic{table}}
  \renewcommand{\theHfigure}{supp.\arabic{figure}}
  \renewcommand{\theHtable}{supp.\arabic{table}}
  \setcounter{figure}{0}
  \setcounter{table}{0}
  \section{Stimulus construction}\label{stimulus-construction}

\subsection{Thatcher faces.}
\begin{figure}[t]
\centering
\includegraphics[width=\linewidth]{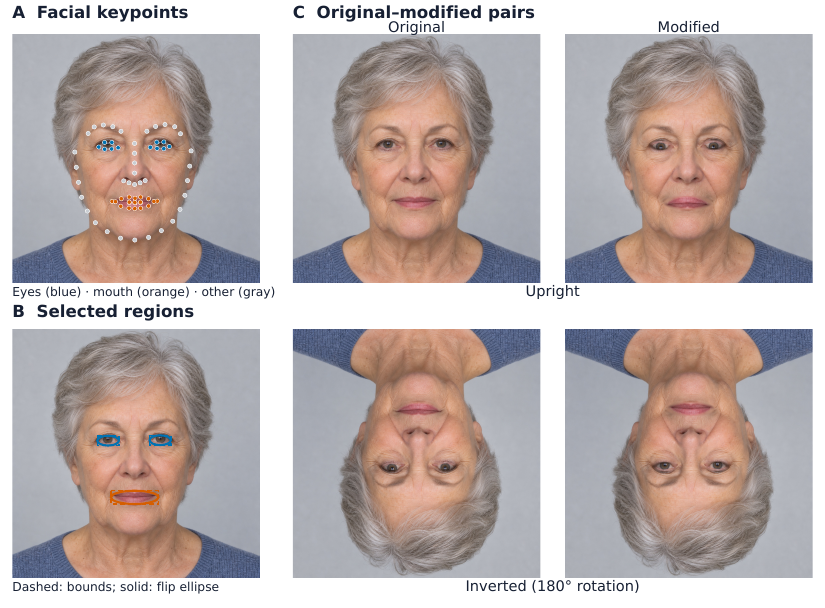}
\caption{\textbf{Thatcher stimulus construction.}
\textbf{(A)}~The 68 facial landmarks detected on an AI-generated fictional face (eyes: blue; mouth: orange; remaining points: gray).
\textbf{(B)}~Selected regions for feature inversion. Dashed rectangles indicate padded landmark bounds; solid ellipses delimit vertically inverted regions feathered along their boundaries.
\textbf{(C)}~Original (left) and Thatcherized (right) versions in upright (top) and inverted (bottom; exact $180^\circ$ rotation) orientations. The face is used only to illustrate the procedure and was not included in the human experiments or model analyses.}
\label{fig:thatcher-stimulus-construction}
\end{figure}

The stimulus set comprises 140 aligned $1024\times1024$ face identities, each evaluated across four conditions: original upright, Thatcherized upright, original inverted, and Thatcherized inverted (560 images total; construction procedure illustrated in Figure~\ref{fig:thatcher-stimulus-construction}).

\paragraph{Source dataset and license filtering.}
Source face images were sampled from the Flickr-Faces-HQ (FFHQ) dataset~\citep{Karras2019-stylegan}, which provides aligned $1024\times1024$ portraits. Source images were restricted to entries marked CC0 1.0 or Public Domain Mark 1.0 in the official FFHQ metadata (\texttt{ffhq-dataset-v2.json}). From this pool, we chose 140 distinct public-domain face identities for the experimental stimuli, and an additional 5 disjoint identities for practice trials (10 practice trials presenting original and modified pairs across orientations).

\paragraph{Feature inversion and boundary blending.}
To generate Thatcherized faces, we adapted the open-source pipeline by \citet{Alimohammadi2020-thatcher} (released under the GNU General Public License v3.0). For each face image, 68 facial landmarks were detected using \texttt{dlib}~\citep{King2009-dlib} with its pre-trained shape predictor~\citep{Kazemi2014-face} (Figure~\ref{fig:thatcher-stimulus-construction}A). Landmarks corresponding to the left eye (points 36--41), right eye (points 42--47), and mouth (points 48--67) were used to define bounding regions with modest padding offsets around each feature (Figure~\ref{fig:thatcher-stimulus-construction}B, dashed rectangles). Within each feature region, pixels inside an elliptical mask were vertically inverted in place across the feature's horizontal midline (Figure~\ref{fig:thatcher-stimulus-construction}B, solid ellipses). To minimize high-contrast seam artifacts and artificial edges, Gaussian blending feathered the transition along the elliptical perimeter into the surrounding skin. Inverted stimuli were generated by rotating both the original and Thatcherized images by $180^\circ$ (Figure~\ref{fig:thatcher-stimulus-construction}C). 

\subsection{Illumination scenes.}
We generated synthetic $1024\times1024$ stimuli using Kubric~\citep{Greff2022-kubric} across three physical operations: cast shadows (\emph{Shadow}), surface reflections (\emph{Reflection}), and lighting direction (\emph{Light direction}). Each operation comprised 48 base configurations. Horizontally mirroring each base pair produced another 48 pairs, yielding 96 pairs per operation and 288 pairs in total.

\begin{figure}[p]
\centering
\includegraphics[width=\linewidth]{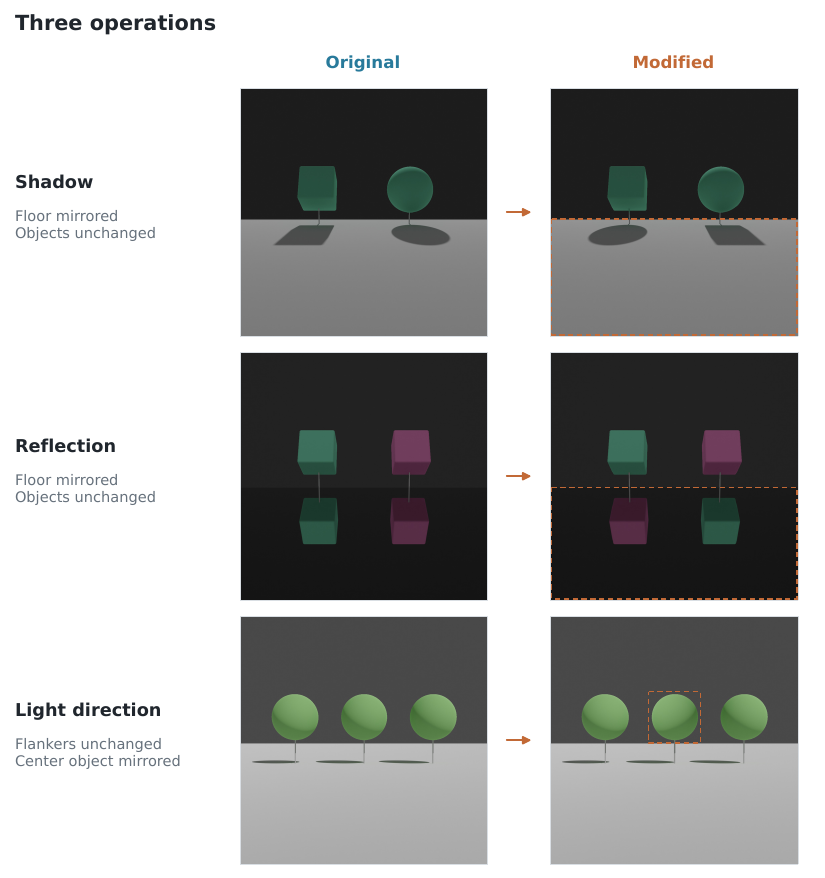}
\caption{\textbf{Illumination stimulus construction.}
Original (left) and modified (right) images for the three operations. The floor region is horizontally flipped in Shadow and Reflection, and the central object in Light direction. Orange outlines indicate the affected regions.}
\label{fig:illumination-stimulus-construction}
\end{figure}

\paragraph{Horizontal-flip manipulations.}
In \emph{Shadow} and \emph{Reflection}, we horizontally flipped the floor region while leaving the objects unchanged, exchanging their shadows or reflections (Figure~\ref{fig:illumination-stimulus-construction}). The two objects were placed symmetrically about the image's vertical midline, so the flip mapped their positions onto one another without introducing a positional offset. Additionally, objects were elevated above the floor (see below) to ensure a spatial gap; if objects contacted the ground, swapping the floor would alter contact contours or introduce spurious edge discontinuities at the object base, confounding loss differences with low-level edge artifacts. In \emph{Light direction}, we horizontally flipped the central object while leaving the floor and flanking objects unchanged. This reversed its shading relative to the flanking objects and, in low-height scenes, disrupted the correspondence between its shading and its unchanged cast shadow. Every intervention rearranged existing pixels without adding newly rendered visual content.

\begin{figure}[p]
\centering
\includegraphics[width=\linewidth]{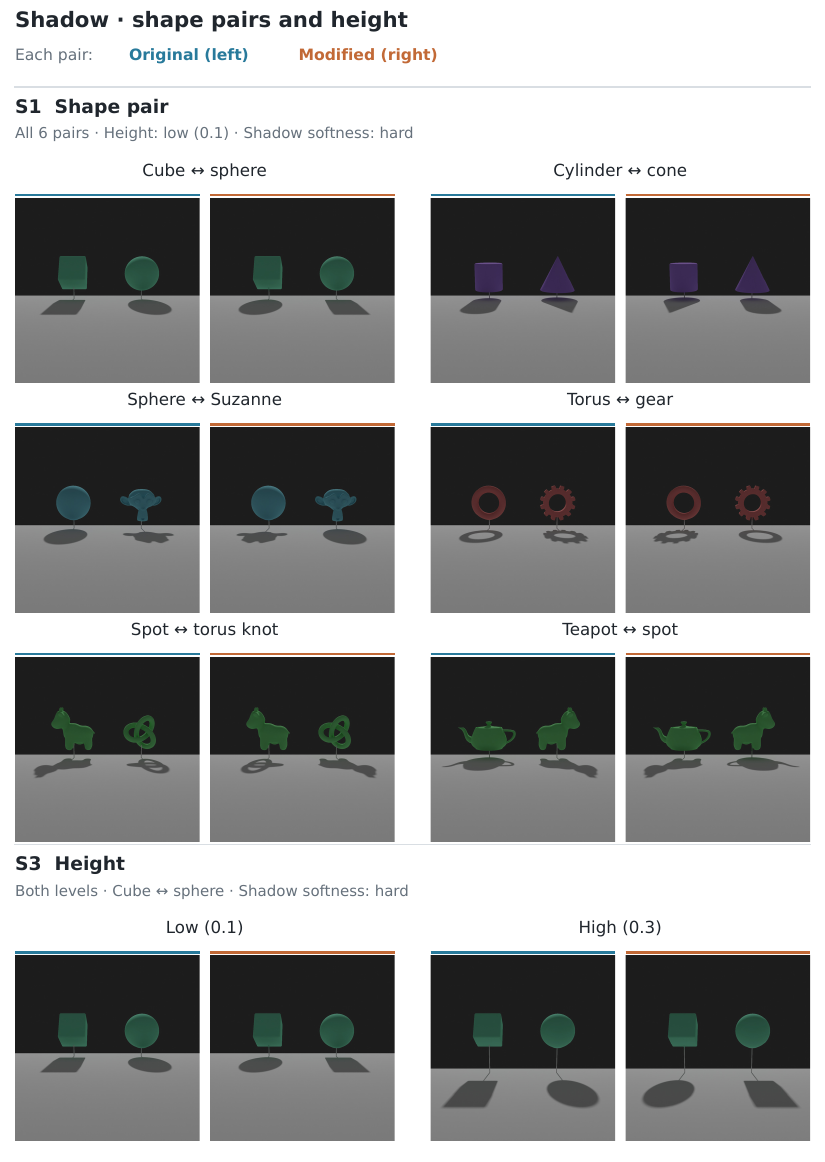}
\caption{\textbf{Shadow: shape pairs and height.}
Each example pairs the full original image (left, blue rule) with its modification (right, orange rule). S1 exchanges shadows between distinct object shapes to introduce salient, categorical geometric violations. S3 compares both object heights for the cube--sphere pair, testing whether elevating objects above the floor reduces the salience of shadow inconsistencies.}
\label{fig:illumination-stimulus-factors}
\label{fig:illumination-stimulus-shadow-shapes}
\end{figure}

\paragraph{Scene factors.}
Figures~\ref{fig:illumination-stimulus-factors}--\ref{fig:illumination-stimulus-light-settings} illustrate the stimuli across all 44 analysis levels evaluated in Appendix~\ref{illumination-factor-profiles}. To present these levels concisely without displaying every factorial combination, object variations (shape, rotation, and color) are shown under shared baseline settings, while separate pairwise panels show both levels of each environmental factor (such as height, shadow softness, or surface roughness).

For \emph{Shadow}, six Shape pair cases (S1) and six Rotation cases (S2) were each crossed with \emph{Height} (S3: low, 0.1; high, 0.3) and \emph{Shadow softness} (S4: soft or hard), giving $(6+6)\times2\times2=48$ base configurations. Soft shadows have blurred edges and were produced by a large area light; hard shadows have sharp edges and were produced by a small, point-like light source.

For \emph{Reflection}, four Shape pair cases (R1), four Rotation cases (R2), and four Color cases (R3) were each crossed with Height (R4: low, 0.1; high, 0.3) and \emph{Floor roughness} (R5: smooth, 0.0; rough, 0.3), giving $(4+4+4)\times2\times2=48$ configurations. The four Color cases combine two shapes, cube and torus, with a small or large hue shift (0.05 or 0.5 turns on the hue circle). Smooth floors produce mirror-like reflections; rough floors blur them. Shape pair, Rotation, and Color are separate sets of cases, not mutually crossed factors.

For \emph{Light direction}, six levels of \emph{Shape} (L1: cube, cylinder, sphere, Spot, teapot, and torus knot) were fully crossed with \emph{Material} (L2: diffuse or metal), Height (L3: low, 0.2; high, 2.0), and \emph{Light angle} (L4: $15^\circ$ or $30^\circ$), giving $6\times2\times2\times2=48$ configurations. Cast shadows were disabled in the high condition. Heights are expressed in renderer scene units.

\begin{figure}[p]
\centering
\includegraphics[width=\linewidth]{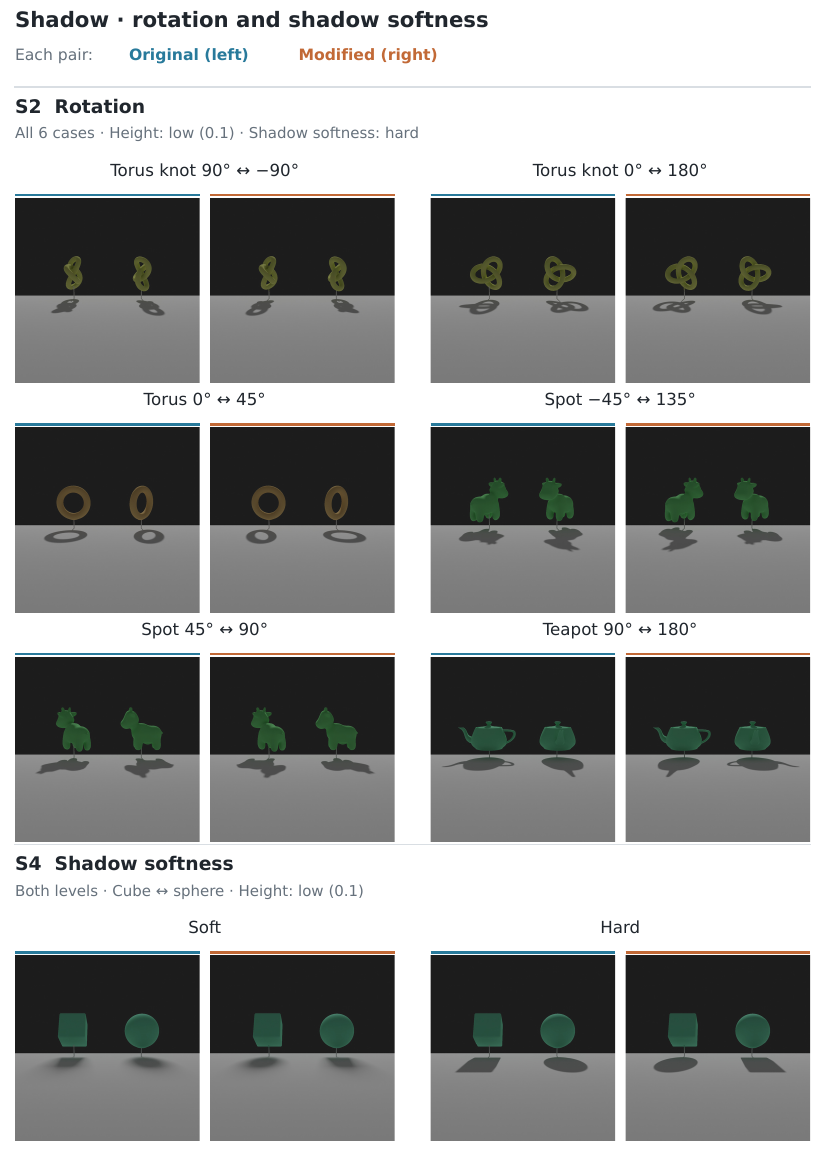}
\caption{\textbf{Shadow: rotations and shadow softness.}
S2 shows all six orientation pairs, with angles indicating object orientations. In modified images, exchanging shadows alters directional correspondence: this inconsistency is subtle for shapes lacking distinct landmarks (e.g., torus knot) or pairs with near-symmetric 2D silhouettes despite 3D depth asymmetry (e.g., Spot $-45^\circ\leftrightarrow135^\circ$), requiring observers to scrutinize depth relationships, whereas prominent directional features (e.g., teapot spout) make it immediately obvious. S4 compares soft and hard shadows, testing whether blurred boundaries increase tolerance to discrepancies.}
\label{fig:illumination-stimulus-shadow-rotations}
\end{figure}

\begin{figure}[p]
\centering
\includegraphics[width=\linewidth]{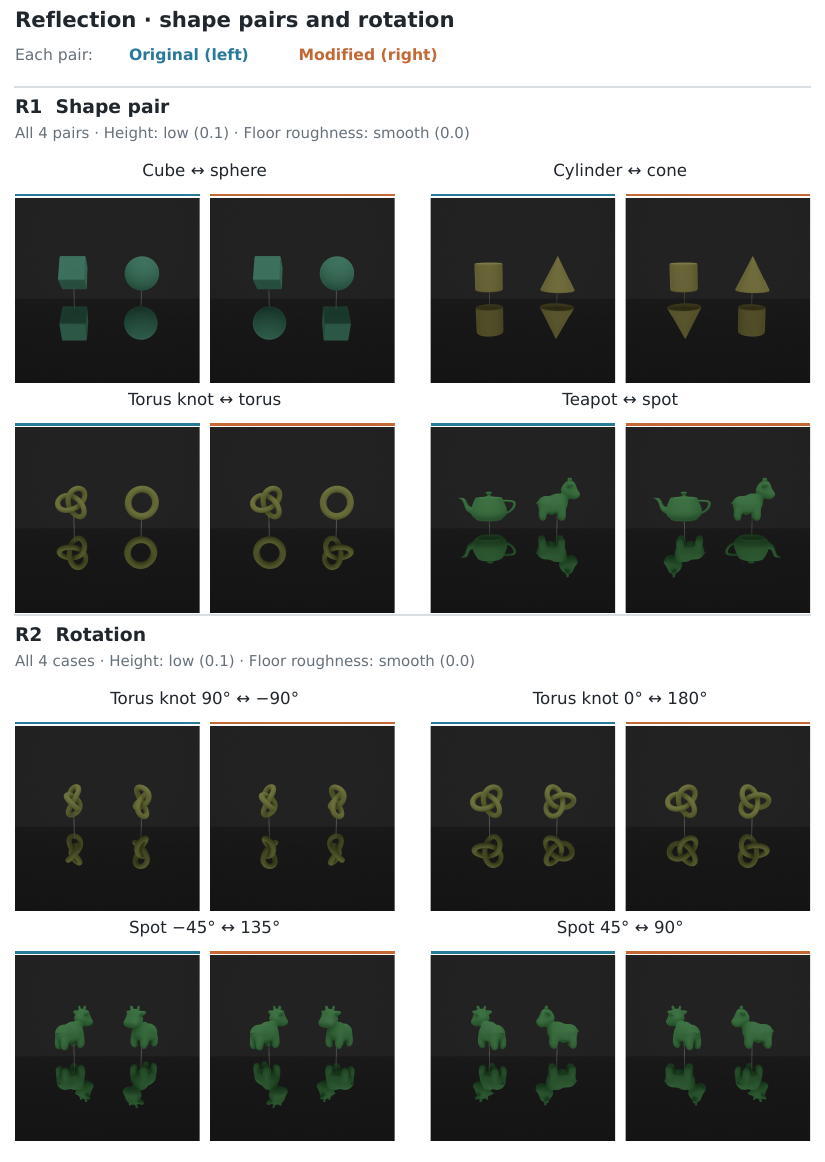}
\caption{\textbf{Reflection: shape pairs and rotations.}
R1 shows all four shape pairs and R2 all four orientation pairs, with height and floor roughness held at the baseline settings. R1 introduces categorical shape discrepancies between objects and their reflections, whereas R2 tests sensitivity to mirror-reflection symmetry across 3D orientations. In R2, inconsistencies are subtle and difficult to detect for shapes lacking distinctive landmarks (e.g., torus knot) or pairs appearing near-symmetric in 2D (e.g., Spot $-45^\circ\leftrightarrow135^\circ$), whereas salient directional asymmetry (e.g., Spot $45^\circ\leftrightarrow90^\circ$) makes violations apparent.}
\label{fig:illumination-stimulus-reflection-shapes}
\end{figure}

\begin{figure}[p]
\centering
\includegraphics[width=\linewidth]{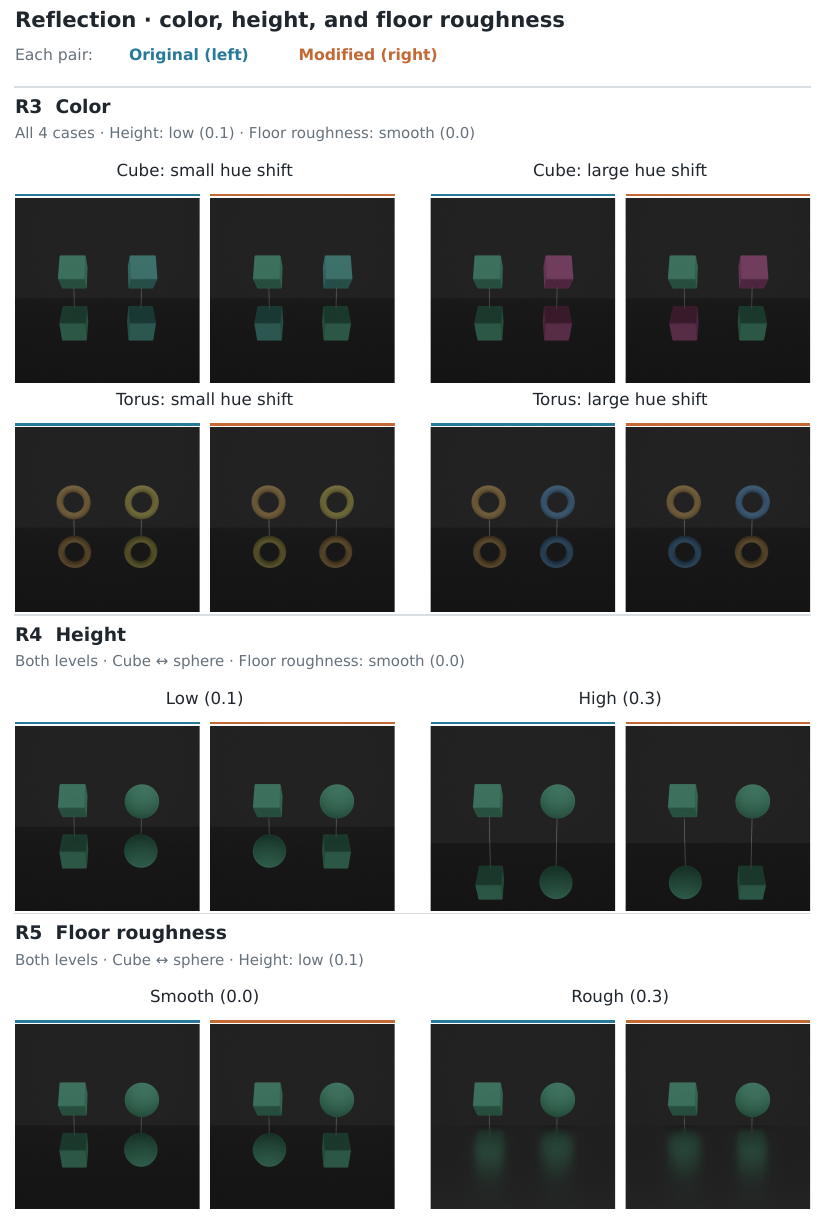}
\caption{\textbf{Reflection: color, height, and floor roughness.}
R3 introduces chromatic mismatches without altering geometry, testing sensitivity to color differences. R4 and R5 compare heights and surface roughness, examining how spatial proximity and blur modulate the salience of reflection anomalies.}
\label{fig:illumination-stimulus-reflection-settings}
\end{figure}

\begin{figure}[p]
\centering
\includegraphics[width=\linewidth]{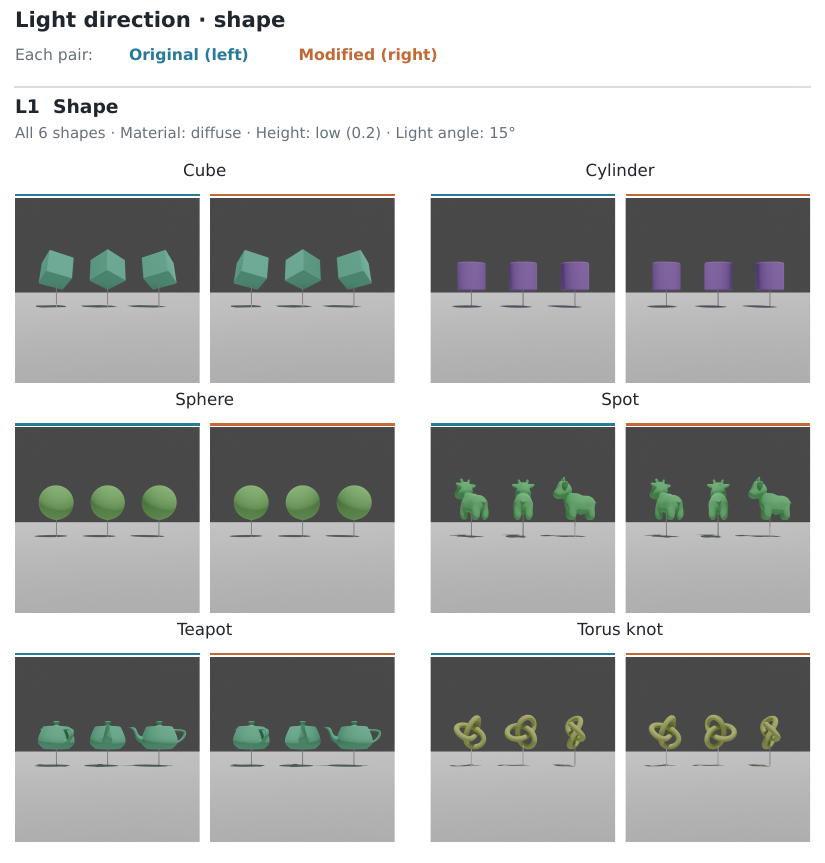}
\caption{\textbf{Light direction: shapes.}
L1 shows every shape at shared baseline settings. Horizontally mirroring only the central object reverses its shading relative to the flanking objects and unchanged cast shadows. This tests how object geometry affects lighting consistency, contrasting simple shapes (e.g., sphere) where illumination direction is unambiguous with complex shapes where lighting discrepancies are easily overlooked.}
\label{fig:illumination-stimulus-light-shapes}
\end{figure}

\begin{figure}[p]
\centering
\includegraphics[width=\linewidth]{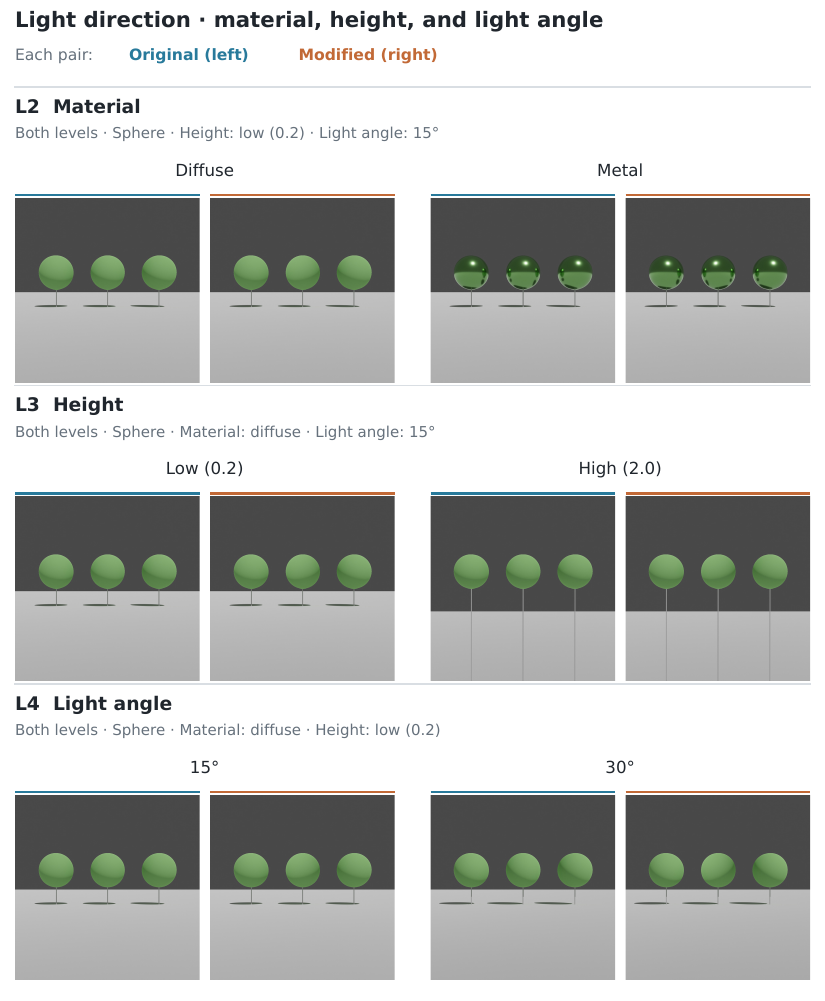}
\caption{\textbf{Light direction: material, height, and light angle.}
Each factor is illustrated at both levels using spheres, holding other factors fixed. L2 tests how surface material properties modulate lighting perception (diffuse shading gradients vs.\ specular highlights). L3 examines whether the presence of a consistent cast shadow strengthens the perceived lighting conflict compared to floating objects without shadows. L4 tests the effect of illumination inclination ($15^\circ$ vs.\ $30^\circ$).}
\label{fig:illumination-stimulus-light-settings}
\end{figure}

\clearpage

\section{Human experiment details}\label{human-experiment-details}

\subsection{Apparatus and display interface}
Online psychophysical experiments were developed using jsPsych (v7)~\citep{deLeeuw2023-jspsych} and hosted on a dedicated web server. Prior to starting the experiment, participants were required to switch their browsers to full-screen mode, and automated checks verified compatible rendering environments. Stimuli were displayed in the center of the screen at $512 \times 512$ pixels against a neutral background.

Participants provided naturalness ratings on a discrete 5-point horizontal slider with text labels: 1 (\emph{Very unnatural}), 2 (\emph{Unnatural}), 3 (\emph{Slightly unnatural}), 4 (\emph{Moderately natural}), and 5 (\emph{Natural}). To avoid anchoring effects, the slider handle was initialized to a random integer position ($1\text{--}5$) on every trial. Participants had to click or move the handle before the continue button was enabled, so that retaining the initial position required an active response. Stimuli remained on screen without time limits until submitted, followed by a 500~ms blank inter-trial interval (ITI).

\subsection{Task instructions}
Because the stimulus manipulations introduced subtle violations of configural or physical regularities within otherwise realistic scenes, relying solely on an unguided, generic query of ``naturalness'' risks inattentive oversights or idiosyncratic evaluation criteria (e.g., judging superficial photographic aesthetics rather than physical plausibility). To ensure that participants attended to the critical structural attributes and to calibrate rating criteria across observers, participants were familiarized with the candidate manipulation types through illustrated instructions and visual examples prior to the experiment:
\begin{itemize}
    \renewcommand{\labelitemi}{\textendash}
    \setlength{\itemsep}{4pt}
    \item \textbf{Thatcher task:} Participants were instructed: \emph{``Rate the naturalness of the facial appearance on a 5-point scale. Internal facial parts (eyes and mouth) may have artificial image manipulations applied. If you perceive a natural face without manipulation, select 5 (Natural). If you feel unnaturalness due to manipulation, select 1--4 according to the degree of unnaturalness. Faces may also appear upside down; please evaluate them using the same criteria, and refrain from physically tilting your head or rotating your display.''}
    \item \textbf{Shadow swap:} Participants were instructed: \emph{``Rate the naturalness of the shadow shape formed on the floor surface on a 5-point scale. If the shadow shape appears consistent with the object shape, select 5 (Natural). If the shadow shape appears inconsistent with the object shape, select 1--4 according to the perceived unnaturalness.''}
    \item \textbf{Reflection swap:} Participants were instructed: \emph{``Rate the naturalness of the reflection on the floor surface on a 5-point scale. If the appearance of the reflected image matches the physical object, select 5 (Natural); if it does not match, select 1--4 according to the degree of unnaturalness.''}
    \item \textbf{Light-direction swap:} Participants were instructed: \emph{``In an image depicting three objects illuminated from above, rate the naturalness of the lighting on the central object on a 5-point scale, focusing on whether the lighting direction on the central object is consistent with the surrounding objects and the floor shadows. If the lighting direction is mutually consistent, select 5 (Natural); if it appears contradictory, select 1--4 according to the degree of unnaturalness.''}
\end{itemize}

\subsection{Experimental design and trial structure}
\paragraph{Thatcher task.} The 140 base face scenes ($140 \times 2 \text{ orientations} = 280 \text{ pairs}$; 560 images) were divided into four disjoint sets of 35 (A--D). The four observer groups viewed A upright and B inverted, B upright and C inverted, C upright and D inverted, or D upright and A inverted. Each participant therefore viewed 35 upright faces and 35 different inverted faces, with each identity presented in only one orientation. Both original and modified versions of each scene were evaluated by the same participant, providing within-subject paired perceptual differences $h_i = R(x_i^{\mathrm{orig}}) - R(x_i^{\mathrm{mod}})$. Each main block contained 140 trials (35 upright faces plus 35 different inverted faces, each shown in original and modified versions) and 6 repeated trials for reliability checking (146 trials total). In data processing, repeated trials were deduplicated by retaining the first presentation.

\paragraph{Illumination task.} The three illumination interventions (Shadow, Reflection, Light direction; 96 scenes each, 288 pairs total) were evaluated in a between-subjects design: each participant evaluated only one intervention type. Within each intervention, the 96 scenes were randomly divided into four disjoint subsets of 24. Each of four observer groups evaluated a different combination of three subsets (72 scenes), so each scene was evaluated by three of the four groups. Both original and modified versions of each assigned scene were evaluated within-subject. Each main block contained 144 trials ($72 \text{ scenes} \times 2 \text{ versions}$) and 6 repeated trials (150 trials total; first presentation retained).

\paragraph{Procedure.} Each session progressed through: full-screen switch $\to$ illustrated instructions $\to$ a practice block (10 trials for Thatcher; condition-specific practice for Illumination) $\to$ an option to re-read instructions $\to$ the main block with trials presented in randomized order $\to$ data submission.

\subsection{Participant demographics, screening, and response counts}
Participants were adults aged 20--49 years, with balanced representation across age groups (20s, 30s, and 40s) and sexes. A total of 162 participant submissions were collected for the Thatcher task and 167 for the Illumination task. Since two Illumination submissions came from duplicate participant identities, we retained each participant's first submission while discarding the latter, leaving 165 participants before response-consistency screening. Although the experiment included 6 repeated trials originally intended for intra-subject test--retest screening, strict consistency filtering on these few trials proved overly conservative due to normal perceptual rating jitter on the 5-point scale, excluding an unnecessarily large fraction of otherwise attentive participants. We therefore adopted a global response-consistency screening criterion. Repeated trials were deduplicated by retaining the first presentation, and for each participant, we computed the Pearson correlation $r$ between their ratings across unique images and the corresponding image-wise mean ratings of all other participants who rated those images.

\begin{figure}[!htbp]
\centering
\includegraphics[width=.96\linewidth]{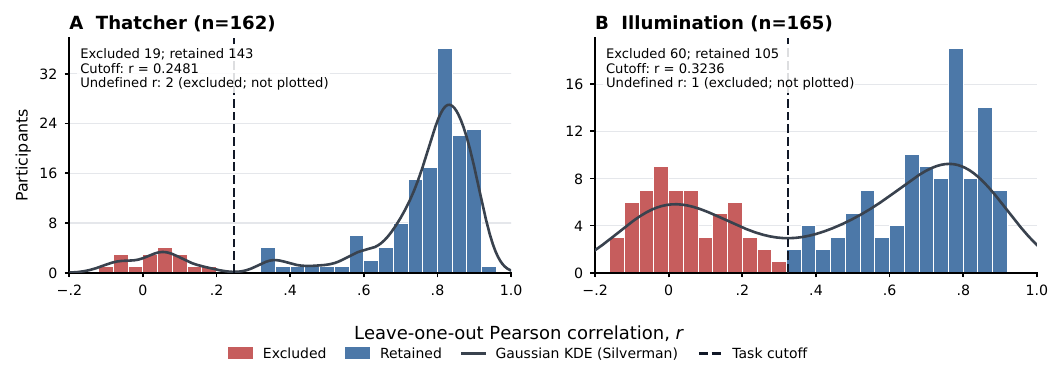}
\caption{\textbf{Participant response consistency and task-specific screening cutoffs.}
Distributions of leave-one-out Pearson correlations $r$ for (A) Thatcher ($n=162$) and (B) Illumination ($n=165$), after duplicate submissions and repeated presentations were reduced by retaining the first occurrence. Each participant's ratings are correlated with the image-wise means of the other participants. Red and blue bars denote excluded and retained participants, respectively; histogram bins have width $.04$. Solid curves show Gaussian kernel density estimates with Silverman's bandwidths ($b=.03836$ and $.11157$), scaled to histogram counts. Dashed lines mark the task-specific cutoffs ($r=.2481$ and $.3236$). Undefined correlations (2 for Thatcher, 1 for Illumination) are excluded from the plots and counted among excluded participants.}
\label{fig:human-response-consistency-histogram}
\end{figure}

We estimated a Gaussian kernel density for each task using Silverman's rule-of-thumb bandwidth~\citep{Silverman1986-density}, $b=0.9\min(s,\mathrm{IQR}/1.34)n^{-1/5}$, where $s$, $\mathrm{IQR}$, and $n$ are the sample standard deviation, interquartile range, and number of finite correlations. We selected the deepest trough relative to its flanking peaks (Figure~\ref{fig:human-response-consistency-histogram}), yielding cutoffs of $r = .2481$ for Thatcher and $r = .3236$ for Illumination. Retaining participants who met or exceeded these thresholds left 143 participants for Thatcher (88.3\%) and 105 for Illumination (63.6\%).

After screening, image items received the following numbers of independent human ratings:
\begin{itemize}
    \renewcommand{\labelitemi}{\textendash}
    \setlength{\itemsep}{0pt}
    \setlength{\parsep}{0pt}
    \item \textbf{Thatcher task:} Mean $35.8$ responses per item (range: $29\text{--}44$ responses; upright mean: $35.8$, inverted mean: $35.8$).
    \item \textbf{Illumination task:} Overall mean $26.2$ responses per item (range: $9\text{--}43$). By condition, Shadow swap averaged $40.5$ responses (range: $38\text{--}43$; 54 participants retained), Reflection swap averaged $26.2$ responses (range: $22\text{--}29$; 35 participants), and Light-direction swap averaged $12.0$ responses (range: $9\text{--}14$; 16 participants).
\end{itemize}

The variation in sample size across Illumination conditions stemmed primarily from differences in exclusion rates during screening: before screening, the Shadow, Reflection, and Light-direction conditions enrolled 69, 44, and 52 participants, of whom 15 (21.7\%), 9 (20.5\%), and 36 (69.2\%), respectively, were excluded. In participant-disjoint split-half analyses (10,000 group-stratified random splits), split-half reliability in the Light-direction condition was $r=.144$ among excluded participants, compared with $r=.668$ among retained participants (and $.960$ for Shadow, $.932$ for Reflection). The excluded Light-direction group showed little average separation between original and modified ratings: the mean original-minus-modified difference across the 96 pairs was $0.051$ points, compared with $0.996$ points in the retained group.

\paragraph{Filtering robustness.} To verify that our findings do not depend on the specific screening threshold, we recomputed all item-level aggregates without any participant exclusions. The unfiltered item aggregates correlated closely with the screened aggregates ($r = .986$ for Thatcher, $r = .942$ for Illumination), and the cross-model sensitivity--alignment association was preserved ($r=.663$ with and without screening on Thatcher; $.821$ with screening and $.847$ without screening on Illumination).
Because Light direction had the highest exclusion rate, we checked whether screening affected model--human correspondence in this condition. Across the 96 image pairs, the mean original--modified rating differences calculated from all 52 participants closely tracked those from the 16 retained participants ($r=.891$). Model--human correlations were also similar (median across all 25 generative models: $.277$ with screening and $.255$ without).

\section{Model details and evaluation specifications}\label{model-and-runtime-manifest}

This section summarizes the common loss probing setup and specifications for all 25 evaluated generative models (Table~\ref{tab:generative-model-suite}). Model-specific schedulers, identifiers, and environment configurations are archived with the accompanying code and configuration files.

\subsection{Generative loss probing}
For each generative model $m$ and test image $x$, the model-native loss $L_m(x)$ is computed by averaging its squared prediction error across a fixed grid of timesteps $T_m$ and noise realizations $E_m$:
\begin{equation}
L_m(x) = \frac{1}{|T_m||E_m|} \sum_{t \in T_m} \sum_{\epsilon \in E_m} \ell_m(x, t, \epsilon).
\end{equation}
Here, $\ell_m$ is the mean squared error in the parameterization of the model's denoising objective (noise, velocity, or flow), averaged over channel, spatial, and, where applicable, temporal dimensions. For latent-space models, losses are evaluated on encoded inputs and do not include an image-reconstruction term. JiT and PixelGen \citep{Li2025-jit,Ma2026-pixelgen} predict clean images and compute their denoising losses in velocity space. We follow this definition, using the residual $(\hat{x}-x)/\max(1-t,0.05)$, where $\hat{x}$ is the predicted clean image, $t$ is the clean-image interpolation coefficient, and the denominator floor follows the original implementations.

\noindent\textbf{Timestep grids ($T_m$).} Image models were evaluated at 100 scheduler-defined timesteps and video models at 50. The grids followed each model's implemented variance-preserving or shifted-flow schedule and were averaged with equal weight; their numerical timestep coordinates are model-specific and not directly comparable across architectures.

\noindent\textbf{Noise realizations ($E_m$).} Each model evaluation used a fixed bank of 20 standard Gaussian noise tensors. This bank was instantiated once and shared across all stimuli and both members of each pair, isolating content-dependent differences under matched noise draws.

\noindent\textbf{Conditioning and guidance.} Unconditional models used their native null-class or unconditional branch. Text-conditioned models used an empty or neutral prompt in a single forward pass without classifier-free guidance (CFG) branch mixing. The embedded guidance scale was set to $1.0$ for both FLUX.1 dev and HunyuanVideo.

\noindent\textbf{Image preprocessing.} All 16 image models received RGB inputs resized to $512 \times 512$ pixels before model-specific normalization or latent encoding.

\noindent\textbf{Video preprocessing.} Each source RGB image was resized by bilinear interpolation to fit the model's input resolution (Table~\ref{tab:generative-model-suite}), preserving aspect ratio, then center-padded with a constant value of $.5$. The padded frame was repeated to form a static clip. CogVideoX1.5 uses 21 repeats to yield six latent frames, satisfying its two-frame temporal patch size; the other eight models use 17.

\noindent\textbf{Compute environment and numerical precision.} Model probing was executed within isolated container environments on NVIDIA RTX A6000 (48\,GB) and A100 (40\,GB) GPUs, using native \texttt{bfloat16}, \texttt{float16}, or \texttt{float32} precision compatible with each model. Full software dependencies and configuration files are provided in the supplementary material.

\begin{table}[!htbp]
\centering
\small
\caption{\textbf{Evaluated open generative models.} Specifications of the 25 generative models across 11 families, grouped by architectural domain: pixel-space image models, latent-space image models, and spatiotemporal video models. Superscripts $^\mathrm{T}$ and $^\mathrm{I}$ mark the top five generative models selected by overall human alignment in Thatcher and Illumination, respectively; these selections are held fixed across conditions (Section~\ref{condition-specific-partial-correlation}).}
\label{tab:generative-model-suite}
\vspace{4pt}
\resizebox{\linewidth}{!}{%
\begin{tabular}{l@{\hspace{1.5em}}lllccc}
\toprule
\textbf{Family} & \textbf{Model} & \textbf{Identifier / Source Repository} & \textbf{Architecture} & \textbf{Evaluated Resolution} & \textbf{Steps ($|T_m|$)} & \textbf{Noise ($|E_m|$)} \\
\midrule
\multicolumn{7}{l}{\textbf{Pixel-space image models}} \\
JiT & JiT-B/32 & \texttt{LTH14/JiT::jit-b-32/checkpoint-last.pth} & Pixel DiT (ImageNet) & $512 \times 512$ & 100 & 20 \\
JiT & JiT-L/32 & \texttt{LTH14/JiT::jit-l-32/checkpoint-last.pth} & Pixel DiT (ImageNet) & $512 \times 512$ & 100 & 20 \\
JiT & JiT-H/32 & \texttt{LTH14/JiT::jit-h-32/checkpoint-last.pth} & Pixel DiT (ImageNet) & $512 \times 512$ & 100 & 20 \\
PixelGen & PixelGen 512 & \texttt{zehongma/PixelGen::PixelGen\_XXL\_T2I.ckpt} & Pixel DiT / Flow & $512 \times 512$ & 100 & 20 \\
HiDream & HiDream-O1-Image & \texttt{HiDream-ai/HiDream-O1-Image} & Pixel-space UiT / Flow & $512 \times 512$ & 100 & 20 \\
\midrule
\multicolumn{7}{l}{\textbf{Latent-space image models}} \\
Stable Diffusion & SD v1.5$^\mathrm{T}$ & \texttt{sd-legacy/stable-diffusion-v1-5} & Latent U-Net (DDPM) & $512 \times 512$ & 100 & 20 \\
Stable Diffusion & SDXL Base 1.0 & \texttt{stabilityai/stable-diffusion-xl-base-1.0} & Latent U-Net (DDPM) & $512 \times 512$ & 100 & 20 \\
Stable Diffusion & SD 3 Medium & \texttt{stabilityai/stable-diffusion-3-medium-diffusers} & Latent MMDiT (Flow) & $512 \times 512$ & 100 & 20 \\
FLUX.1 & FLUX.1 dev & \texttt{black-forest-labs/FLUX.1-dev} & Latent MMDiT (Flow) & $512 \times 512$ & 100 & 20 \\
FLUX.1 & FLUX.1 schnell$^\mathrm{T}$ & \texttt{black-forest-labs/FLUX.1-schnell} & Latent MMDiT (Distilled) & $512 \times 512$ & 100 & 20 \\
FLUX.2 & FLUX.2 klein base 4B & \texttt{black-forest-labs/FLUX.2-klein-base-4B} & Latent MMDiT (Flow) & $512 \times 512$ & 100 & 20 \\
FLUX.2 & FLUX.2 klein 4B$^\mathrm{T}$ & \texttt{black-forest-labs/FLUX.2-klein-4B} & Latent MMDiT (Flow) & $512 \times 512$ & 100 & 20 \\
FLUX.2 & FLUX.2 klein base 9B & \texttt{black-forest-labs/FLUX.2-klein-base-9B} & Latent MMDiT (Flow) & $512 \times 512$ & 100 & 20 \\
FLUX.2 & FLUX.2 klein 9B$^\mathrm{T}$ & \texttt{black-forest-labs/FLUX.2-klein-9B} & Latent MMDiT (Flow) & $512 \times 512$ & 100 & 20 \\
Qwen-Image & Qwen-Image$^\mathrm{I}$ & \texttt{Qwen/Qwen-Image} & Latent MMDiT (Flow) & $512 \times 512$ & 100 & 20 \\
Qwen-Image & Qwen-Image-2512$^\mathrm{I}$ & \texttt{Qwen/Qwen-Image-2512} & Latent MMDiT (Flow) & $512 \times 512$ & 100 & 20 \\
\midrule
\multicolumn{7}{l}{\textbf{Spatiotemporal video models (17 frames; CogVideoX1.5: 21)}} \\
CogVideoX & CogVideoX-2B & \texttt{THUDM/CogVideoX-2b} & 3D VAE Latent DiT & $720 \times 480$ & 50 & 20 \\
CogVideoX & CogVideoX-5B & \texttt{THUDM/CogVideoX-5b} & 3D VAE Latent DiT & $720 \times 480$ & 50 & 20 \\
CogVideoX & CogVideoX1.5-5B & \texttt{THUDM/CogVideoX1.5-5B} & 3D VAE Latent DiT & $1360 \times 768$ & 50 & 20 \\
LTX-Video & LTX-Video 2B v0.9.0 & \texttt{Lightricks/LTX-Video} & 3D VAE Latent DiT & $768 \times 512$ & 50 & 20 \\
LTX-Video & LTX-Video 13B dev$^\mathrm{T}$ & \texttt{Lightricks/LTX-Video-0.9.7-dev} & 3D VAE Latent DiT & $768 \times 512$ & 50 & 20 \\
Wan & Wan2.1 T2V 1.3B & \texttt{Wan-AI/Wan2.1-T2V-1.3B-Diffusers} & 3D VAE Latent DiT & $832 \times 480$ & 50 & 20 \\
Wan & Wan2.1 T2V 14B$^\mathrm{I}$ & \texttt{Wan-AI/Wan2.1-T2V-14B-Diffusers} & 3D VAE Latent DiT & $832 \times 480$ & 50 & 20 \\
Wan & Wan2.2 T2V A14B$^\mathrm{I}$ & \texttt{Wan-AI/Wan2.2-T2V-A14B-Diffusers} & 3D VAE Latent DiT (MoE) & $832 \times 480$ & 50 & 20 \\
HunyuanVideo & HunyuanVideo$^\mathrm{I}$ & \texttt{hunyuanvideo-community/HunyuanVideo} & 3D VAE Latent DiT & $960 \times 544$ & 50 & 20 \\
\bottomrule
\end{tabular}%
}
\par\vspace{3pt}
{\footnotesize\raggedright HiDream-O1 in figure labels abbreviates HiDream-O1-Image. LTX-Video 2B uses the initial v0.9.0 checkpoint in Diffusers format.\par}
\end{table}

\section{Non-generative baseline details and scoring methods}\label{comparator-baseline-details}

To benchmark generative loss against non-generative alternatives, we evaluate a heterogeneous pool of 12 frozen image encoders alongside two prespecified five-metric IQA batteries (full-reference and no-reference; Table~\ref{tab:comparator-baseline-suite}). All model weights and scoring protocols are held fixed without task-specific fine-tuning.

Encoders were selected by predefined criteria independent of observed human alignment to represent standard anchors (ResNet-50, CLIP, and DINOv2), diverse pretraining objectives, varying model scales (0.024B--8.145B parameters), and within-family scaling pairs (e.g., Qwen3-VL-Embedding 2B vs.\ 8B, PE-Core B/16 vs.\ G/14). The resulting pool consists of 12 frozen encoders spanning supervised classification (ResNet-50; \citealp{He2016-resnet}), self-supervised visual representation learning (DINOv2, DINOv3; \citealp{Oquab2023-dinov2,Simeoni2025-dinov3}), and contrastive image--language alignment (CLIP, SigLIP~2, PE-Core, Jina CLIP~v2; \citealp{Radford2021-clip,Tschannen2025-siglip2,Bolya2025-pe,Koukounas2024-jinaclipv2}). Qwen3-VL-Embedding provides a multimodal embedding baseline trained for representation extraction, using the representation of the last valid input token (\citealp{Li2026-qwen3vlembedding}).

\subsection{Frozen encoder distances}
\label{encoder-layer-readout-protocol}

For model $m$ and block $\ell$, let $\phi_{m\ell}(x)$ denote the feature obtained by applying a fixed model-specific readout to that block's output. For global image representations, the pair score is the unsigned cosine distance between independently extracted representations of the original and modified images:
\begin{equation}
d_{\mathrm{enc}}(x_i^{\mathrm{orig}},x_i^{\mathrm{mod}})
=1-\frac{\phi(x_i^{\mathrm{orig}})^\top\phi(x_i^{\mathrm{mod}})}
{\|\phi(x_i^{\mathrm{orig}})\|_2\|\phi(x_i^{\mathrm{mod}})\|_2}.
\end{equation}
Here $\phi=\phi_{m\ell}$ for the candidate being evaluated. No classifier or readout is fitted to human judgments.

Because the 12 encoders differ in their internal architectures (spanning convolutional feature maps, Vision Transformers with or without class tokens, and multimodal language decoders), extracting comparable representations $\phi_{m\ell}(x)$ requires architecture-specific readout rules. We hold these readout rules fixed across all blocks and tasks. Below, we specify how representations are extracted across three structural groups: models using direct global pooling or unprojected class tokens (ResNet-50, CLIP, and PE-Core); models evaluated via spatially matched patch-token distances (DINOv2 and DINOv3); and models using terminal pooling, projection, or normalization transformations (SigLIP~2, Jina CLIP~v2, and Qwen3-VL-Embedding).

\paragraph{ResNet, CLIP, and PE-Core.}
ResNet-50 uses spatial global average pooling after each of its 16 bottleneck blocks, without the classification head. CLIP uses the block's unprojected \texttt{[CLS]} token, without the terminal LayerNorm or projection. This choice follows the default unnormalized intermediate-feature convention in OpenCLIP, while retaining the original OpenAI CLIP checkpoint. PE-Core uses the spatial mean of block tokens after excluding prefix tokens, without terminal normalization or attention pooling. The PE feature API exposes unnormalized intermediate tokens; their spatial averaging is our fixed aggregation choice. Pooling rules therefore follow the native token structure of each architecture.

\paragraph{DINOv2/v3 spatial distances.}
For DINOv2 and DINOv3, we apply the pretrained terminal LayerNorm to each candidate block, following the normalized intermediate-feature convention. We then exclude the leading \texttt{[CLS]} token and, for DINOv3, four register tokens. The score is the mean one-minus-cosine distance between corresponding spatial patch tokens. Tokens are not averaged before computing cosine distance, and no self-supervised projection head is used. This preserves a spatially matched comparison under the controlled original--modified pairing.

\paragraph{SigLIP, Jina, and Qwen output transformations.}
SigLIP~2 applies its pretrained terminal LayerNorm and learned attention pooling to each block. Jina CLIP~v2 applies its terminal normalization, \texttt{[CLS]} extraction, and projection. Qwen3-VL-Embedding processes images through its full visual tower into a multimodal language decoder (28 blocks for 2B, 36 for 8B). Following its official representation extraction protocol, we apply terminal RMSNorm and $\ell_2$ normalization to the representation of the last valid input token at each decoder block. Both Qwen models use the fixed default instruction ``Represent the user's input.'' without task-specific prompting. These transformations are applied directly to intermediate blocks without retraining.

\paragraph{Candidate layer evaluation.}
To identify the strongest representations achievable by each architecture, we evaluated pair distances across all 314 intermediate backbone blocks alongside each model's default final output (e.g., CLIP's projected embedding or PE-Core's attention-pooled vector). For each task, the representative layer reported in Table~\ref{tab:encoder-selected-layers} was selected by maximizing the signed Pearson correlation with human judgment differences across all stimulus pairs. The downstream protocol for incorporating these selected layers and top-performing models into partial correlation analyses is described in Appendix~\ref{selection-aware-partial-correlation-protocol}.

\begin{table}[ht]
\centering
\small
\caption{\textbf{Selected encoder layers.} Entries indicate the selected layer index and total model depth for each task. Entries marked ``(output)'' use the model's default final embedding rather than intermediate block readouts (Section~\ref{encoder-layer-readout-protocol}). For Qwen, indices denote language-decoder blocks. Selected layers remain fixed across all condition-specific comparisons within each task.}
\label{tab:encoder-selected-layers}
\begin{tabular}{@{}lcc@{}}
\toprule
Encoder & Thatcher & Illumination \\
\midrule
ResNet-50 & $7/16$ & $16/16$ (output) \\
CLIP ViT-B/16 & $12/12$ & $9/12$ \\
DINOv2-Base & $2/12$ & $12/12$ (output) \\
DINOv2-g/14 & $10/40$ & $8/40$ \\
DINOv3-H+/16 & $32/32$ (output) & $1/32$ \\
DINOv3-7B/16 & $36/40$ & $5/40$ \\
SigLIP 2 Base/16 & $9/12$ & $7/12$ \\
PE-Core B/16 & $12/12$ (output) & $11/12$ \\
PE-Core G/14 & $50/50$ (output) & $33/50$ \\
Jina CLIP v2 & $23/24$ & $22/24$ \\
Qwen3-VL-Embedding 2B & $22/28$ & $28/28$ (output) \\
Qwen3-VL-Embedding 8B & $31/36$ & $35/36$ \\
\bottomrule
\end{tabular}
\end{table}

\subsection{Full-reference IQA (FR-IQA)}
Full-reference metrics evaluate the distortion of $x_i^{\mathrm{mod}}$ relative to the pristine original reference $x_i^{\mathrm{orig}}$. To align all FR-IQA metrics along a consistent \emph{degradation / distance} axis where larger values indicate greater unnaturalness or perceptual difference:
\begin{equation}
u_{\mathrm{FR}}(x_i^{\mathrm{orig}}, x_i^{\mathrm{mod}}) = \begin{cases}
- \mathrm{Metric}(x_i^{\mathrm{mod}}, x_i^{\mathrm{orig}}) & \text{if higher is better (PSNR, SSIM, VIF)}, \\
\phantom{-}\mathrm{Metric}(x_i^{\mathrm{mod}}, x_i^{\mathrm{orig}}) & \text{if lower is better (LPIPS-VGG, DISTS)}.
\end{cases}
\end{equation}
Specifically, PSNR is computed from pixel-level mean squared error; SSIM evaluates local luminance, contrast, and structure over sliding windows; VIF models visual information fidelity in the wavelet domain; LPIPS-VGG computes $\ell_2$-distance across five intermediate VGG-16 feature activation layers weighted by BAPPS perceptual judgments; and DISTS evaluates multi-scale structural and texture similarities using spatial deep features.

\subsection{No-reference IQA (NR-IQA)}
No-reference metrics score each image independently without access to a reference image: $q(x)$. To measure directional quality changes on original--modified pairs, scores are differenced strictly according to each metric's published valence:
\begin{equation}
u_{\mathrm{NR}}(x_i^{\mathrm{orig}}, x_i^{\mathrm{mod}}) = \begin{cases}
q(x_i^{\mathrm{orig}}) - q(x_i^{\mathrm{mod}}) & \text{if higher is better (HyperIQA, MUSIQ, LIQE)}, \\
q(x_i^{\mathrm{mod}}) - q(x_i^{\mathrm{orig}}) & \text{if lower is better (NIQE, BRISQUE)}.
\end{cases}
\end{equation}
Under this definition, positive values ($u_{\mathrm{NR}} > 0$) consistently signify that the original image is evaluated as higher quality than the modified image. The battery spans NIQE, an opinion-unaware natural-scene-statistics (NSS) model fitted to pristine luminance images; BRISQUE, spatial NSS features paired with an opinion-trained support vector regressor; HyperIQA, a CNN hypernetwork predicting content-adaptive quality weights; MUSIQ, a multi-scale Vision Transformer trained on KonIQ-10k; and LIQE, a multimodal quality evaluator using CLIP semantic prompts trained on KonIQ.

\subsection{Preprocessing and execution environment}
All input images were resized to $512 \times 512$ pixels without cropping, followed by model-specific channel normalization. IQA metrics used their default processors from PyIQA (v0.1.9)~\citep{Chen2022-pyiqa}. Probing was executed on NVIDIA RTX A6000 GPUs, with dependencies and configurations provided in the supplementary material.

\begin{table}[ht]
\centering
\scriptsize
\caption{\textbf{Evaluated non-generative baseline suite.} Specifications of the 12 frozen vision encoders alongside five full-reference (FR-IQA) and five no-reference (NR-IQA) metrics. Parameter counts (B) report total network parameters in billions. Superscripts $^\mathrm{T}$ and $^\mathrm{I}$ mark the top five encoders selected by overall human alignment on Thatcher and Illumination, respectively; these selections are held fixed across conditions (Section~\ref{condition-specific-partial-correlation}).}
\label{tab:comparator-baseline-suite}
\vspace{4pt}
\setlength{\tabcolsep}{3pt}
\begin{tabular}{@{}>{\raggedright\arraybackslash}p{.07\linewidth}>{\raggedright\arraybackslash}p{.15\linewidth}>{\raggedright\arraybackslash}p{.23\linewidth}>{\raggedright\arraybackslash}p{.17\linewidth}>{\raggedleft\arraybackslash}p{.07\linewidth}@{\hspace{8pt}}>{\raggedright\arraybackslash}p{.21\linewidth}@{}}
\toprule
\textbf{Pool} & \textbf{Predictor} & \textbf{Identifier / checkpoint} & \textbf{Training objective} & \shortstack{\textbf{Params.}\\\textbf{(B)}} & \textbf{Readout / score} \\
\midrule
Encoder & ResNet-50 (ImageNet-1K, torchvision V2) & \path{torchvision::IMAGENET1K_V2} & ImageNet supervision & .024 & Block spatial mean; cosine distance \\
Encoder & CLIP ViT-B/16$^{\mathrm{T},\mathrm{I}}$ & \path{openai/clip-vit-base-patch16} & Image--text contrastive & .086 & Unprojected block CLS; cosine distance \\
Encoder & DINOv2-Base & \path{facebook/dinov2-base} & Vision-only self-supervision & .087 & Corresponding patch-token distance \\
Encoder & SigLIP~2 Base/16$^\mathrm{I}$ (224) & \path{google/siglip2-base-patch16-224} & Image--text sigmoid contrastive & .093 & LayerNorm, attention pooling; cosine distance \\
Encoder & PE-Core B/16$^\mathrm{I}$ (224) & \path{facebook/PE-Core-B16-224} & Vision--language representation & .094 & Block spatial mean; cosine distance \\
Encoder & DINOv2-g/14 spatial & \path{facebook/dinov2-giant} & Vision-only self-supervision & 1.136 & Corresponding patch-token distance \\
Encoder & PE-Core G/14$^{\mathrm{T},\mathrm{I}}$ (448) & \path{timm/vit_pe_core_gigantic_patch14_448.fb} & Vision--language representation & 1.882 & Block spatial mean; cosine distance \\
Encoder & DINOv3-H+/16 spatial$^\mathrm{T}$ & \path{facebook/dinov3-vith16plus-pretrain-lvd1689m} & Vision-only self-supervision & .840 & Corresponding patch-token distance \\
Encoder & DINOv3-7B/16 spatial$^\mathrm{T}$ & \path{facebook/dinov3-vit7b16-pretrain-lvd1689m} & Vision-only self-supervision & 6.716 & Corresponding patch-token distance \\
Encoder & Qwen3-VL-Embedding 2B & \path{Qwen/Qwen3-VL-Embedding-2B} & Multimodal embedding & 2.128 & RMSNorm, last valid token; cosine distance \\
Encoder & Qwen3-VL-Embedding 8B$^\mathrm{T}$ & \path{Qwen/Qwen3-VL-Embedding-8B} & Multimodal embedding & 8.145 & RMSNorm, last valid token; cosine distance \\
Encoder & Jina CLIP v2$^\mathrm{I}$ & \path{jinaai/jina-clip-v2} & Multilingual image--text contrastive & .304 & Terminal norm, CLS projection; cosine distance \\
\midrule
\multicolumn{6}{l}{\textbf{Full-reference IQA}} \\
FR-IQA & PSNR & \path{pyiqa::psnr} & Pixel error & --- & $-\mathrm{PSNR}(x^{\mathrm{mod}},x^{\mathrm{orig}})$ \\
FR-IQA & SSIM & \path{pyiqa::ssim} & Structural similarity & --- & $-\mathrm{SSIM}(x^{\mathrm{mod}},x^{\mathrm{orig}})$ \\
FR-IQA & VIF & \path{pyiqa::vif} & Visual information fidelity & --- & $-\mathrm{VIF}(x^{\mathrm{mod}},x^{\mathrm{orig}})$ \\
FR-IQA & LPIPS-VGG & \path{pyiqa::lpips-vgg} & Learned perceptual distance & --- & $\mathrm{LPIPS}(x^{\mathrm{mod}},x^{\mathrm{orig}})$ \\
FR-IQA & DISTS & \path{pyiqa::dists} & Structure and texture similarity & --- & $\mathrm{DISTS}(x^{\mathrm{mod}},x^{\mathrm{orig}})$ \\
\midrule
\multicolumn{6}{l}{\textbf{No-reference IQA}} \\
NR-IQA & NIQE & \path{pyiqa::niqe} & Natural scene statistics & --- & $q(x^{\mathrm{mod}})-q(x^{\mathrm{orig}})$ \\
NR-IQA & BRISQUE & \path{pyiqa::brisque} & Natural scene statistics + SVR & --- & $q(x^{\mathrm{mod}})-q(x^{\mathrm{orig}})$ \\
NR-IQA & HyperIQA & \path{pyiqa::hyperiqa} & CNN hypernetwork & --- & $q(x^{\mathrm{orig}})-q(x^{\mathrm{mod}})$ \\
NR-IQA & MUSIQ & \path{pyiqa::musiq (koniq10k)} & Multi-scale Transformer & --- & $q(x^{\mathrm{orig}})-q(x^{\mathrm{mod}})$ \\
NR-IQA & LIQE & \path{pyiqa::liqe (koniq)} & Language-prompted IQA & --- & $q(x^{\mathrm{orig}})-q(x^{\mathrm{mod}})$ \\
\bottomrule
\end{tabular}
\end{table}

\clearpage

\section{Statistical procedures and analysis details}
\label{detailed-statistical-methods}
\label{detailed-analysis-protocols}

This section provides methodological details for the statistical inference, reliability estimation, and baseline partial correlation analysis reported in the main text.

\subsection{Bootstrap resampling and statistical uncertainty}
To account for sampling uncertainty across both observers and stimuli, primary human-dependent estimates (such as model--human alignment and human violation sensitivity) are evaluated via a nonparametric two-factor bootstrap ($10{,}000$ draws with fixed random seeds), independently resampling participants and base scenes with replacement. Participants are sampled within each stimulus-assignment group at their original sample sizes. Stimuli are sampled at the level of base scenes: Thatcher stimuli were resampled across the $140$ face identities (keeping upright and inverted pairs together); Illumination stimuli were resampled at the level of the $48$ base configurations within each operation, keeping horizontally mirrored variants together. Generative-model sensitivities do not depend on human ratings and are evaluated using base-scene resampling alone. In each iteration, condition means and the task-wide standard deviation denominator ($s_{m,\mathrm{task}}$) are recomputed dynamically, propagating scaling uncertainty into the resulting estimates. Identical bootstrap draws are shared across all models, conditions, and paired comparisons within each task. Pointwise 95\% confidence intervals correspond to the empirical 2.5th and 97.5th percentiles across draws, summarizing sampling variability across observers and stimuli.

\paragraph{Human factor-profile intervals.}
The error bars for human profile means in Figure~\ref{fig:stimuli-reference-conditioning}B and Appendix~\ref{illumination-factor-profiles} evaluate observer sampling variability conditional on the fixed stimulus set. These pointwise 95\% intervals are constructed from $10{,}000$ participant-bootstrap iterations within each stimulus-assignment group, recomputing image-wise means and task-scaled effects within each draw.

\paragraph{Cross-model associations.}
To evaluate uncertainty in cross-model sensitivity--alignment correlations while accounting for dependence among related models, we combine the participant-and-scene resampling above with a cluster bootstrap over 11 model families (CogVideoX, FLUX.1, FLUX.2, HiDream, HunyuanVideo, JiT, LTX-Video, PixelGen, Qwen-Image, Stable Diffusion, and Wan). In each of $10{,}000$ iterations, 11 families are sampled with replacement, retaining all models within each selected family to recompute the cross-model correlation. Resampled family draws are held consistent across tasks and conditions. Reported 95\% confidence intervals correspond to the pointwise 2.5th--97.5th percentiles across iterations, summarizing joint sampling variability across observers, stimuli, and model families.

\subsection{Human split-half reliability estimation}
\label{human-split-half-reliability-estimation}
To establish an empirical reference for human rating consistency, we evaluate split-half reliability across $10{,}000$ random participant partitions. Within each stimulus-assignment group, participants are randomly divided into two disjoint halves ($A$ and $B$) of equal (or near-equal) size, holding partitions consistent across experimental conditions. For each partition, item-level mean unnaturalness effects ($h_{A,i}$ and $h_{B,i}$) are computed independently for each half, and split-half reliability is evaluated as their Pearson correlation:
\begin{equation}
r_{\mathrm{half}} = \operatorname{corr}_i(h_{A,i}, h_{B,i}).
\end{equation}
In Figures~\ref{fig:cross-model-sensitivity-alignment} and~\ref{fig:comparator-alignment-and-partial-correlation} (and their condition breakdowns, Figures~\ref{fig:condition-specific-sensitivity-alignment} and~\ref{fig:condition-specific-comparator-partial-correlations}), reference lines and markers indicate the mean split-half correlation across partitions on the complete stimulus set. To reflect variability from both rater partitioning and stimulus sampling, each partition is paired with a synchronized base-scene bootstrap draw, with the 2.5th and 97.5th percentiles defining the 95\% resampling intervals (shown as shaded bands and error bars).

\subsection{Bidirectional partial correlation analysis}
\label{selection-aware-partial-correlation-protocol}
To test whether Generative scores retain human-aligned variation beyond strong frozen encoders and quality metrics, we compare the 25 Generative scores with the 12-encoder pool and the two prespecified five-metric IQA batteries described in Section~\ref{baseline-comparators} and Appendix~\ref{comparator-baseline-details}.

\paragraph{Model and layer selection.}
To compare generative models against the strongest non-generative baselines, top-performing models are selected in two steps. First, for each of the 12 frozen encoders, we identify the layer that achieves the highest Pearson correlation with human unnaturalness ratings across the full task (Table~\ref{tab:encoder-selected-layers}). Second, from the 25 generative models and the 12 selected-layer encoders, we rank models by their correlation with human ratings and retain the top five from each pool. All five metrics in each IQA battery are retained without selection. These selected top-five models and encoder layers are held constant across all analyses within each task, including the condition breakdowns and factor profiles (Appendix~\ref{condition-specific-partial-correlation}), ensuring that differences across conditions reflect stimulus properties rather than shifts in model composition.

\paragraph{Bidirectional partial correlations.}
For every selected Generative--comparator pair $(G,X)$, we compute
\begin{equation}
r(h,G\mid X)=
\frac{r_{hG}-r_{hX}r_{GX}}
{\sqrt{(1-r_{hX}^{2})(1-r_{GX}^{2})}},
\qquad
r(h,X\mid G)=
\frac{r_{hX}-r_{hG}r_{GX}}
{\sqrt{(1-r_{hG}^{2})(1-r_{GX}^{2})}}.
\end{equation}
The two directional summaries are the medians of the 25 values in the selected $5\times5$ grid.

\paragraph{Accounting for selection uncertainty in bootstrap intervals.}
To include participant, stimulus, and selection variation, confidence intervals use the same 10,000 participant-and-scene bootstrap draws. On each draw, the two-step selection process is repeated on the resampled full-task data: identifying the best-performing layer for each encoder, and then ranking models to select the top five generative models and encoders. These selections are applied consistently across the overall task and condition breakdowns for that draw. For each comparison, 95\% confidence intervals for the directional partial correlations are obtained from the 2.5th and 97.5th percentiles across the bootstrap distribution.

\section{Diagnostic controls on native generative loss}
\label{diagnostic-controls-generic-distortions-element-deletion}
\label{boundary-controls-csiq-shadow}
\label{one-shadow-deletion-boundary-control}

Generative model loss is heavily sensitive to low-level image complexity and high-frequency statistics~\citep{Nalisnick2018-jh,Serra2020-input}. Consequently, paired loss differences ($L_{\mathrm{modified}} - L_{\mathrm{original}}$) fail to track perceptual naturalness whenever an intervention alters these low-level properties, and unconditioned standalone losses fail to reflect human ratings across differing visual compositions. To evaluate the necessity of content-preserving relational manipulations and paired score differencing, we conducted two diagnostic evaluations: generic low-level distortions alongside element deletion, and standalone single-image loss evaluations.

\subsection{Generic distortions on CSIQ and element deletion}

\begin{figure}[!htb]
\centering
\includegraphics[width=.98\textwidth]{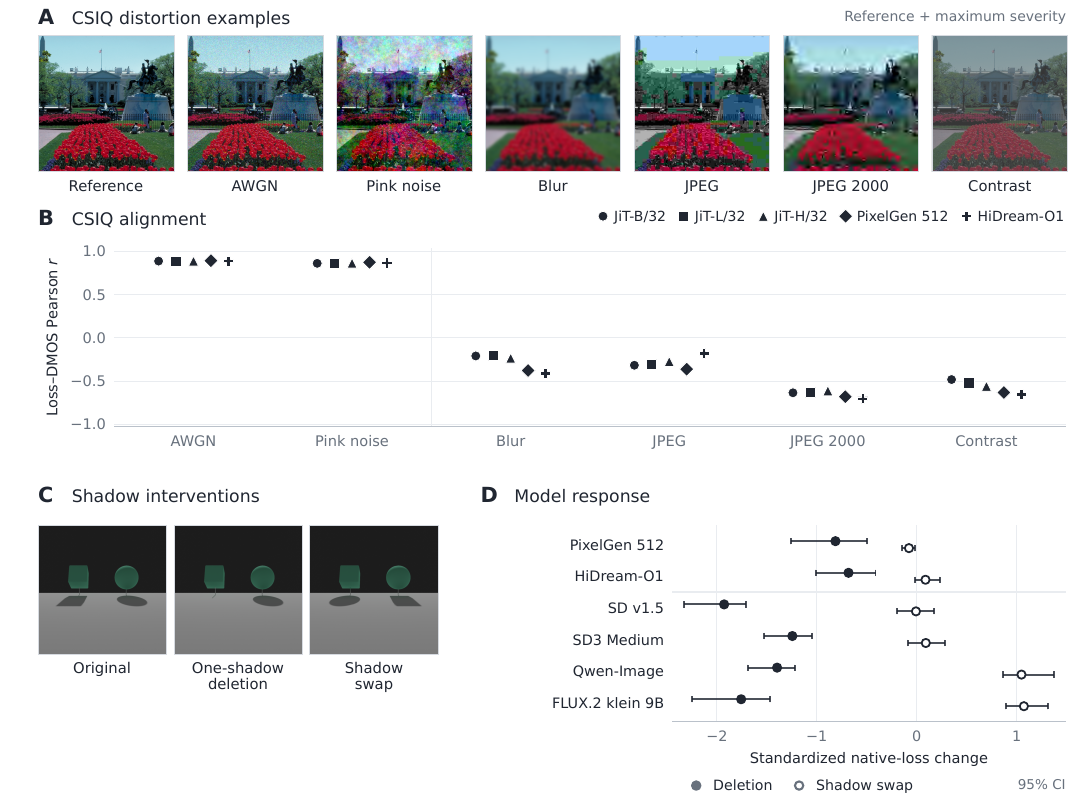}
\caption{\textbf{Impact of low-level image statistics on native generative loss.}
(A) Illustrative CSIQ reference image alongside corrupted examples at maximum severity across six distortion types.
(B) Paired loss difference ($L_{\mathrm{distorted}} - L_{\mathrm{reference}}$)--DMOS Pearson correlations within each CSIQ distortion type across five pixel-space diffusion models.
(C) Example synthetic scenes: original baseline, one-shadow deletion, and shadow swap (relocation).
(D) Mean loss changes ($L_{\mathrm{modified}} - L_{\mathrm{original}}$), divided by their sample standard deviation over the 96 pairs of each manipulation, for one-shadow deletion (filled circles) versus shadow swap (open circles). Error bars denote percentile 95\% intervals from 10,000 resamples of the 48 base scenes.}
\label{fig:csiq-shadow-boundary}
\end{figure}

To test these limitations, we evaluated paired model loss under two scenarios: generic low-level distortions on the CSIQ benchmark~\citep{Larson2010-xm} and element deletion in synthetic 3D scenes (Figure~\ref{fig:csiq-shadow-boundary}).

\paragraph{Generic distortions on CSIQ.}
First, we tested whether loss differences relative to the reference image ($L_{\mathrm{distorted}} - L_{\mathrm{reference}}$) track human quality judgments across standard corruptions using the CSIQ benchmark~\citep{Larson2010-xm} (Figure~\ref{fig:csiq-shadow-boundary}A). We restricted this evaluation to pixel-space diffusion models (three JiT variants, PixelGen 512, and HiDream-O1-Image) to avoid confounding generative priors with VAE compression characteristics.
Evaluating loss--DMOS correlations within each distortion type reveals a severe statistical bias (Figure~\ref{fig:csiq-shadow-boundary}B). For distortions that add high-frequency variance (AWGN and pink noise), model loss correlates positively with perceived degradation. In contrast, for distortions that smooth textures or attenuate contrast (blur, JPEG, JPEG 2000, and contrast reduction), the correlation systematically reverses to negative. Because denoising objectives naturally penalize high-frequency variance while favoring smoother images, paired loss differences cannot overcome unconstrained shifts in low-level statistics, rendering native loss ineffective as a general-purpose quality metric.

\paragraph{Element deletion versus relational relocation.}
Second, we examined whether altering image content undermines the evaluation of physical consistency. Using 96 rendered 3D scenes, we compared model responses under two shadow manipulations (Figure~\ref{fig:csiq-shadow-boundary}C): (1)~\emph{one-shadow deletion}, which removes a cast shadow entirely, introducing a physical anomaly while reducing overall image complexity; and (2)~\emph{shadow swap} (our main paradigm), which relocates shadows between objects, creating physical inconsistency while preserving low-level image statistics.

Standardized loss differences (defined for this control as the mean paired loss difference divided by the sample standard deviation over the 96 pairs) show distinct patterns between the two manipulations (Figure~\ref{fig:csiq-shadow-boundary}D). For one-shadow deletion, loss changes were uniformly \emph{negative} across all models: although removing a shadow makes the scene physically unnatural, the reduced image complexity dominates the loss landscape, causing the model to evaluate the corrupted scene as more probable. In contrast, under shadow swap, preserving low-level statistics mitigates this negative artifact, allowing more capable architectures to exhibit positive loss penalties that reflect sensitivity to the physical inconsistency.

\clearpage

\subsection{Single-image absolute score analysis}
\label{additional-score-analyses}

\begin{figure}[!htbp]
\centering
\includegraphics[width=\linewidth]{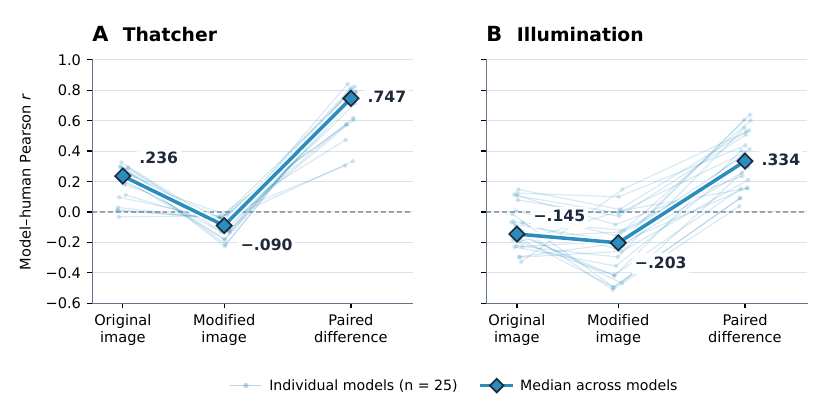}
\caption{\textbf{Single-image losses and paired loss changes in relation to human judgments.}
Item-level Pearson correlations for all 25 generative models, pooled across conditions within (A) Thatcher and (B) Illumination. Original- and modified-image losses are correlated with the corresponding image's mean unnaturalness, $5-R(x)$; paired loss changes are correlated with $R(x^{\mathrm{orig}})-R(x^{\mathrm{mod}})$. Faint lines and small points show individual models, with each line connecting estimates from the same model. Thick lines, diamonds, and numeric labels indicate the median across all 25 models.}
\label{fig:single-image-naturalness}
\end{figure}

Raw single-image model losses ($L_m(x)$ evaluated without reference-image subtraction) do not show consistently positive alignment with human unnaturalness judgments ($a(x) = 5 - R(x)$) across image variants and tasks (Figure~\ref{fig:single-image-naturalness}). Across 280 Thatcher and 288 Illumination pairs, we correlated each original- or modified-image loss with the rating of that same image. The 25-model median Pearson correlations were $.236$ (original) and $-.090$ (modified) on Thatcher faces, and $-.145$ (original) and $-.203$ (modified) on Illumination scenes. Paired score differences ($u_{m,i} = L_m(x_i^{\mathrm{mod}}) - L_m(x_i^{\mathrm{orig}})$) correlated positively with human paired differences ($h_i = R(x_i^{\mathrm{orig}}) - R(x_i^{\mathrm{mod}})$) for all 25 models in both tasks, with median correlations of $.747$ on Thatcher and $.334$ on Illumination. 

\clearpage

\section{Model-wise sensitivity and human alignment}
\label{model-wise-condition-sensitivity-breakdown}

\begin{figure}[!htbp]
\centering
\includegraphics[width=\linewidth]{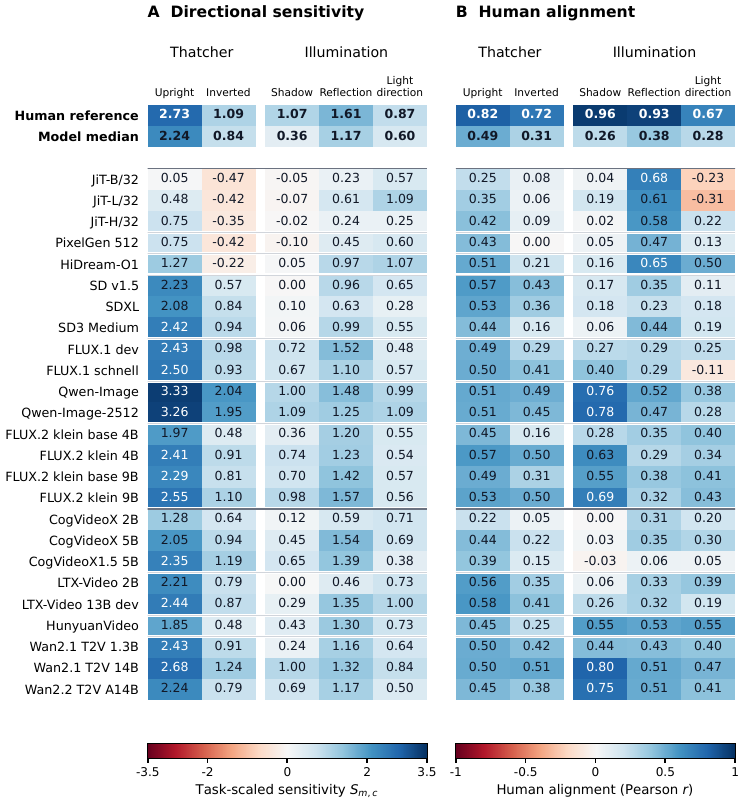}
\caption{\textbf{Model-wise sensitivity and human alignment across conditions.}
(A) Violation sensitivity, $S_{m,c} = \bar u_{m,c} / s_{m,\mathrm{task}}$: the condition mean loss difference divided by the model's sample SD over the full task.
(B) Human alignment: Pearson correlation between model loss differences and human paired rating differences within each condition. The top rows show human references and the median across all 25 models; human alignment is the mean raw split-half correlation across 10,000 disjoint participant splits.}
\label{fig:condition-sensitivity-heatmap-supplement}
\end{figure}

Figure~\ref{fig:condition-sensitivity-heatmap-supplement} shows violation sensitivity and human alignment for all 25 generative models across the five stimulus conditions. Appendix~\ref{condition-specific-sensitivity-alignment} shows the relationship between these two measures as condition-specific scatter plots.

\section{Sensitivity and human alignment across individual conditions}
\label{condition-specific-sensitivity-alignment}

\begin{figure}[t]
\centering
\includegraphics[width=\linewidth]{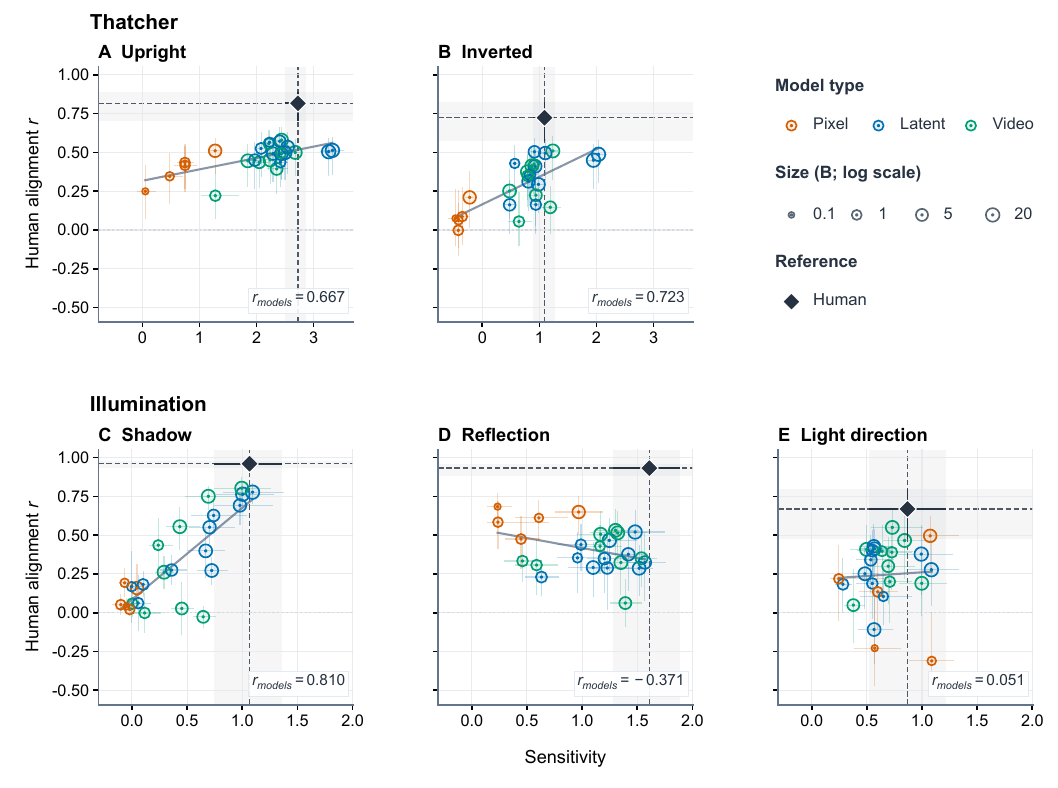}
\caption{\textbf{Sensitivity and human alignment across stimulus conditions.}
Violation sensitivity versus human alignment ($r$) for 25 generative models across individual conditions: upright Thatcher (A), inverted Thatcher (B), cast shadow (C), surface reflection (D), and light direction (E).
Error bars and shaded bands indicate 95\% bootstrap intervals (Appendix~\ref{detailed-statistical-methods}).
Black diamonds and dashed crosshairs mark human sensitivity and mean split-half reliability.
Solid gray lines show linear fits across models.}
\label{fig:condition-specific-sensitivity-alignment}
\end{figure}

Figure~\ref{fig:condition-specific-sensitivity-alignment} shows sensitivity--alignment associations across the 25 models for each stimulus condition. Pointwise 95\% intervals combine family, participant, and scene resampling (Appendix~\ref{detailed-statistical-methods}).

Cross-model sensitivity--alignment associations varied across stimulus conditions (Figure~\ref{fig:condition-specific-sensitivity-alignment}). Positive associations were robust for inverted faces ($r = .723$, 95\% CI $[.118,.923]$) and cast shadows ($r = .810$, $[.424,.960]$), where models with greater violation sensitivity also exhibited closer alignment with continuous human ratings. In contrast, this relationship became weak or inconsistent in the remaining conditions: while upright faces showed a positive point estimate ($r = .667$, $[-.215,.904]$), its confidence interval was wide, and associations were absent or negative for light direction ($r = .051$, $[-.417,.570]$) and surface reflections ($r = -.371$, $[-.704,.364]$). %

\section{Condition-specific evaluation against non-generative baselines}
\label{condition-specific-partial-correlation}
\label{primary-comparator-partial-correlation}

\begin{figure}[!b]
\centering
\includegraphics[width=\linewidth]{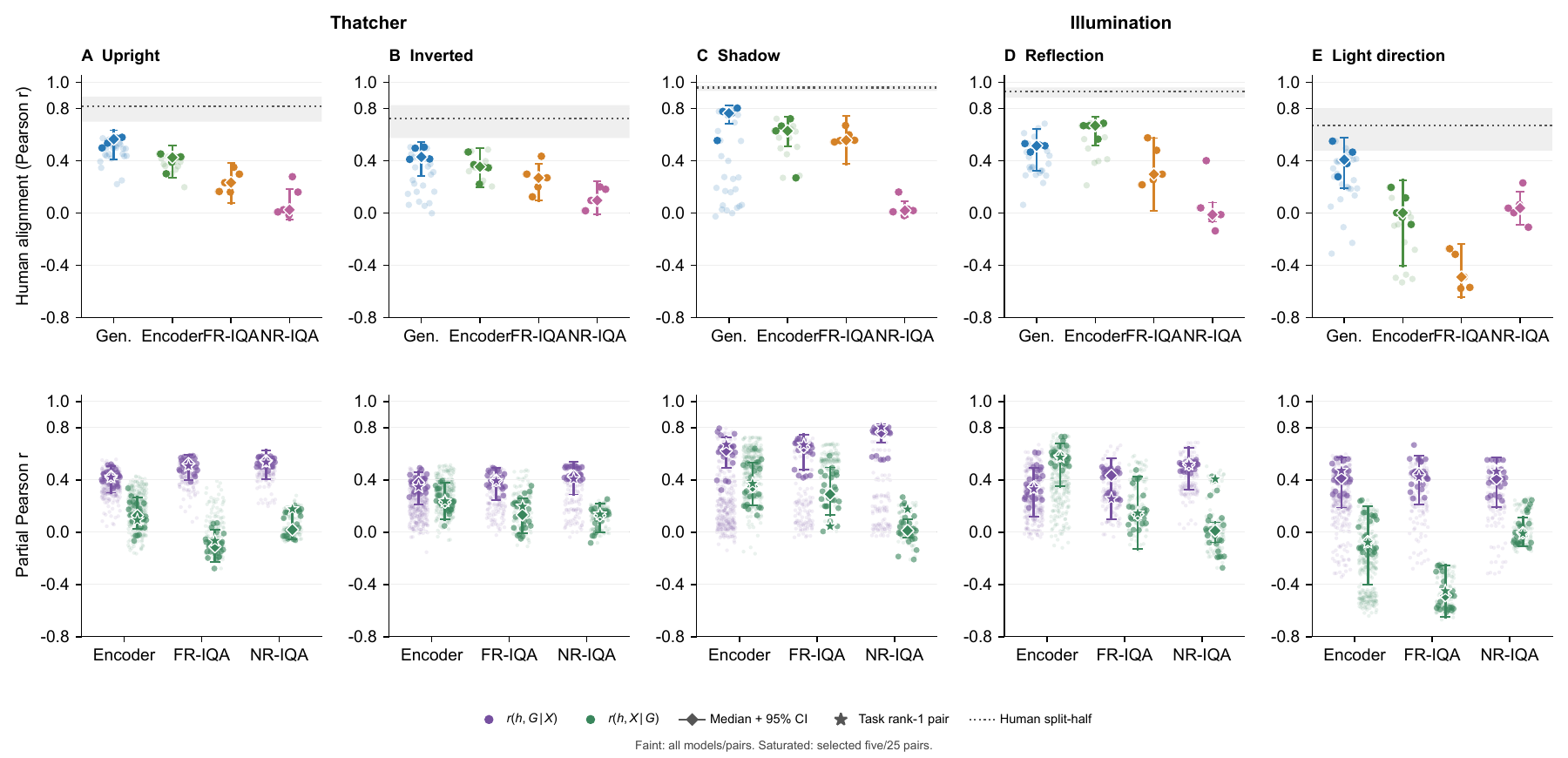}
\caption{\textbf{Condition-level alignment and partial correlation with baseline models.}
Panels show Thatcher (\textbf{A}, upright; \textbf{B}, inverted) and Illumination (\textbf{C}, shadow; \textbf{D}, reflection; \textbf{E}, light direction), evaluated with the top-five models and encoder layers selected over each full task.
Top panels show human alignment (faint points: all candidates; saturated points: selected models; diamonds and error bars: selected median and 95\% bootstrap CI; dotted lines and bands: mean human split-half reliability and its 95\% interval).
Bottom panels show bidirectional partial correlations (faint points: all pairs; saturated points: selected $5\times5$ pairs; stars: rank-1 pair; diamonds and bars: selected median and 95\% CI). Purple circles show $r(h,G\mid X)$ and green circles show $r(h,X\mid G)$.
Bootstrap intervals account for task-level layer and model reselection (Appendix~\ref{selection-aware-partial-correlation-protocol}).}
\label{fig:condition-specific-comparator-partial-correlations}
\end{figure}

We evaluate baseline comparisons separately within each stimulus condition. To facilitate direct comparison without confusing model shifts across conditions, we retain the models selected over each entire task: the top five generative models, top five encoders, and all IQA metrics. Figure~\ref{fig:condition-specific-comparator-partial-correlations} displays the resulting human alignment and bidirectional partial correlations.

Across individual conditions, generative models consistently retained positive residual correlations with human judgments ($r(h,G\mid X)$) after controlling for each non-generative baseline (Figure~\ref{fig:condition-specific-comparator-partial-correlations}, bottom panels). For upright faces, cast shadows, and light direction, generative scores accounted for substantial human-aligned variance beyond visual encoders and standard quality metrics, whereas baseline residual associations were comparatively low. Under face inversion, residual correlations were attenuated for both generative models and baselines, consistent with the overall reduction in perceptual sensitivity. For surface reflections, both generative models and visual encoders retained positive residual associations ($r(h,G\mid\mathrm{Encoder}) \approx .346$, $r(h,\mathrm{Encoder}\mid G) \approx .584$), indicating that both representations capture distinct aspects of reflection naturalness.

Because condition-level aggregates combine multiple scene properties, they do not isolate which factors drive these agreements and discrepancies. In Appendix~\ref{illumination-factor-profiles}, we break down the Illumination stimuli into specific scene factors (including object shape, orientation, and surface roughness) to examine differences across model classes.

\clearpage

\section{Factor-level structure of Illumination judgments}
\label{illumination-factor-profiles}

While overall evaluations measure aggregate alignment across the full stimulus set, they obscure which specific scene properties accompany human--model consistency or divergence. We therefore examined how perceived unnaturalness and model predictions vary across the object shapes, rotations, colors, and physical parameters of the Illumination stimuli (Appendix~\ref{stimulus-construction}). We evaluate the task-pooled top-five models (Section~\ref{baseline-comparators}): generative models Wan2.1 T2V 14B, Qwen-Image, Wan2.2 T2V A14B, Qwen-Image-2512, and HunyuanVideo (Table~\ref{tab:generative-model-suite}); and encoders PE-Core B/16, PE-Core G/14, CLIP ViT-B/16, SigLIP 2 Base/16, and Jina CLIP v2 (Table~\ref{tab:comparator-baseline-suite}). All five metrics from each IQA battery are also retained. The same model sets are used in the condition-level comparison in Figure~\ref{fig:condition-specific-comparator-partial-correlations}.

\subsection{Factor profiles and comparison metric}

For each of the 13 scene factors (Appendix~\ref{stimulus-construction}), a profile plots the mean paired response at each factor level, marginalizing over other scene parameters. Each human or model score is divided by its own sample standard deviation across all 288 Illumination pairs ($\mathrm{ddof}=1$), without subtracting its mean; factor-level means therefore retain the original zero point ($S = \bar{u} / s_{\mathrm{task}}$, Section~\ref{outcome-measures}). FR-IQA metrics are converted to distance forms before task-SD scaling: PSNR becomes unit-range pixel MSE ($10^{-\mathrm{PSNR}/10}$), SSIM and VIF become $1-\mathrm{SSIM}$ and $1-\mathrm{VIF}$, and LPIPS and DISTS retain their native distance values. To quantify human--model agreement, we compute the mean squared error (MSE) between model and human profiles:
\begin{equation}
 \mathrm{MSE}_{mf} = \frac{1}{K_f} \sum_{\ell=1}^{K_f} (y_{mf\ell} - y_{hf\ell})^2,
 \qquad
 \mathrm{MSE}_m = \sum_f v_f \mathrm{MSE}_{mf},
 \label{eq:illumination-profile-mse}
\end{equation}
where $y_{mf\ell}$ and $y_{hf\ell}$ are the scaled model and human means at level $\ell$, $K_f$ is the number of levels in factor $f$, and weights $v_f$ weight each of the three operations equally ($1/3$ per operation, divided evenly among its factors). Figures~\ref{fig:illumination-factor-shadow}--\ref{fig:illumination-factor-light} display the level-wise profiles, and Figure~\ref{fig:illumination-factor-disagreement} summarizes the resulting MSE scores. Human profile means include 95\% participant bootstrap intervals (Appendix~\ref{detailed-statistical-methods}), while category summaries show medians across the five models in each class, displaying model spread rather than confidence intervals.

\subsection{Human perceptual structure and model alignment across scene factors}

\begin{figure}[p]
\centering
\includegraphics[width=\textwidth]{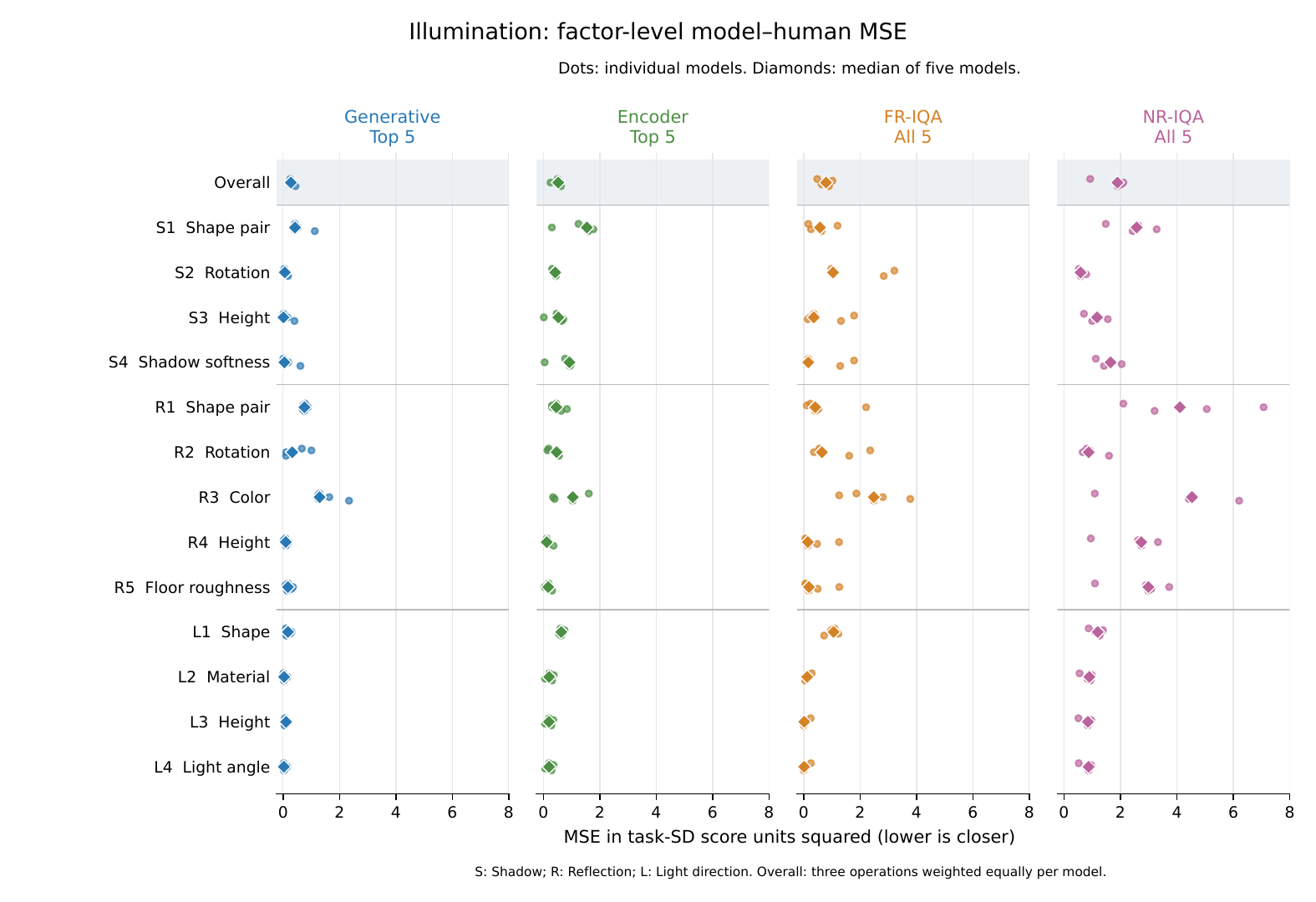}
\caption{\textbf{Factor-profile errors across model categories.}
Columns separate categories and rows separate factors. Dots show individual models (five per category: generative models, visual encoders, FR-IQA, and NR-IQA); diamonds indicate category medians. Models and encoder layers are fixed by task-pooled alignment. The spread describes model variation, not confidence intervals.
Overall displays the aggregate MSE across all 13 factor profiles ($\mathrm{MSE}_m$, Eq.~\ref{eq:illumination-profile-mse}) for each model.
Factor codes correspond to the detailed profiles in Figures~\ref{fig:illumination-factor-shadow}--\ref{fig:illumination-factor-light}.}
\label{fig:illumination-factor-disagreement}
\end{figure}

\begin{figure}[p]
\centering
\includegraphics[width=\textwidth]{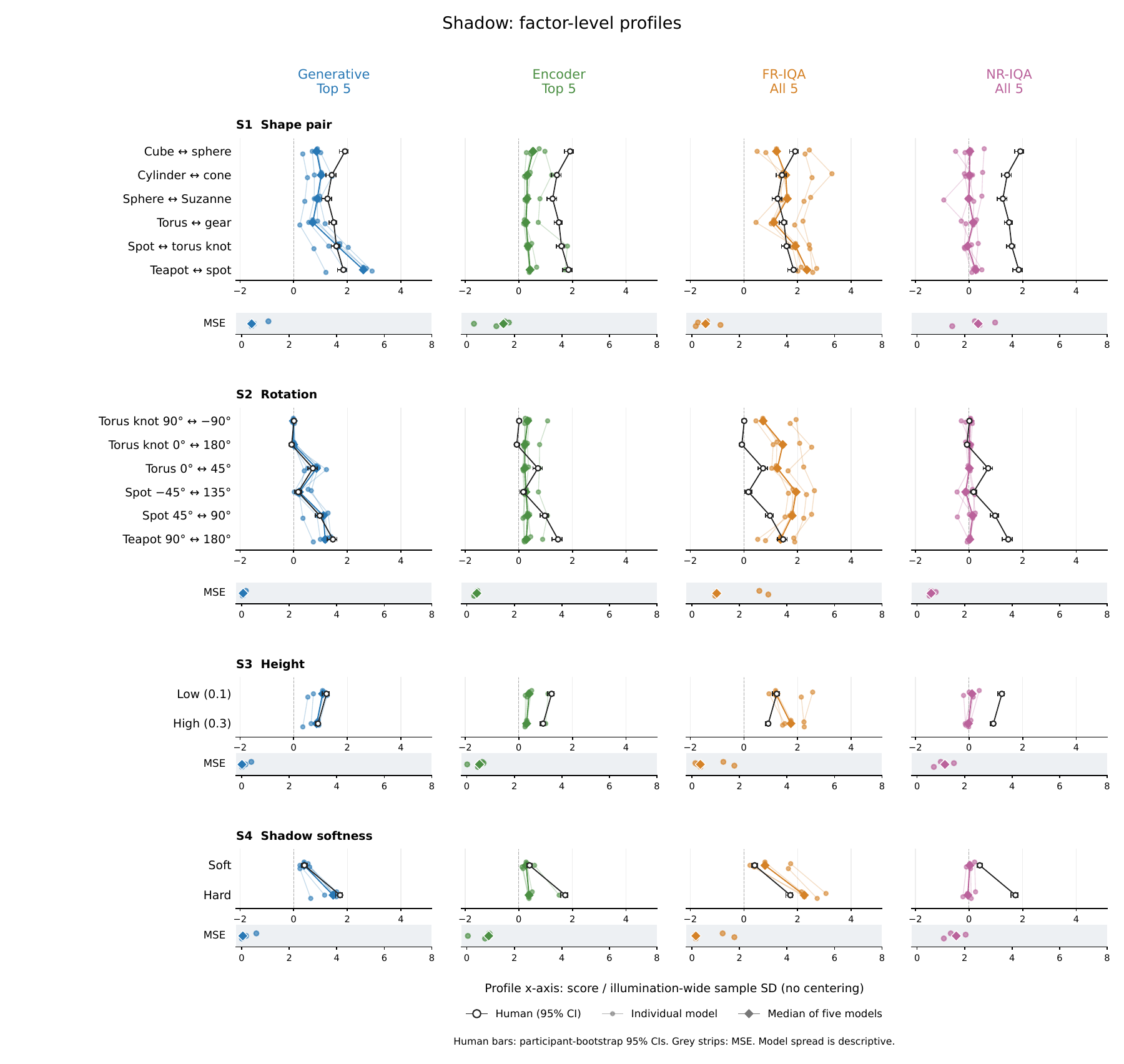}
\caption{\textbf{Shadow: factor profiles and model error.}
Columns separate the four model categories (generative models, visual encoders, FR-IQA, and NR-IQA).
Black open circles and horizontal bars show human means and 95\% bootstrap intervals (Appendix~\ref{detailed-statistical-methods}), repeated across columns. Colored dots show individual models, faint lines connect the same model across factor levels, and filled diamonds denote category medians.
Gray strips show corresponding factor MSEs. Responses are scaled by task standard deviation without centering (encoder distances are unsigned), with model sets and encoder layers held constant across factors.}
\label{fig:illumination-factor-shadow}
\end{figure}

\begin{figure}[p]
\centering
\includegraphics[width=\textwidth]{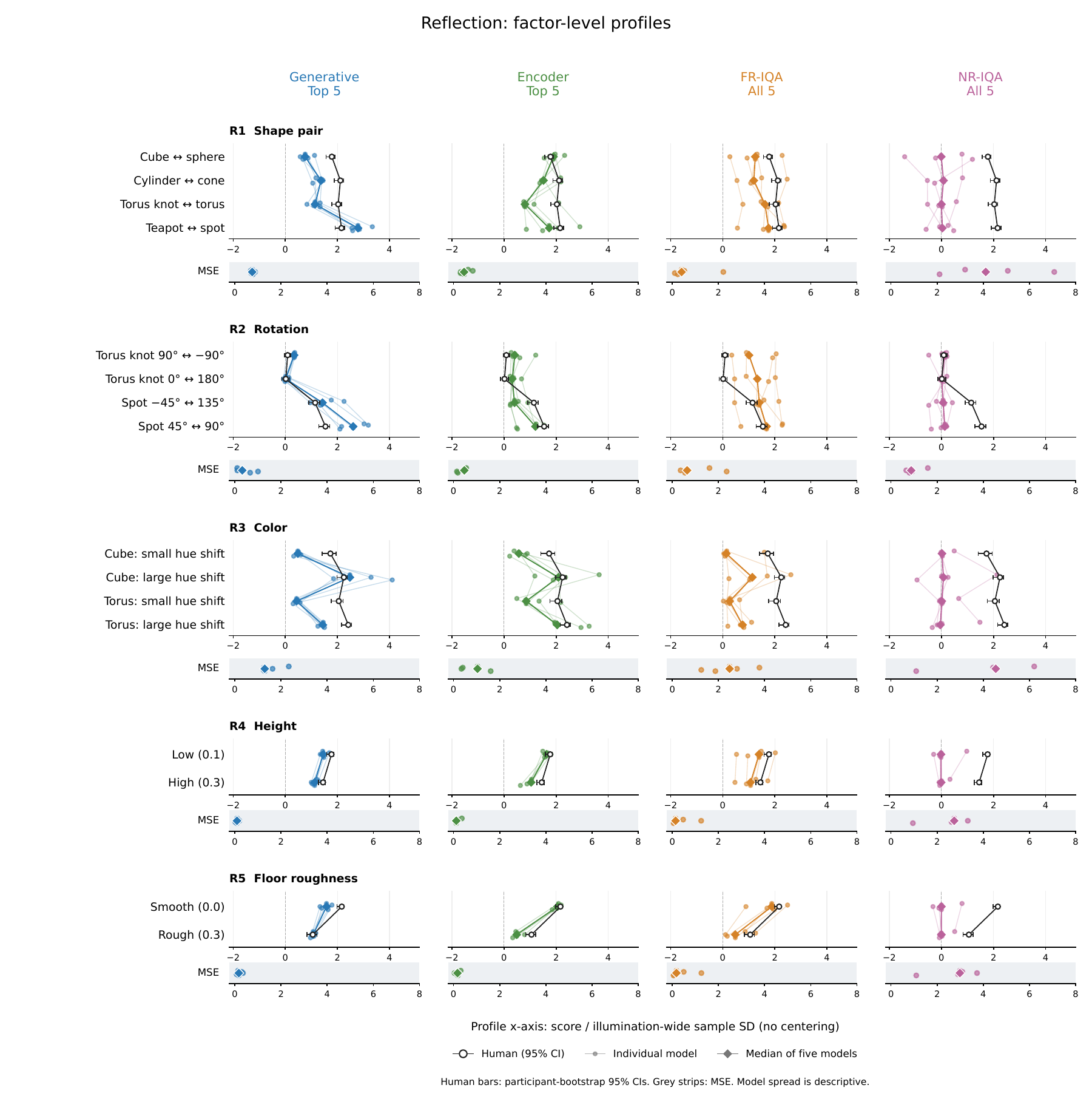}
\caption{\textbf{Reflection: factor profiles and model error.}
Plot conventions match Figure~\ref{fig:illumination-factor-shadow}.}
\label{fig:illumination-factor-reflection}
\end{figure}

\begin{figure}[p]
\centering
\includegraphics[width=\textwidth]{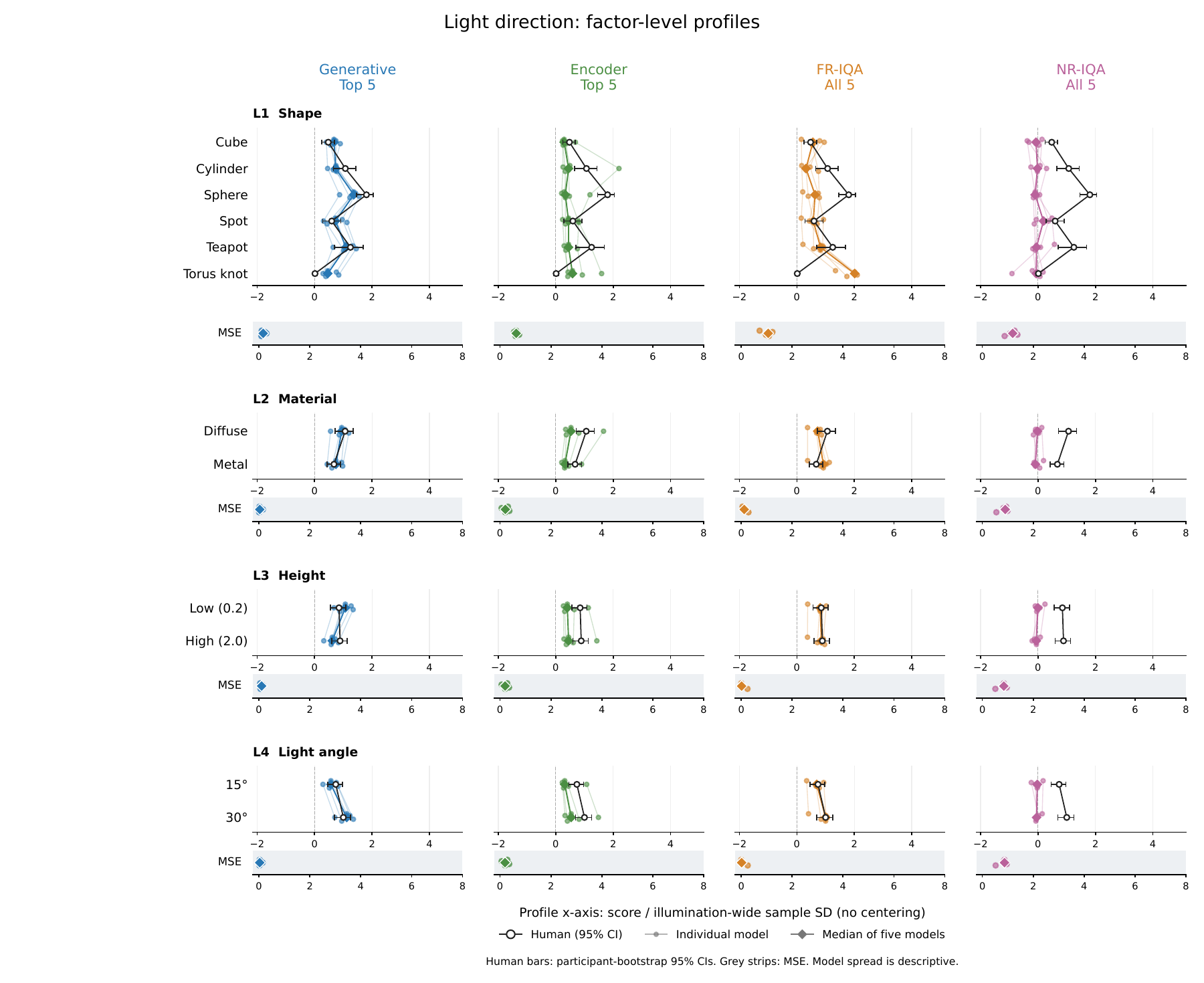}
\caption{\textbf{Light direction: factor profiles and model error.}
Plot conventions match Figure~\ref{fig:illumination-factor-shadow}.}
\label{fig:illumination-factor-light}
\end{figure}

\paragraph{Geometric transformations and rotation across shadow and reflection (S2, R2, L1).}
Human sensitivity to inconsistencies was strongly modulated by object geometry. In both cast shadows and surface reflections, rotating objects with distinct, familiar axes produced highly salient unnaturalness: Shadow scores reached $0.714$ for the torus and $1.462$ for the teapot (S2), while Spot cow rotations in Reflection elicited similarly marked unnaturalness ($1.145$ and $1.540$; R2). In Shadow, unnaturalness for the Spot cow was much higher when rotations introduced salient 2D contour asymmetry ($45^\circ\leftrightarrow90^\circ$: $0.973$) than when the 2D silhouette was near mirror-symmetric ($-45^\circ\leftrightarrow135^\circ$: $0.170$). In contrast, rotating the complex torus knot produced negligible perceived unnaturalness in both Shadow ($-0.077$ and $0.013$) and Reflection ($0.018$ and $0.089$). Similarly, in Light direction (L1), mirroring the central sphere elicited strong unnaturalness ($1.805$), whereas the same manipulation on the torus knot was virtually unnoticed ($0.014$). Generative models closely tracked this selective sensitivity across both shadows and reflections (median $\mathrm{MSE} = 0.061$ in Shadow rotation, $0.324$ in Reflection rotation, and $0.170$ in Light direction; Figure~\ref{fig:illumination-factor-disagreement}). As shown in Figures~\ref{fig:illumination-factor-shadow}--\ref{fig:illumination-factor-light}, generative scores reproduced the distinction between subtle torus-knot rotations and conspicuous teapot or Spot rotations in both shadow and reflection, as well as the difference between the symmetric and asymmetric Spot shadow configurations. Visual encoders ($\mathrm{MSE} = 0.418$--$0.632$) and FR-IQA metrics ($0.645$--$1.058$), by contrast, showed larger discrepancies, responding strongly to torus-knot rotations and lighting changes despite human judgments being near baseline.

\paragraph{Shape swaps and reflection color shifts (S1, R1, R3).}
In contrast to rotation, human judgments of shape swaps showed consistently salient unnaturalness across all tested pairs in both Shadow (S1: $1.261$--$1.913$) and Reflection (R1: $1.784$--$2.155$). Generative models failed to capture this uniform human intolerance: rather than remaining consistently high, median generative scores varied widely across object geometries in both Shadow ($0.711$ for Torus $\leftrightarrow$ gear to $2.597$ for Teapot $\leftrightarrow$ Spot) and Reflection ($0.761$ for Cube $\leftrightarrow$ sphere to $2.789$ for Teapot $\leftrightarrow$ Spot; Figures~\ref{fig:illumination-factor-shadow}--\ref{fig:illumination-factor-reflection}). 
A distinct profile divergence also appeared for reflection color shifts (R3). Whereas human observers judged subtle hue discrepancies as already markedly unnatural ($1.730$ on Cube, $2.052$ on Torus) with only modest increases for large shifts ($2.247$ and $2.416$), generative models scaled strongly with the magnitude of chromatic discrepancy—remaining largely insensitive to subtle hue deviations ($0.480$ and $0.431$) but surging for large shifts (reaching $2.479$ on Cube). This contrast between threshold-like human sensitivity and continuous, magnitude-scaling generative losses produced the largest generative profile error across all factors (median $\mathrm{MSE} = 1.292$ vs.\ encoder $1.037$). In Reflection shape swaps (R1) as well, encoders aligned more closely with uniform human elevations ($\mathrm{MSE} = 0.456$ vs.\ generative $0.753$). These factor-level discrepancies suggest why visual encoders retained substantial residual predictive power specifically for surface reflections (Figure~\ref{fig:condition-specific-comparator-partial-correlations}).

\paragraph{Physical scene parameters (S3--S4, R4--R5, L2--L4).}
Physical scene parameters also systematically affected human judgments and model predictions.
In cast shadows, human perceived unnaturalness was higher for hard than soft shadows (S4: $1.731$ vs.\ $0.400$) and when objects were placed near the ground (S3: $1.224$ vs.\ $0.908$). Generative models closely tracked both contrasts (median $\mathrm{MSE} = 0.043$ for softness and $0.011$ for height), whereas visual encoders exhibited larger errors ($0.922$ and $0.530$).
Similarly, in surface reflections, human ratings decreased from smooth to rough floors (R5: $2.166$ vs.\ $1.054$) and from low to high object positions (R4: $1.773$ vs.\ $1.447$). All model classes captured these reductions with comparable accuracy ($\mathrm{MSE} \approx 0.09$--$0.18$).
In Light direction, human perceived unnaturalness for lighting inconsistencies was higher on diffuse than metallic surfaces (L2: $1.063$ vs.\ $0.675$) and higher for lateral lighting ($30^\circ$) than overhead lighting ($15^\circ$; L4: $1.001$ vs.\ $0.736$), contrasts that generative models matched with low error ($\mathrm{MSE} \le 0.035$). Finally, for object height in Light direction (L3), height also determined whether floor cast shadows were rendered as a reference cue (present at low height, disabled at high height; Appendix~\ref{stimulus-construction}). In this condition, human ratings showed little difference between heights ($0.850$ vs.\ $0.887$), whereas generative models scored higher when cast shadows were present ($1.070$ vs.\ $0.615$, median $\mathrm{MSE} = 0.104$; Figure~\ref{fig:illumination-factor-light}).

\clearpage

\clearpage

\section{Effects of model capacity on sensitivity and human alignment}
\label{model-size-diagnostic}

\begin{figure}[t]
\centering
\includegraphics[width=\linewidth]{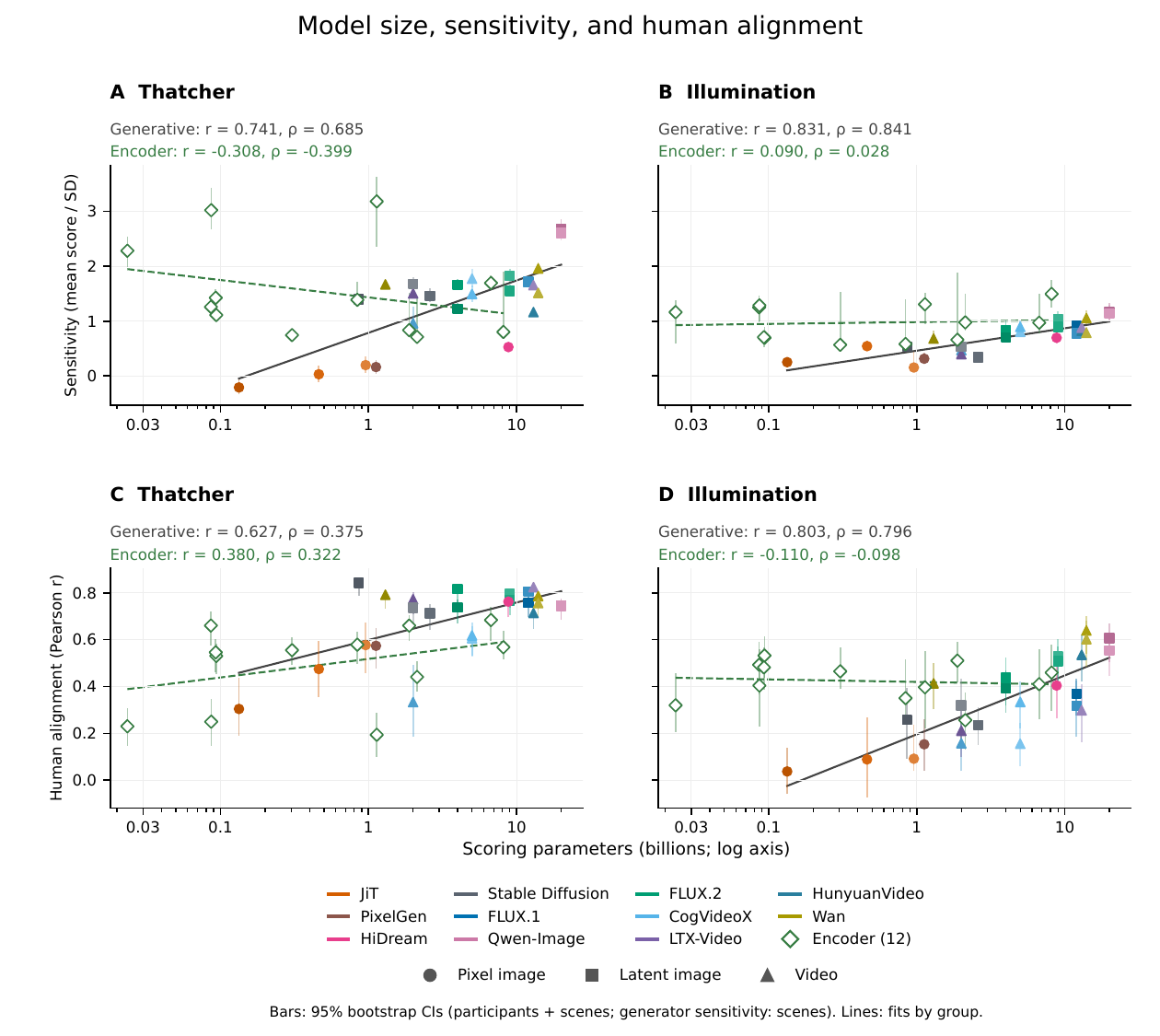}
\caption{\textbf{Model size, sensitivity, and human alignment.}
(\textbf{A, B}) Violation sensitivity and (\textbf{C, D}) human alignment ($r$) versus scoring parameter count for 25 generators and 12 encoders.
Solid and dashed lines show linear fits against log parameter count for generators and encoders, respectively; annotations report Pearson $r$ and Spearman $\rho$.
Error bars indicate 95\% bootstrap intervals (Appendix~\ref{detailed-statistical-methods}).
Colors and markers match Figure~\ref{fig:cross-model-sensitivity-alignment} and the bottom legend.}
\label{fig:model-size-sensitivity-alignment}
\end{figure}

We evaluated how violation sensitivity and human alignment relate to model capacity across the 25 generative models (Figure~\ref{fig:model-size-sensitivity-alignment}). For generative models, model size is defined as the parameter count of the denoising or flow backbone network used to compute native scores (ranging from 0.14B to 20B parameters).

Larger generators generally tended to show both stronger violation sensitivity and closer human alignment, though these associations varied across tasks and metrics. Log parameter count correlated positively with violation sensitivity (Thatcher: Pearson $r = .741$, Spearman $\rho = .685$; Illumination: $r = .831$, $\rho = .841$) and human alignment (Thatcher: $r = .627$, $\rho = .375$; Illumination: $r = .803$, $\rho = .796$). The weaker rank correlation for Thatcher alignment ($\rho = .375$) suggests that while larger models were generally more sensitive, human alignment did not follow model size equally closely across all evaluated models.

To provide an empirical reference over an overlapping parameter range, we also evaluated all 12 frozen vision and multimodal encoders at the layer selected for each task (Table~\ref{tab:encoder-selected-layers}; 0.02B to 8.1B parameters; Figure~\ref{fig:model-size-sensitivity-alignment}, open green diamonds). Encoder size denotes the full scoring network, not the prefix ending at the selected layer. The same standardization (mean feature distance divided by its standard deviation across the task) was applied to unsigned distances, which measure representational divergence rather than directional unnaturalness. At the selected layers, encoder sensitivity was moderately negatively associated with log parameter count on Thatcher and weakly associated on Illumination (Thatcher: $r=-.308$, $\rho=-.399$; Illumination: $r=.090$, $\rho=.028$). Encoder human alignment was positively associated with size on Thatcher ($r=.380$, $\rho=.322$), but weakly associated on Illumination ($r=-.110$, $\rho=-.098$). Thus, whereas larger generators show both stronger violation sensitivity and closer human alignment, scaling encoder size does not automatically improve either metric, particularly on Illumination.

\begin{figure}[p]
\centering
\includegraphics[width=\linewidth]{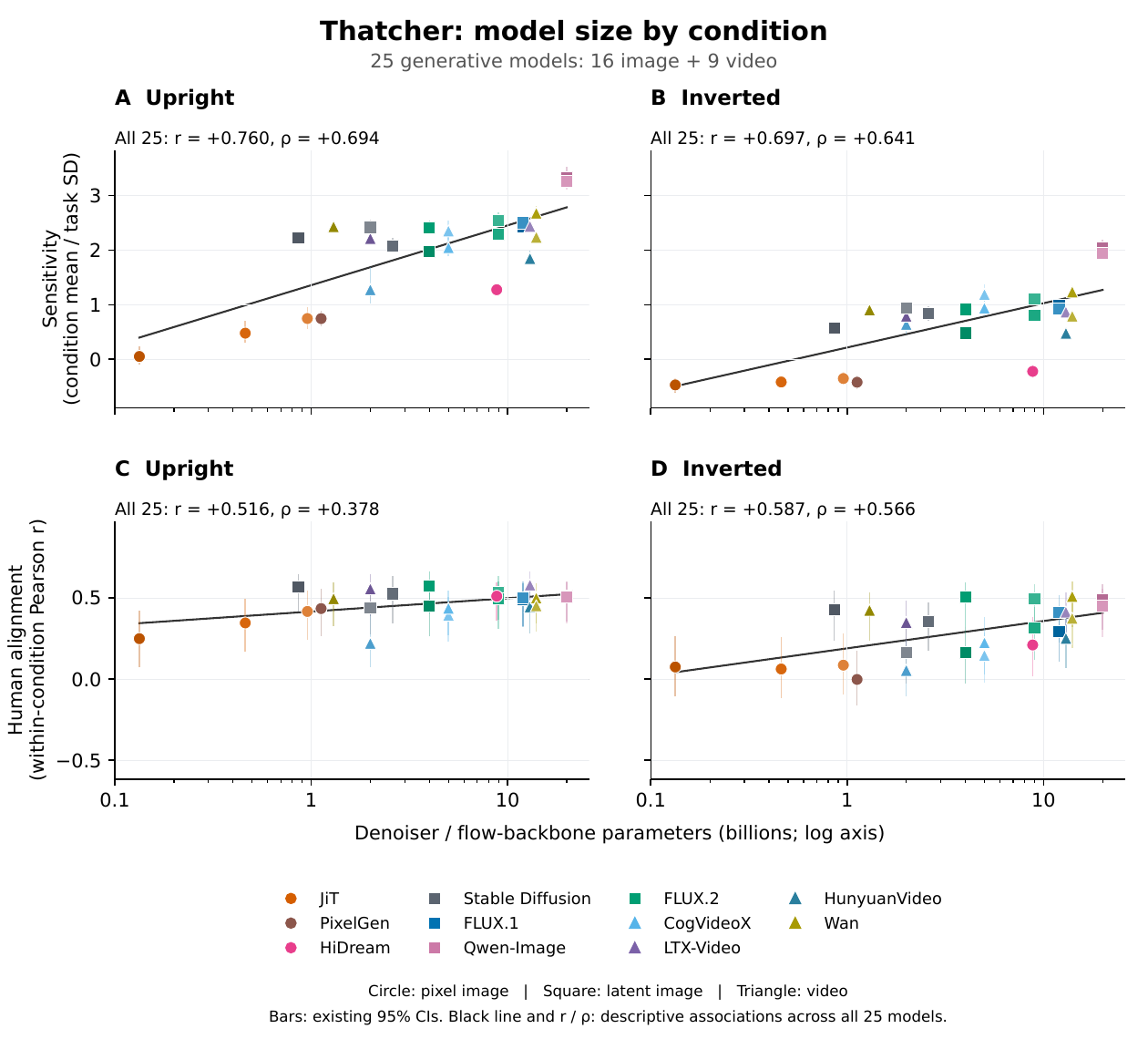}
\caption{\textbf{Model size and condition-specific metrics in the Thatcher task.}
(\textbf{A, B}) Violation sensitivity and (\textbf{C, D}) human alignment ($r$) versus scoring parameter count for upright and inverted faces across 25 generators.
Solid lines show linear fits against log parameter count; annotations report Pearson $r$ and Spearman $\rho$.
Error bars indicate 95\% bootstrap intervals (Appendix~\ref{detailed-statistical-methods}).
Colors and markers match Figure~\ref{fig:cross-model-sensitivity-alignment} and the bottom legend.}
\label{fig:model-size-conditions-thatcher}
\end{figure}

\begin{figure}[p]
\centering
\includegraphics[width=\linewidth]{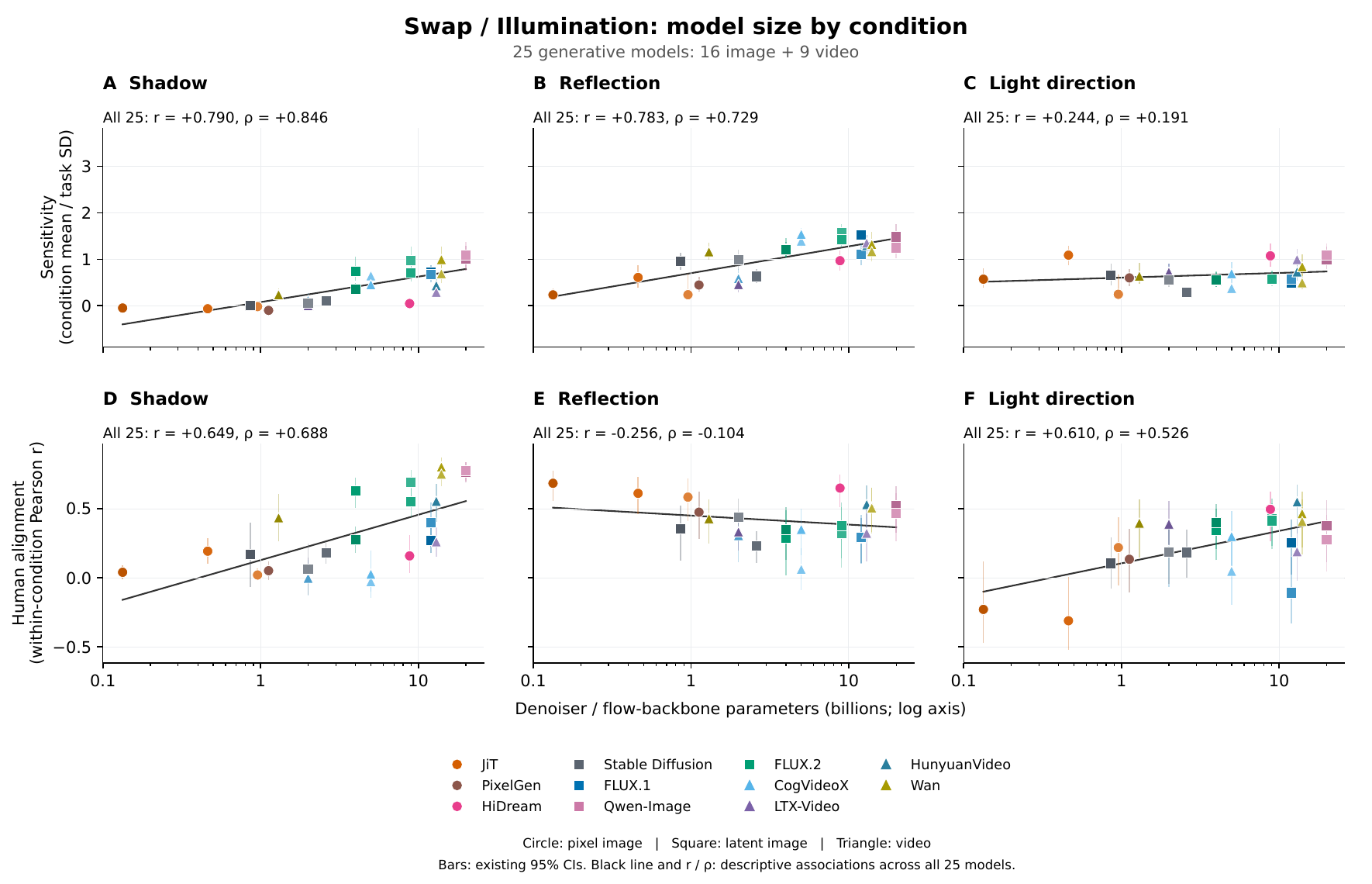}
\caption{\textbf{Model size and condition-specific metrics in the Illumination task.}
(\textbf{A--C}) Violation sensitivity and (\textbf{D--F}) human alignment ($r$) versus scoring parameter count for cast shadow, surface reflection, and light direction across 25 generators.
Solid lines show linear fits against log parameter count; annotations report Pearson $r$ and Spearman $\rho$.
Error bars indicate 95\% bootstrap intervals (Appendix~\ref{detailed-statistical-methods}).
Colors and markers match Figure~\ref{fig:model-size-conditions-thatcher} and the bottom legend.}
\label{fig:model-size-conditions-illumination}
\end{figure}

\paragraph{Condition-specific scaling trajectories.}
Decomposing this relationship by stimulus condition (Figures~\ref{fig:model-size-conditions-thatcher} and~\ref{fig:model-size-conditions-illumination}) reveals that model capacity affects violation sensitivity and human alignment through distinct trajectories. Violation sensitivity scaled positively with parameter count across nearly all conditions (Thatcher upright: Pearson $r = .760, \rho = .694$; inverted: $r = .697, \rho = .641$; cast shadow: $r = .790, \rho = .846$; surface reflection: $r = .783, \rho = .729$; light direction: $r = .244, \rho = .191$), indicating that larger backbones systematically acquire stronger penalties for relational alterations across diverse visual domains. In contrast, human alignment exhibited pronounced domain selectivity: while alignment scaled with capacity for cast shadows ($r = .649, \rho = .688$) and light direction ($r = .610, \rho = .526$), it showed no positive association with model size for surface reflections ($r = -.256, \rho = -.104$) and saturated early for upright faces ($r = .516, \rho = .378$).

Crucially, the absence of alignment scaling for surface reflections cannot be attributed to human rating noise, as participants exhibited high split-half reliability in this condition ($r = .932$; Appendix~\ref{human-split-half-reliability-estimation}). Instead, it reflects a genuine divergence between the statistical features captured by generative loss minimization (which readily penalizes surface modifications) and the physical or perceptual criteria humans employ to judge reflection naturalness.

\clearpage

\clearpage

\clearpage

\section{Published generation evaluations and human alignment}
\label{external-generation-evaluation}
\label{external-generation-size-adjustment}
\label{external-generation-correlation-difference}

We conducted a post-hoc comparison with published generation evaluations to ask whether models' human alignment in our tasks is associated with broader generation performance, beyond what can be explained by overall violation sensitivity.

\subsection{Evaluation choice and model matching}

We evaluated a subset of 11 image and seven video models from our primary sample for which corresponding public benchmark records were available (Table~\ref{tab:public-generation-models}). Image models are evaluated on Artificial Analysis text-to-image Elo \citep{ArtificialAnalysis2026-image-methodology}, derived from blinded pairwise human judgments across shared prompts. Video models are evaluated on the automated VBench Quality score \citep{Huang2024-vbench}, which assesses visual and temporal fidelity using pretrained vision models and motion algorithms (spanning consistency, flickering, smoothness, aesthetics, and imaging quality).\footnote{Snapshots retrieved September 11, 2026 from \url{https://artificialanalysis.ai/image/leaderboard/text-to-image} and \url{https://huggingface.co/spaces/Vchitect/VBench_Leaderboard}.} Model correspondence was established from published model names and official checkpoint links; models without corresponding public evaluation records (e.g., JiT, PixelGen, LTX) were excluded.

\begin{table}[!htbp]
\centering
\caption{\textbf{Models matched to public generation evaluations.} Parameters refer to the scoring backbone in billions. The same IDs, colors, and markers identify models in Figures~\ref{fig:public-generation-alignment} and~\ref{fig:public-generation-partial}. Published names do not guarantee identical checkpoint revisions or generation pipelines.}
\label{tab:public-generation-models}
\footnotesize
\setlength{\tabcolsep}{3pt}
\renewcommand{\arraystretch}{1.12}
\begin{tabular}{l>{\raggedright\arraybackslash}p{.30\linewidth}r>{\raggedright\arraybackslash}p{.32\linewidth}r}
\toprule
ID & Evaluated model & B & Published entry & Score \\
\midrule
\multicolumn{5}{l}{\textit{Image models: Artificial Analysis human-preference Elo}} \\
P05 & HiDream-O1-Image & 8.805 & HiDream-O1-Image & 979 \\
L01 & Stable Diffusion v1.5 & 0.86 & Stable Diffusion 1.5 & 468 \\
L02 & SDXL Base 1.0 & 2.6 & Stable Diffusion XL 1.0 & 688 \\
L03 & Stable Diffusion 3 Medium & 2 & Stable Diffusion 3 Medium & 726 \\
L04 & FLUX.1 dev & 12 & FLUX.1 [dev] & 842 \\
L05 & FLUX.1 schnell & 12 & FLUX.1 [schnell] & 807 \\
L06 & Qwen-Image & 20 & Qwen Image & 883 \\
L08 & FLUX.2 klein base 4B & 4 & FLUX.2 [klein] Base 4B & 776 \\
L09 & FLUX.2 klein 4B & 4 & FLUX.2 [klein] 4B & 864 \\
L10 & FLUX.2 klein base 9B & 9 & FLUX.2 [klein] Base 9B & 902 \\
L11 & FLUX.2 klein 9B & 9 & FLUX.2 [klein] 9B & 947 \\
\addlinespace[4pt]
\multicolumn{5}{l}{\textit{Video models: VBench Quality (\%)}} \\
V01 & CogVideoX-2B & 2 & CogVideoX-2B (Diffusers) & 82.480 \\
V02 & CogVideoX-5B & 5 & CogVideoX-5B (Diffusers) & 83.050 \\
V03 & CogVideoX1.5-5B & 5 & CogVideoX1.5-5B & 82.720 \\
V06 & HunyuanVideo & 13 & HunyuanVideo (Open-Source Version) & 85.090 \\
V07 & Wan2.1 T2V 1.3B & 1.3 & Wan2.1-T2V-1.3B (2025-05-03) & 85.230 \\
V08 & Wan2.1 T2V 14B & 14 & Wan2.1-T2V-14B & 85.590 \\
V09 & Wan2.2 T2V A14B & 14 & Wan2.2-T2V-A14B (w/o prompt-optimized) & 85.030 \\
\bottomrule
\end{tabular}
\end{table}

\subsection{Associations before and after sensitivity adjustment}

\begin{figure}[!t]
\centering
\includegraphics[width=\linewidth]{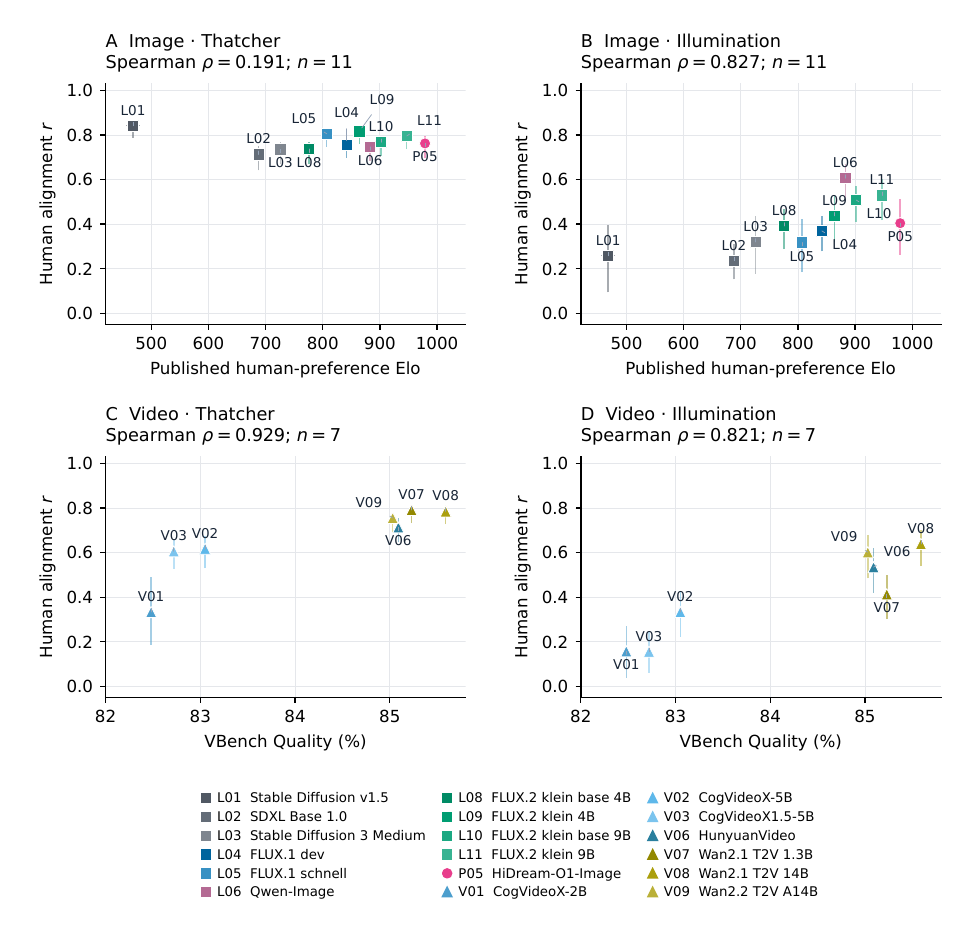}
\caption{\textbf{Published generation evaluations and human alignment before adjustment.}
Image models (A, B; $n=11$) and video models (C, D; $n=7$), with Thatcher on the left and Illumination on the right. Each point represents one model; IDs, colors, and markers match Figure~\ref{fig:public-generation-partial}. The vertical axis is human alignment ($r$). Vertical bars show 95\% participant-and-scene bootstrap CIs for each model; horizontal bars in A and B show published Elo intervals.}
\label{fig:public-generation-alignment}
\end{figure}

We retained the same 11 image and seven video models for both tasks. Let $Q$ denote the public generation score, $A$ the model's human alignment ($A_m = r_m$), and $S$ violation sensitivity ($S_m$). Before adjustment, the model-level Spearman correlations between $Q$ and $A$ were $.191$ and $.827$ for image models on Thatcher and Illumination, respectively, and $.929$ and $.821$ for video models (Figure~\ref{fig:public-generation-alignment}; Table~\ref{tab:public-generation-sensitivity-alignment}).

\begin{figure}[!t]
\centering
\includegraphics[width=\linewidth]{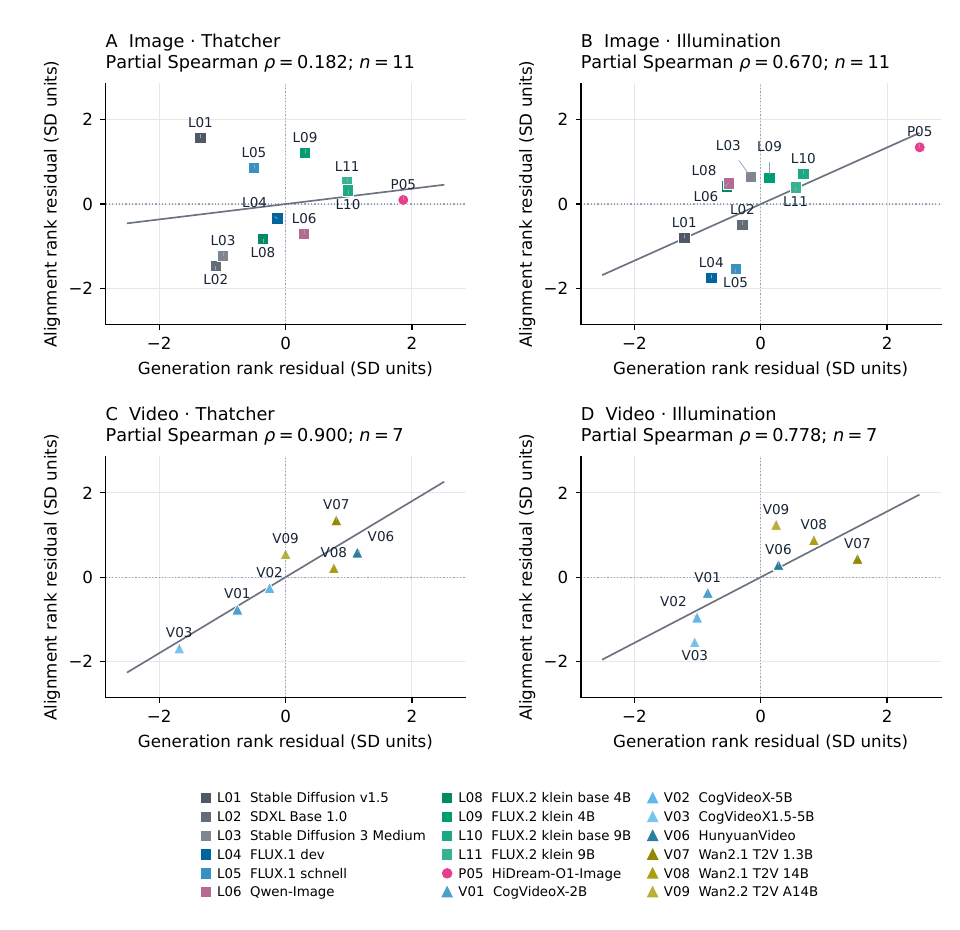}
\caption{\textbf{Generation evaluations and human alignment after controlling for violation sensitivity.}
The same models and panel arrangement as Figure~\ref{fig:public-generation-alignment}. Both axes show residuals from regressing the corresponding model ranks on sensitivity rank, divided by their own sample standard deviations. Zero denotes agreement with the sensitivity-based regression prediction. Gray lines show least-squares fits, whose slope equals the annotated partial Spearman correlation. These are the sensitivity-adjusted values in Table~\ref{tab:public-generation-sensitivity-alignment}.}
\label{fig:public-generation-partial}
\end{figure}

To control for sensitivity, we transformed $Q$, $A$, and $S$ to average ranks within each image/video task panel, separately regressed the ranks of $Q$ and $A$ on an intercept and the rank of $S$, and correlated the two residuals using Pearson correlation, without reranking them. The resulting partial Spearman coefficients were $.182$ and $.670$ for image models and $.900$ and $.778$ for video models (Figure~\ref{fig:public-generation-partial}). Thus, positive associations remained on Illumination for image models and both tasks for video models, whereas the association on Thatcher for image models was weak.

\begin{table}[!t]
\vspace{-2pt}
\noindent\begin{minipage}[c]{0.65\linewidth}
\small
\setlength{\tabcolsep}{3pt}
\renewcommand{\arraystretch}{1.12}
\begin{tabular*}{\linewidth}{@{\extracolsep{\fill}}llrrr@{}}
\toprule
Models ($n$) & Task & $\rho(Q,A)$ & $\rho(Q,A\mid S)$ & $\rho(Q,S\mid A)$ \\
\midrule
Image (11) & Thatcher & $.191$ & $.182$ & $.201$ \\
 & Illumination & $.827$ & $.670$ & $-.045$ \\
\addlinespace[3pt]
Video (7) & Thatcher & $.929$ & $.900$ & $.122$ \\
 & Illumination & $.821$ & $.778$ & $.218$ \\
\bottomrule
\end{tabular*}
\end{minipage}\hfill
\begin{minipage}[c]{0.32\linewidth}
\caption{\textbf{Generation evaluations and human alignment.} Spearman associations of generation evaluations $Q$, alignment $A$, and sensitivity $S$; variables after $\mid$ are controlled. Scatter plots are in Figures~\ref{fig:public-generation-alignment}--\ref{fig:public-generation-sensitivity-given-alignment}.}
\label{tab:public-generation-sensitivity-alignment}
\end{minipage}
\vspace{-4pt}
\end{table}

\clearpage
\subsection{Sensitivity after accounting for human alignment}

\begin{figure}[!t]
\centering
\includegraphics[width=\linewidth]{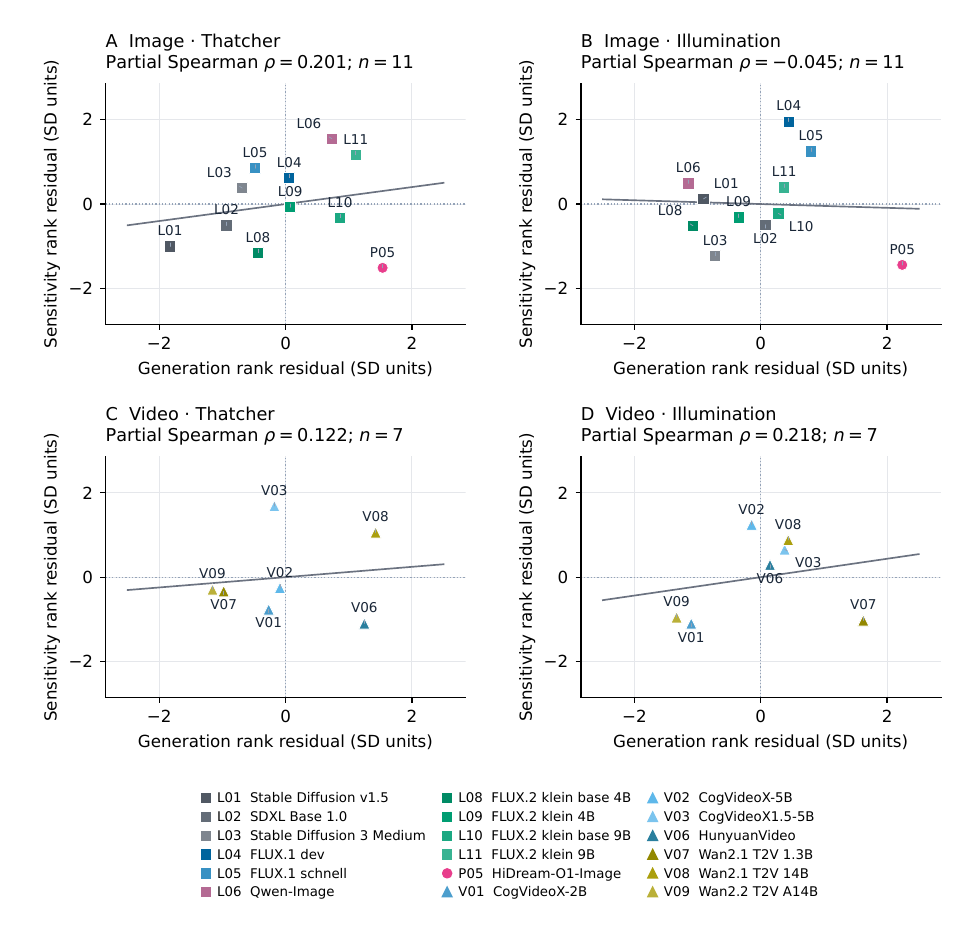}
\caption{\textbf{Generation evaluations and sensitivity after controlling for human alignment.}
Same format, models, and conventions as Figure~\ref{fig:public-generation-partial}, showing standardized rank residuals of generation evaluations (horizontal) and violation sensitivity (vertical) after regressing out human alignment rank. These are the alignment-adjusted values in Table~\ref{tab:public-generation-sensitivity-alignment}.}
\label{fig:public-generation-sensitivity-given-alignment}
\end{figure}

We also examined the reciprocal association by separately regressing the ranks of $Q$ and $S$ on an intercept and the rank of $A$, using the same matched models and residual-correlation procedure. Partial Spearman correlations were $.201$ and $-.045$ for image models on Thatcher and Illumination, respectively, and $.122$ and $.218$ for video models (Table~\ref{tab:public-generation-sensitivity-alignment}; Figure~\ref{fig:public-generation-sensitivity-given-alignment}). These weak residual associations are consistent with the possibility that generation evaluations relate more to item-level human alignment than to overall violation sensitivity alone, although given the modest sample sizes, these observations should be viewed as preliminary.

\clearpage
\subsection{Additional size adjustment and interpretation}

As a supplementary check, we added scoring-backbone parameter count $P$ (Table~\ref{tab:public-generation-models}) to the sensitivity adjustment. Partial Spearman correlations controlled for the ranks of both $S$ and $P$. Positive associations remained on Illumination for image models and both tasks for video models, while the image-Thatcher association remained weak (Table~\ref{tab:public-generation-adjustments}). Controlling for parameter count did not attenuate these associations, suggesting that the correspondence between perceptual alignment and practical generation quality persists beyond simple scaling effects.

\begin{table}[!htbp]
\centering
\caption{\textbf{Generation-evaluation associations with sensitivity and size adjustment.} All coefficients relate generation evaluation $Q$ to human alignment $A$. $S$ is violation sensitivity and $P$ is scoring-backbone parameter count. Partial Spearman coefficients control ranks of the indicated covariates.}
\label{tab:public-generation-adjustments}
\small
\setlength{\tabcolsep}{5pt}
\renewcommand{\arraystretch}{1.12}
\begin{tabular}{lllrr}
\toprule
& & & \multicolumn{2}{c}{Partial Spearman} \\
\cmidrule(lr){4-5}
Models & Task & $n$ & $S$ & $S,P$ \\
\midrule
Image & Thatcher & 11 & $.182$ & $.222$ \\
 & Illumination & 11 & $.670$ & $.783$ \\
\addlinespace[3pt]
Video & Thatcher & 7 & $.900$ & $.913$ \\
 & Illumination & 7 & $.778$ & $.914$ \\
\bottomrule
\end{tabular}
\end{table}

\section{Spatial loss-difference maps across generative models}
\label{mean-spatial-loss-maps}
\label{extended-diagnostic-results}

\begin{figure}[p]
\centering
\includegraphics[width=\linewidth]{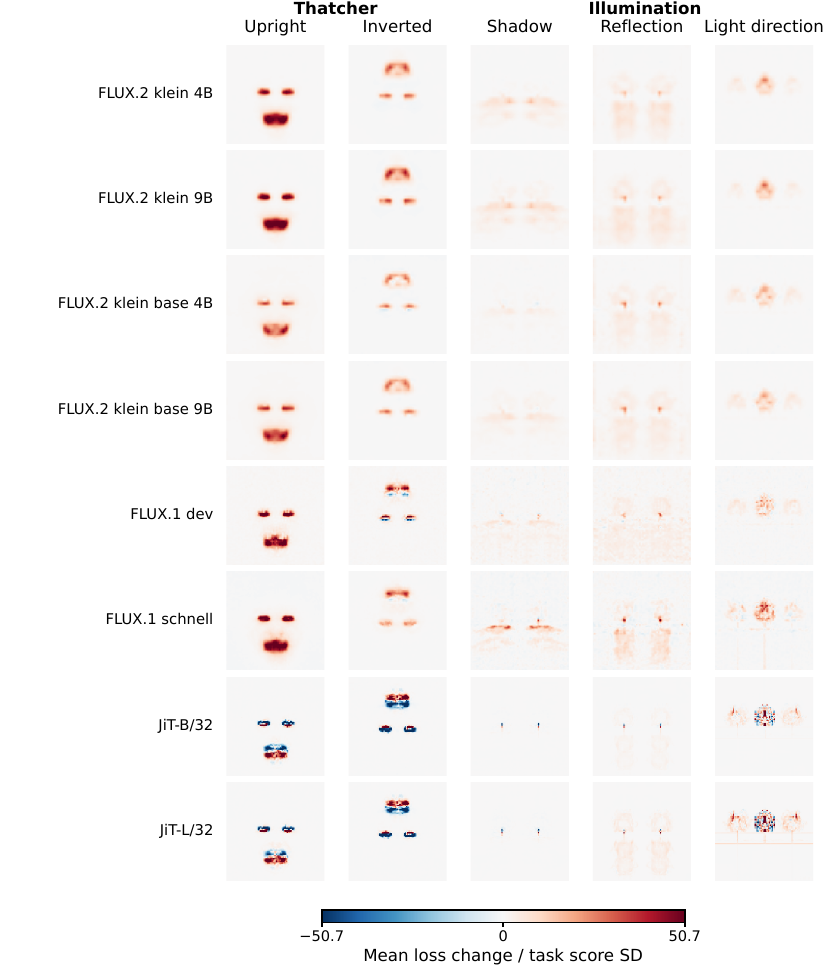}
\caption{\textbf{Mean loss-change maps for image models (Part 1 of 2).}}
\label{fig:spatial-localization}
\end{figure}

\begin{figure}[p]
\centering
\includegraphics[width=\linewidth]{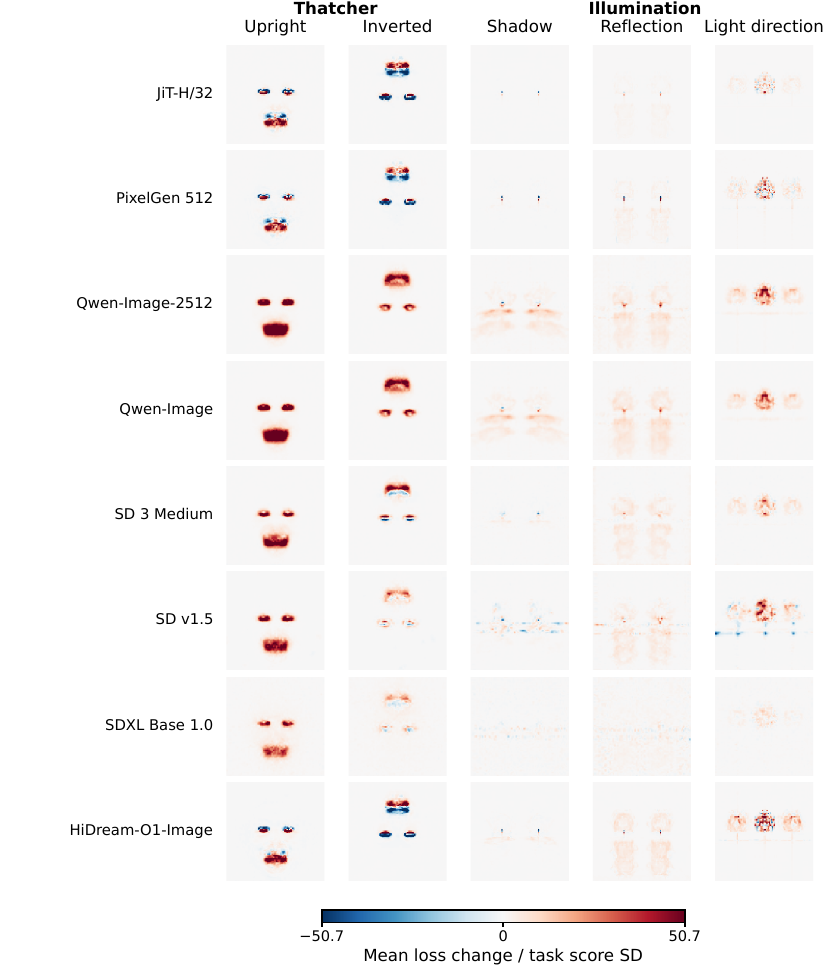}
\caption{\textbf{Mean loss-change maps for image models (Part 2 of 2).}}
\label{fig:spatial-maps-image-2}
\end{figure}

\begin{figure}[p]
\centering
\includegraphics[width=\linewidth]{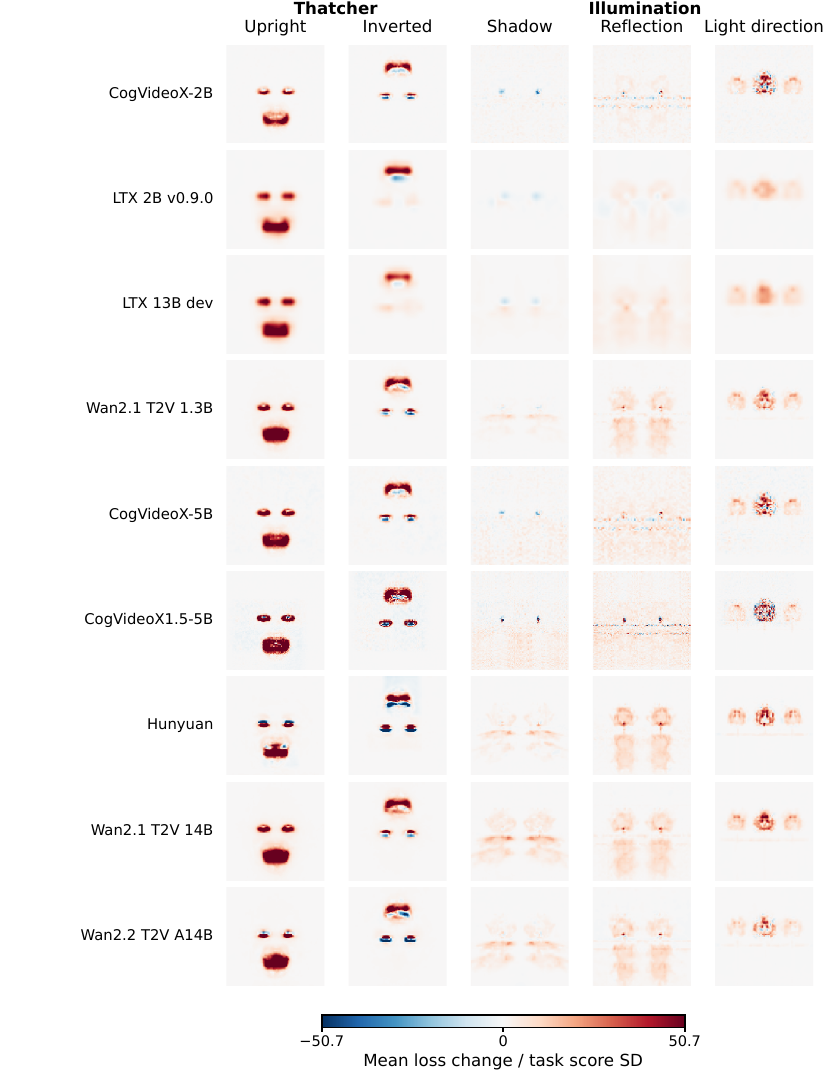}
\caption{\textbf{Mean loss-change maps for video models.}}
\label{fig:spatial-maps-video}
\end{figure}

Figures~\ref{fig:spatial-localization}--\ref{fig:spatial-maps-video} display condition-averaged loss-difference maps across all 25 generative models (16 image and 9 video models). At each native spatial position, we compute the paired difference by subtracting original-image loss from modified-image loss, averaged uniformly across timesteps and noise draws (and latent frames for video models). For model $m$, task $d$, and condition $c$, the mean loss-change map is defined as:
\begin{equation}
\overline{M}_{m,d,c}(x,y)
= \frac{1}{N_{d,c}\,s_{m,\mathrm{task}}} \sum_{i\in I_{d,c}} \left[ L^{\mathrm{mod}}_{m,i}(x,y) - L^{\mathrm{orig}}_{m,i}(x,y) \right],
\label{eq:mean-spatial-loss-map}
\end{equation}
where $I_{d,c}$ is the set of stimulus pairs in task $d$ and condition $c$, $N_{d,c}=|I_{d,c}|$, and $s_{m,\mathrm{task}}$ is the sample standard deviation of $u_{m,i}$ across all pairs in task $d$, as defined in Section~\ref{outcome-measures} ($N=280$ for Thatcher, $N=288$ for Illumination, using $N-1$ degrees of freedom). Here, $L^{\mathrm{mod}}_{m,i}(x,y)$ and $L^{\mathrm{orig}}_{m,i}(x,y)$ denote the spatial loss maps for the modified and original images of pair $i$, respectively (averaged across timesteps and noise realizations). For visualization in the figures, maps are cropped to remove video padding and resized to a common $96\times96$ square grid by bilinear interpolation, retaining each stimulus's presented orientation. All panels share a common symmetric color scale (the 99.5th percentile of absolute display-cell values across all 125 maps, $\pm 50.69$), where warm colors denote loss increases, cool colors denote loss decreases, and white indicates zero change.

\clearpage
\clearpage
\clearpage
\clearpage

\section{Timestep profiles and peak locations}
\label{timestep-profiles}
\label{timestep-profiles-and-optimization}

\begin{figure}[!htbp]
\centering
\includegraphics[width=\linewidth]{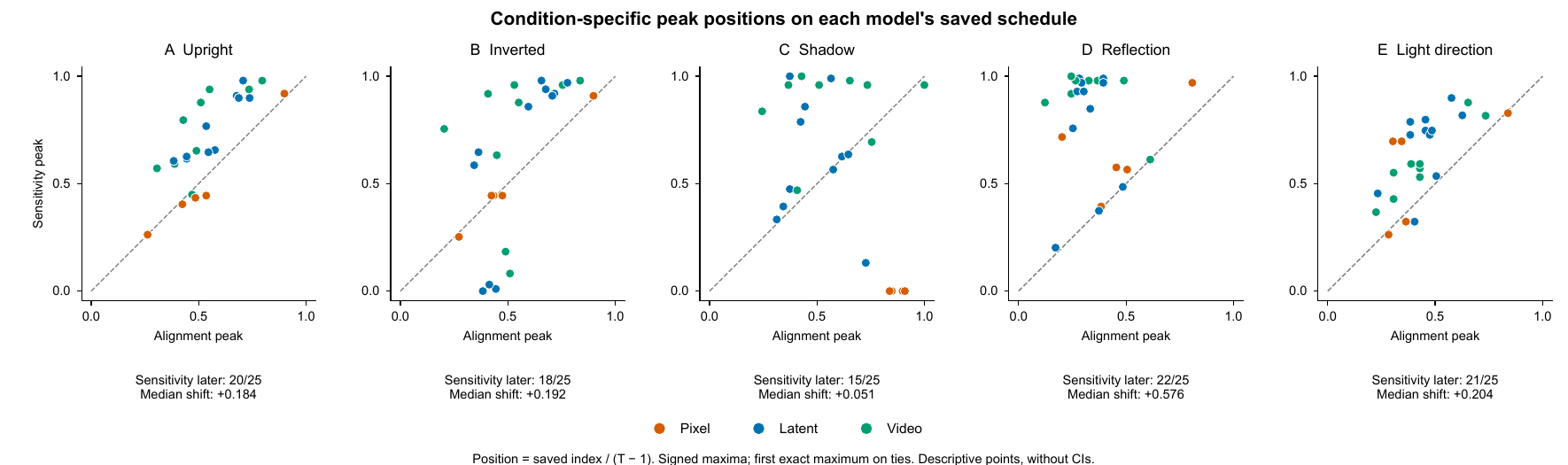}
\caption{\textbf{Condition-specific alignment and sensitivity peak positions.}
Each panel displays all 25 models (dots colored by model type: pixel, latent, and video, following Figure~\ref{fig:cross-model-sensitivity-alignment}). Horizontal and vertical axes show condition-specific alignment and sensitivity peak positions, respectively, on each model's normalized schedule ($[0,1]$). Points above the diagonal indicate a later sensitivity peak (counts printed below each panel).}
\label{fig:timestep-condition-peaks}
\end{figure}

\begin{figure}[!htbp]
\centering
\includegraphics[width=\linewidth]{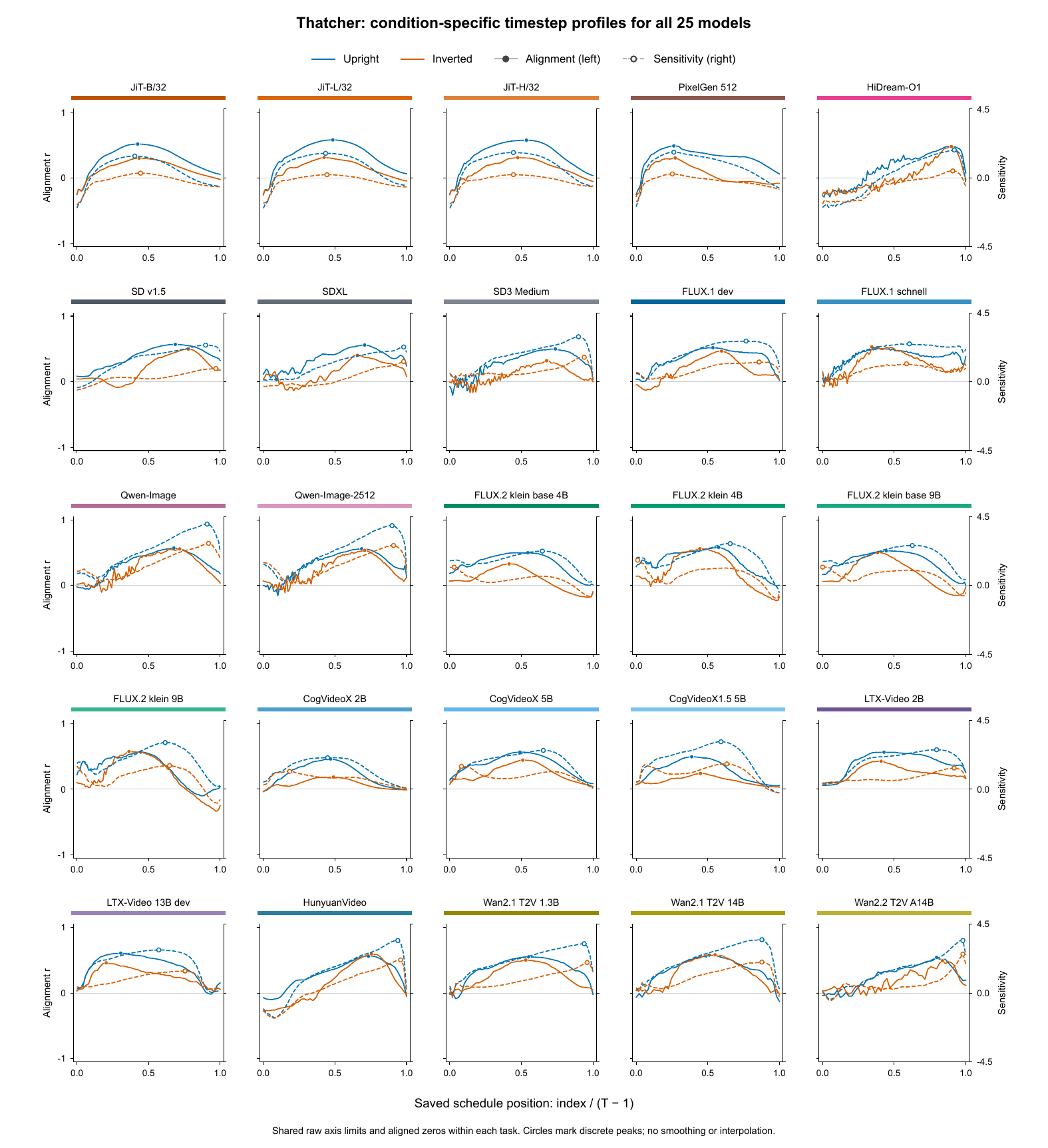}
\caption{\textbf{Condition-specific Thatcher timestep profiles for all 25 models.}
Panels show individual models (top bar colored as in Figure~\ref{fig:cross-model-sensitivity-alignment}). Blue and orange curves denote upright and inverted faces; solid curves show item-level Pearson alignment (left axis) and dashed curves show task-scaled sensitivity (right axis). Filled and open circles mark alignment and sensitivity maxima, respectively. The horizontal axis indicates normalized schedule position ($[0,1]$).}
\label{fig:timestep-logsnr-all-model-conditions}
\end{figure}

\begin{figure}[!htbp]
\centering
\includegraphics[width=\linewidth]{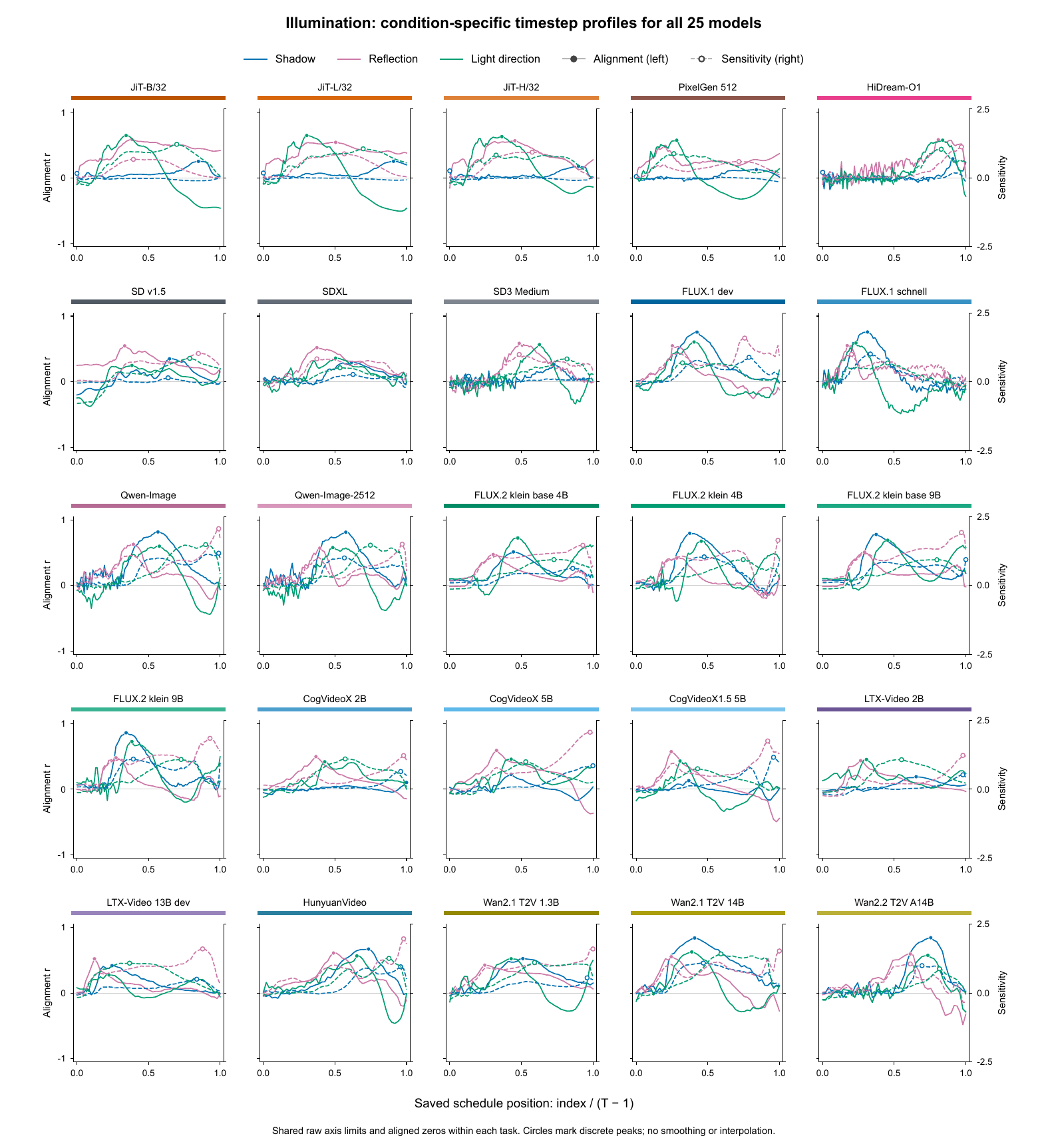}
\caption{\textbf{Condition-specific Illumination timestep profiles for all 25 models.}
Panels show individual models (top bar colored as in Figure~\ref{fig:cross-model-sensitivity-alignment}). Blue, purple, and green curves denote shadow, reflection, and light-direction interventions; solid curves show item-level Pearson alignment (left axis) and dashed curves show task-scaled sensitivity (right axis). Filled and open circles mark alignment and sensitivity maxima, respectively. The horizontal axis indicates normalized schedule position ($[0,1]$).}
\label{fig:timestep-all-model-conditions-illumination}
\end{figure}

In the main text (Section~4, Figure~\ref{fig:timestep-profiles-and-peaks}), we observed that sensitivity tended to peak later than human perceptual alignment along diffusion denoising schedules. This section provides the underlying evaluation details and complete model-wise data: (1) formal definitions of the profile metrics and schedule coordinates, (2) condition-specific peak comparisons across all 25 models (Figure~\ref{fig:timestep-condition-peaks}), and (3) complete profiles for all evaluated models across conditions (Figures~\ref{fig:timestep-logsnr-all-model-conditions} and~\ref{fig:timestep-all-model-conditions-illumination}).

\paragraph{Metric definitions and schedule coordinates.}
To compare trajectories across heterogeneous model architectures and noise schedulers, we define:
\begin{itemize}
    \setlength{\itemsep}{2pt}
    \item \textbf{Normalized schedule coordinate:} For a model with $|T|$ stored timesteps, discrete index $j \in \{0, \dots, |T|-1\}$ is mapped to $t_{\mathrm{norm}} = j / (|T|-1) \in [0, 1]$, where $0$ denotes the earliest (maximum noise) step and $1$ the final (minimum noise) step.
    \item \textbf{Standardized sensitivity:} For condition $c$ at timestep $t$, sensitivity is the mean loss difference ($L_{\mathrm{modified}} - L_{\mathrm{original}}$) divided by the sample SD ($n-1$ denominator) pooled over all items in the task at that timestep. Sharing this task-wide denominator preserves relative response magnitudes across conditions.
    \item \textbf{Human alignment:} Item-level Pearson correlation $r$ between model loss differences and human naturalness ratings, computed either pooled across the task or within each condition.
    \item \textbf{Peak locations:} Peaks are defined as signed maxima on each model's native discrete grid without smoothing or interpolation (exact ties break by the first stored index).
\end{itemize}

\paragraph{Condition-specific peak separation.}
Figure~\ref{fig:timestep-condition-peaks} presents within-model peak positions evaluated separately for each experimental condition. The extent of peak separation varied across interventions:
\begin{itemize}
    \setlength{\itemsep}{2pt}
    \item \textbf{Thatcher task:} Sensitivity peaked later than alignment in 20 of 25 models for upright faces (median shift $\Delta = +0.184$) and 18 of 25 for inverted faces (median $\Delta = +0.192$).
    \item \textbf{Illumination task:} Sensitivity peaked later in 21 of 25 models for light direction ($\Delta = +0.204$), 22 of 25 for reflection ($\Delta = +0.576$), and 15 of 25 for cast shadow ($\Delta = +0.051$).
\end{itemize}
These condition-level breakdowns show that sensitivity generally peaks later than alignment across visual manipulations. The magnitude of this shift varies across physical cues: it is largest for surface reflections and lighting direction, and smaller for cast shadows.

\paragraph{Individual model profiles.}
Figures~\ref{fig:timestep-logsnr-all-model-conditions} and~\ref{fig:timestep-all-model-conditions-illumination} display the complete condition-specific alignment (solid curves, left axis) and sensitivity (dashed curves, right axis) trajectories for each of the 25 models. Raw axis limits are held constant across all panels within each task, with vertical origins aligned at zero without within-model rescaling. Filled and open circles mark the discrete alignment and sensitivity maxima, respectively, matching the coordinates plotted in Figure~\ref{fig:timestep-condition-peaks}.

\section{Control for text conditioning}
\label{control-for-text-conditioning}
\label{conditioning-and-objective-controls}
To test whether our findings depend on empty text conditioning, we compared empty prompts with fixed descriptions of the stimulus content in a representative nine-model panel spanning diverse architectures (Stable Diffusion v1.5, SDXL, SD3, FLUX.1 dev, PixelGen, Qwen-Image, Wan2.1 T2V 1.3B, CogVideoX-2B, and LTX-Video 2B v0.9.0). Prompts described stimulus content (face orientation or a tabletop scene) without mentioning the visual manipulation or judging naturalness, and were identical for both the original and modified images within each pair (Table~\ref{tab:text-conditioning-prompts}). Both prompt conditions were evaluated without classifier-free guidance, using the same timestep grids and noise draws (20 timesteps for image models, 10 for video models; five draws per timestep). 

\begin{table}[h!]
\caption{\textbf{Fixed descriptive prompts for text conditioning.} Prompts used in the conditional evaluation. Each prompt was identical for the original and modified images within each pair and did not mention the visual manipulation or judge naturalness. The baseline condition used an empty string.}
\label{tab:text-conditioning-prompts}
\centering
\small
\begin{tabular}{@{}p{0.20\linewidth}p{0.37\linewidth}p{0.37\linewidth}@{}}
\toprule
Condition & Image models & Video models \\
\midrule
Thatcher (upright) & A close-up frontal photograph of a person's face, shown upright. & A static camera shot showing a close-up frontal view of a person's face, shown upright. \\
\addlinespace
Thatcher (inverted) & An upside-down close-up frontal photograph of a person's face. & A static camera shot showing an upside-down close-up frontal view of a person's face. \\
\addlinespace
Illumination (all conditions) & A computer-generated 3D scene of multiple geometric objects on a flat tabletop against a dark background. & A static camera shot of a computer-generated 3D scene with multiple geometric objects on a flat tabletop against a dark background. \\
\bottomrule
\end{tabular}
\end{table}

Changes in human alignment (conditional minus empty Pearson $r$) ranged from $-.034$ to $.014$ on Thatcher and from $-.021$ to $.041$ on Illumination; Table~\ref{tab:text-conditioning-control} reports the paired estimates and 95\% bootstrap confidence intervals for each model. Because the Thatcher prompts specified face orientation, we also assessed alignment within conditions: after removing orientation-specific means, correlations remained positive for all nine models under both prompt conditions. Across these nine models, human-alignment rankings had Spearman correlations of $.967$ on Thatcher and $.983$ on Illumination between prompt conditions. The sensitivity--alignment correlation changed from $.659$ to $.594$ on Thatcher and from $.947$ to $.954$ on Illumination. These results suggest that the observed differences in human alignment across models are largely preserved when simple descriptions of the stimulus content replace empty prompts.

\begin{table}[htbp]
\caption{\textbf{Paired text-conditioning control.} Human alignment (Pearson $r$) under empty and descriptive (Desc.) prompts, with $\Delta r = r_{\mathrm{Desc.}} - r_{\mathrm{Empty}}$. Brackets give paired percentile 95\% bootstrap confidence intervals for $\Delta r$ from 10,000 participant-and-scene draws (140 face identities in Thatcher or the 48 base configurations within each operation in Illumination), shared between empty and descriptive prompts.}
\label{tab:text-conditioning-control}
\centering
\small
\setlength{\tabcolsep}{3.5pt}
\begin{tabular}{@{}lrrcrrc@{}}
\toprule
 & \multicolumn{3}{c}{Thatcher} & \multicolumn{3}{c}{Illumination} \\
\cmidrule(lr){2-4}\cmidrule(l){5-7}
Model & Empty & Desc. & $\Delta r$ [95\% CI] & Empty & Desc. & $\Delta r$ [95\% CI] \\
\midrule
SD v1.5 & $.825$ & $.822$ & $-.003\;[-.006, .000]$ & $.234$ & $.268$ & $+.033\;[.017, .050]$ \\
SDXL & $.660$ & $.662$ & $+.002\;[-.028, .033]$ & $.226$ & $.233$ & $+.007\;[-.011, .025]$ \\
SD3 & $.730$ & $.744$ & $+.014\;[-.002, .031]$ & $.263$ & $.267$ & $+.004\;[-.034, .044]$ \\
FLUX.1 dev & $.751$ & $.752$ & $+.001\;[-.007, .009]$ & $.353$ & $.364$ & $+.011\;[-.012, .031]$ \\
PixelGen & $.485$ & $.493$ & $+.008\;[-.044, .060]$ & $.101$ & $.117$ & $+.016\;[-.067, .091]$ \\
Qwen-Image & $.743$ & $.709$ & $-.034\;[-.049, -.018]$ & $.576$ & $.555$ & $-.021\;[-.060, .015]$ \\
\addlinespace
Wan2.1 T2V 1.3B & $.784$ & $.779$ & $-.004\;[-.017, .008]$ & $.365$ & $.406$ & $+.041\;[.004, .083]$ \\
CogVideoX-2B & $.387$ & $.379$ & $-.007\;[-.014, -.001]$ & $.221$ & $.215$ & $-.006\;[-.044, .030]$ \\
LTX-Video 2B & $.753$ & $.747$ & $-.006\;[-.015, .003]$ & $.100$ & $.109$ & $+.010\;[-.031, .047]$ \\
\bottomrule
\end{tabular}
\end{table}

\section{Control for shared sampling noise}
\label{control-for-shared-sampling-noise}
\label{independent-noise-stress-test}
Violation sensitivity and human alignment are both functions of the same estimated paired scores. Computing both metrics from the same finite noise draws can therefore introduce correlated sampling errors, potentially inflating their cross-model association. Testing this control requires individual per-draw losses. While all 16 image models preserved per-draw losses across 100 timesteps in the primary evaluation, the primary 50-timestep video evaluation did not retain per-draw outputs. We therefore sampled three representative video models (Wan2.1 T2V 1.3B, CogVideoX-2B, and LTX-Video 2B v0.9.0) from a separate 10-timestep evaluation where per-draw losses were recorded, forming a combined panel of 19 models. Both subsets retained 20 draws and all stimulus pairs (280 Thatcher and 288 Illumination).

Within each model, we divided draws into fixed even/odd halves A and B of 10, using the same split across all items and timesteps, and recomputed sensitivity $S$ and human alignment $r$ in each half. The same-half estimate averages $\operatorname{corr}_m(S_A,r_A)$ and $\operatorname{corr}_m(S_B,r_B)$; the cross-half estimate averages $\operatorname{corr}_m(S_A,r_B)$ and $\operatorname{corr}_m(S_B,r_A)$. As a result, the same-half and cross-half associations differed by less than $.001$ in both tasks (Table~\ref{tab:shared-noise-associations}), indicating that the relationship between violation sensitivity and human alignment is not driven by shared sampling noise.

\begin{table}[htbp]
\caption{\textbf{Shared sampling noise control across 19 models.} Cross-model correlations between violation sensitivity and human alignment under same-half and cross-half draw splits, with $\Delta r = r_{\mathrm{Cross}} - r_{\mathrm{Same}}$.}
\label{tab:shared-noise-associations}
\centering
\small
\begin{tabular}{@{}lrrr@{}}
\toprule
Task & Same-half & Cross-half & $\Delta r$ \\
\midrule
Thatcher & $.6962$ & $.6971$ & $+.0009$ \\
Illumination & $.9242$ & $.9238$ & $-.0005$ \\
\bottomrule
\end{tabular}
\end{table}

\end{document}